\documentclass[11pt]{article}

\usepackage[preprint]{acl}

\usepackage{xcolor}
\usepackage{times}
\usepackage{latexsym}
\usepackage{float}
\usepackage{multirow}
\usepackage[most]{tcolorbox}
\tcbset{enhanced jigsaw, breakable}
\usepackage{enumitem}
\usepackage{bbm}
\usepackage{inconsolata} 
\usepackage{bbm}

\usepackage[T1]{fontenc}

\usepackage{amsmath}

\usepackage[utf8]{inputenc}

\usepackage{microtype}

\usepackage{inconsolata}

\usepackage{graphicx}
\usepackage{subcaption}

\title{Scaling Unverifiable Rewards: A Case Study on Visual Insights }
\author{
\textbf{Shuyu Gan},
\textbf{James Mooney},
\textbf{Pan Hao},
\textbf{Renxiang Wang}, \\[1mm]
\textbf{Mingyi Hong},
\textbf{Qianwen Wang},
\textbf{Dongyeop Kang} \\[2mm]
University of Minnesota\\[2mm]
\small
\{gan00067, moone174, pan00342, wan03409, mhong, qianwen, dongyeop\}@umn.edu 
}

\begin{document}
\maketitle
\begin{abstract}

Large Language Model (LLM) agents can increasingly automate complex reasoning through Test-Time Scaling (TTS), iterative refinement guided by reward signals. 
However, many real-world tasks involve multi-stage pipeline whose final outcomes lack verifiable rewards or sufficient data to train robust reward models, making judge-based refinement prone to accumulate error over stages.
We propose Selective TTS, a \textit{process-based refinement} framework that scales inference across different stages in multi-agent pipeline, instead of repeated refinement over time by prior work. 
By distributing compute across stages and pruning low-quality branches early using process-specific judges, Selective TTS mitigates the judge drift and stabilizes refinement.
Grounded in the data science pipeline, we build an end-to-end multi-agent pipeline for generating visually insightful charts and report of given dataset, and design a reliable LLM-based judge model, aligned with human experts (Kendall's $\tau$=0.55).
Our proposed selective TTS then improves insight quality under a fixed compute budget, increasing mean scores from 61.64 to 65.86 while reducing variance.
We hope our findings serve as the first step toward to  scaling complex, open-ended tasks with unverifiable rewards, such as scientific discovery and story generation. 
Our code and generated reports are publicly available \footnote{\url{https://minnesotanlp.github.io/insight-scaling-webpage}}

\end{abstract}

\section{Introduction}
LLM agents increasingly improve performance by allocating more inference-time compute to \emph{branch, reflect, and verify} before committing to answers.
Approaches such as Self-Consistency \citep{wang2023selfconsistencyimproveschainthought}, Tree-of-Thoughts \citep{yao2023treethoughtsdeliberateproblem}, and evolutionary refinement like AlphaEvolve \citep{novikov2025alphaevolvecodingagentscientific} show that iterative reasoning and resampling at test time can scale solution quality comparably to model or training-time gains \citep{snell2024scalingllmtesttimecompute, liu20251bllmsurpass405b}.


Many complex, real-world tasks involve multi-stage pipelines, such as analysis, generation, evaluation, and refinement, where the final outcomes are \textit{unverifiable} (e.g., insightfulness of generated charts, engagingness of generated story).
In most cases, these unverifiable objectives lack of sufficient training data to train robust reward models due to the difficulty of defining the problem and underlying principles to define the problem tasks.
In such cases, practitioners often rely on LLM-as-Judge feedback \citep{zheng2023judgingllmasajudgemtbenchchatbot}, yet repeated refinement using judge-generated scores tends to accumulate and amplify early misjudgments. Consequently, traditional time-based TTS can drift away from genuinely human-preferred outcomes, degrading rather than improving quality.

\begin{figure}[t]
  \includegraphics[width=1.05\linewidth,trim=3cm 3.6cm 3cm 0.1cm,clip]{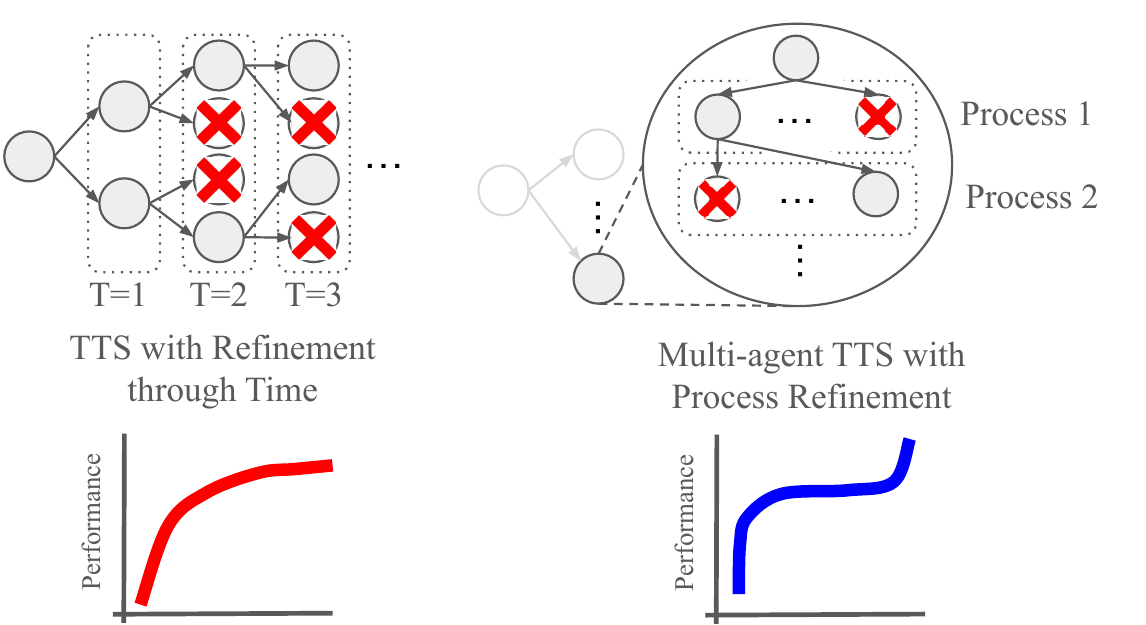}
  \caption{Comparison between traditional TTS through refinement over time (left) \citep{novikov2025alphaevolvecodingagentscientific}, and our proposed process refinement in multi-agent pipeline (right).
  They show different styles of scaling behaviors.
  }
  \label{fig:conceptual}
\end{figure}

\begin{figure*}[t]
  \includegraphics[width=\linewidth,trim=0.5cm 4.6cm 3.5cm 4cm,clip]{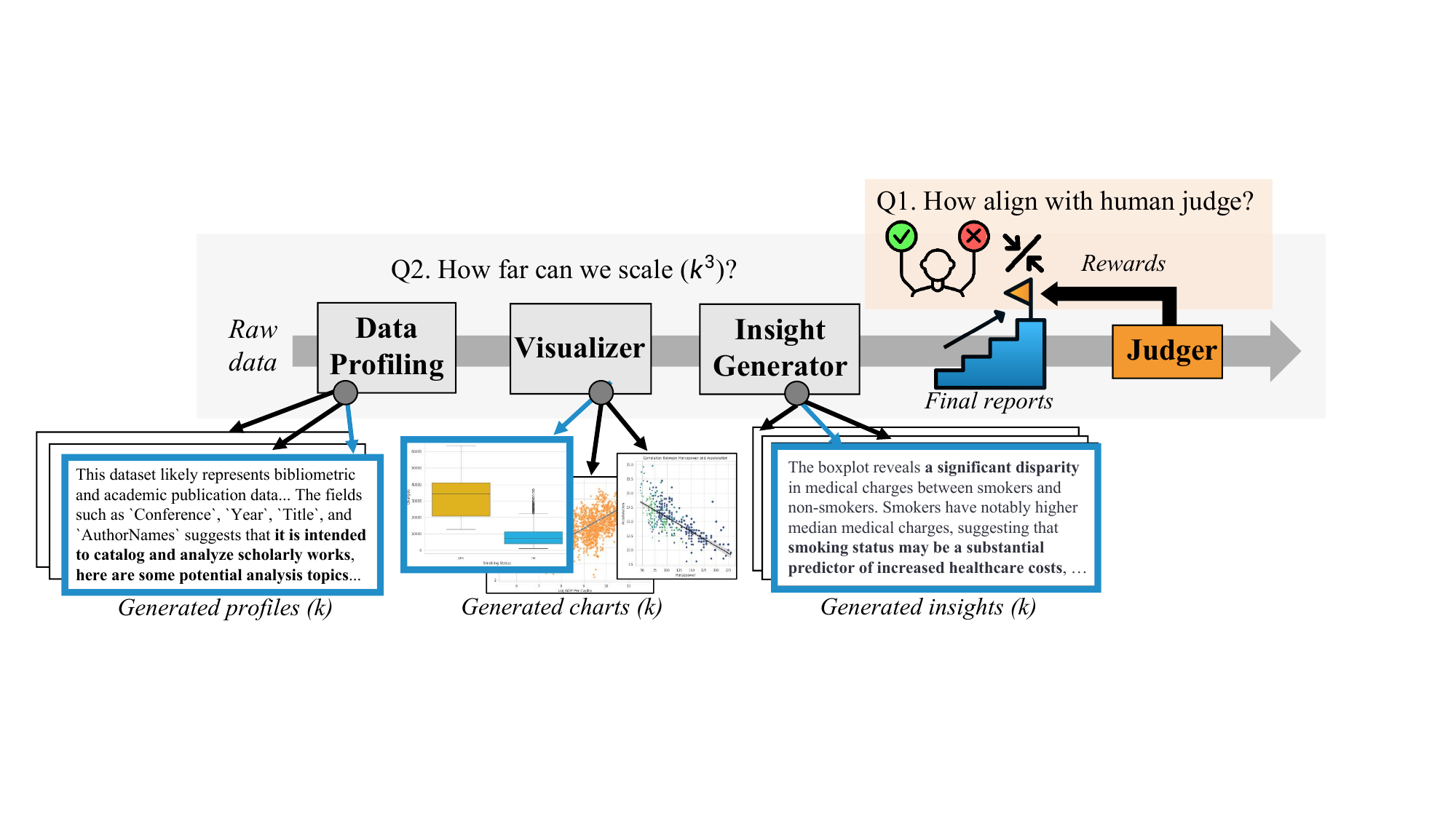}
  \caption{Our multi-agent data analysis pipeline aims to answer two key questions: (RQ1) whether judge-guided scaling can align with human experts, and (RQ2) how much performance can be further scaled within the same compute budget using the proposed selective TTS.}
  \label{fig:pipline}
\end{figure*}


To address these challenges, we propose Selective Test-Time Scaling (Selective TTS), a \emph{process-based, stage-wise} refinement strategy for multi-agent systems.
Instead of repeatedly refining outputs through time, Selective TTS allocates compute across pipeline stages and applies stage-specific evaluators to prune low-quality branches early. 
This design confines each judge's influence to its local stage, reducing error propagation and yielding more stable, interpretable scaling behavior (Fig.~\ref{fig:conceptual}).
While traditional refinement accumulates noise over time, Selective TTS redistributes scaling horizontally across independent pipeline executions to achieve consistent quality improvements under fixed compute.

Data analysis transforms raw data into visualizations and then into insight driven reports. Each stage, from choosing what to analyze to designing and interpreting charts, requires substantial expertise and intuition, making it difficult to evaluate the final report with a single scalar reward because the notion of a ``good'' insight is often subjective and depends heavily on domain knowledge and analytical experience~\citep{wang2025chartinsighterapproachmitigatinghallucination}.
Most prior work on automating data analysis like Data Science Agents \citep{google_dsa_labs_2025} fall into following rigid linear scripts and therefore plotting simple charts of surface-level outputs.

We take this inherently challenging and unverifiable process as our case study. This setting provides a grounded environment for systematically examining how \emph{process-based refinements} during test time can improve overall pipeline quality under a fixed compute budget.



Our contributions are threefold:
\begin{itemize}[noitemsep,nolistsep]
    \item We build an end-to-end data analysis pipeline (from raw data to final reports), decomposing the workflow into data profiling, chart generation, insight extraction, and verification.  
    \item We design stage-wise \textit{LLM-as-Judges} to evaluate the generated report quality, and identify a pseudo–ground-truth judge that best matches human expert judgment ($\tau$=0.55), providing a human-aligned reference for subsequent experiments. 
    \item We propose and evaluate \textit{Selective TTS}, a pruning-based compute allocation strategy across pipeline stages, that improves insight quality of reports (+4.2 gain, –39\% variance) under the same compute budget.
\end{itemize}
Together, these contributions reframe this unverifiable insight optimization as a 
\textit{test-time scaling and allocation} problem, demonstrating a principled recipe to scale outcome quality under unverifiable rewards, while bridging progress in agentic TTS with real-world data science workflows.

\section{Proposed Methods}
\label{sec:prelim}

We investigate whether unverifiable rewards in visual insight generation can be operationalized 
and improved via test-time scaling. 
To this end, we build a simple multi-agent pipeline that produces insightful reports from raw data, together with a human-aligned LLM-based judge to evaluate report quality (\S\ref{sec:pipeline}). 
Using this judging signal, we then extend our work to see how far we may scale quality under a fixed compute budget (\S\ref{sec:pruning}).


\subsection{Multi-Agent Data Analysis Pipeline}\label{sec:pipeline}
As shown in Fig.~\ref{fig:pipline}, the pipeline takes raw data and produces a final report with curated charts and chart-grounded insights. It comprises four stages, \emph{Data Profiling}, \emph{Visualization}, and \emph{Insight Generation} and \emph{Judger Verification} each implemented as an agent with stage-specific prompts, powered by an LLM or a vision–language model (VLM). The overall design is inspired by work by \citet{gan2025a2pvis} but simplified for the sake of efficient scaling. All examples shown in \S2.1 are illustrative outputs generated using the pipeline with \texttt{GPT-4o} instantiated as the LLM backbone on the VIS publication dataset~\citep{Isenberg:2017:VMC}.
This use is solely for producing clear examples for exposition and does not influence any experimental results reported in later sections. Detailed explanation of the pipeline design is in Appendix \S\ref{app:pipeline}

\paragraph{(1) Data Profiling.}
The \emph{Data Profiling} stage summarizes key metadata of the dataset (shape, schema, inferred types, small samples) and emits a concise \textit{metadata report} with plausible analysis directions.
This compact contract guides downstream visualization (valid columns, sensible encoding), reduce hallucinations and routine failures (empty plots), and avoid streaming raw records to the model, improving robustness and efficiency.
\begin{tcolorbox}[colback=gray!5, colframe=gray!50, 
                  boxrule=0.5pt, arc=2mm, left=4pt, right=4pt, top=3pt, bottom=3pt,
                  title={An example generated metadata report}]
\noindent \small
This dataset appears to represent a bibliographic collection of conference papers, likely covering scientific or technical domains. It contains a mixture of numeric, categorical, and textual attributes … Such data can support analyses of research trends, citation patterns, and scholarly impact. Here are several potential directions …\end{tcolorbox}

\begin{table*}[!t]
\centering
\small
\begin{tabular}
{@{}p{.28\linewidth}|p{.27\linewidth}|p{.41\linewidth}@{}}
\hline
\textbf{Easy Judge} & \textbf{Moderate Judge} & \textbf{Harsh Judge}\\
\hline
\raggedright
\textbf{Task:} objective evaluation using chart-only evidence.\\
\textbf{Traits:} Readability; OnTopic; TrendAlignment.\\
\textbf{Process:} direct observation and scoring.\\
\textbf{Scoring:} integers 0--100 per trait.\\
\textbf{Output:} JSON \{scores, evidence, conclusion\}. &
\raggedright
\textbf{Task:} objective evaluation using chart-only evidence.\\
\textbf{Traits:} Correctness; Specificity; InterpretiveValue.\\
\textbf{Process:} identify chart elements and rate insight clarity.\\
\textbf{Scoring:} integers 0--100 per trait.\\
\textbf{Output:} JSON \{scores, evidence, conclusion\}. &
\raggedright
\textbf{Task:} objective evaluation using chart-only evidence.\\
\textbf{Traits: }Correctness and Factuality; Specificity and Traceability; Insightfulness and Depth; So-what quality.\\
\textbf{\textcolor{blue}{CoT Process:}} observe chart → decompose insight → map evidence → score → conclude.\\
\textbf{Scoring:} integers 0--100 per trait.\\
\textbf{Output:} JSON \{scores, evidence, conclusion\}.
\end{tabular}
\caption{Prompt snippets for the three judgers. The overall evaluation task, scoring scale, and output format are consistent, while the evaluation traits and reasoning process become progressively more demanding. The final evaluation score is computed externally as the mean of the summed trait scores. Full templates are in Appendix~\S\ref{app:prompt}.}
\label{tab:judger-prompts-snippets}
\end{table*}

\paragraph{(2) Visualization.}
Conditioned on the metadata report, the \textit{Visualization} stage: (i) proposes analysis directions (\texttt{topic}, \texttt{chart\_type}, \texttt{variables}); (ii) compiles executable plotting code; (iii) executes and, if needed, \emph{rectifies} code using error traces; and (iv) filters meaningless figures via a chart judger. The result is a vetted set of figures ready for interpretation. (see an example plot generated below from the previous profile).
\begin{tcolorbox}[colback=gray!5, colframe=gray!50, 
                  boxrule=0.5pt, arc=2mm, left=4pt, right=4pt, top=3pt, bottom=3pt,
                  floatplacement=none,
                  title={An example generated chart}]
\noindent 
\small
        \includegraphics[width=\linewidth]{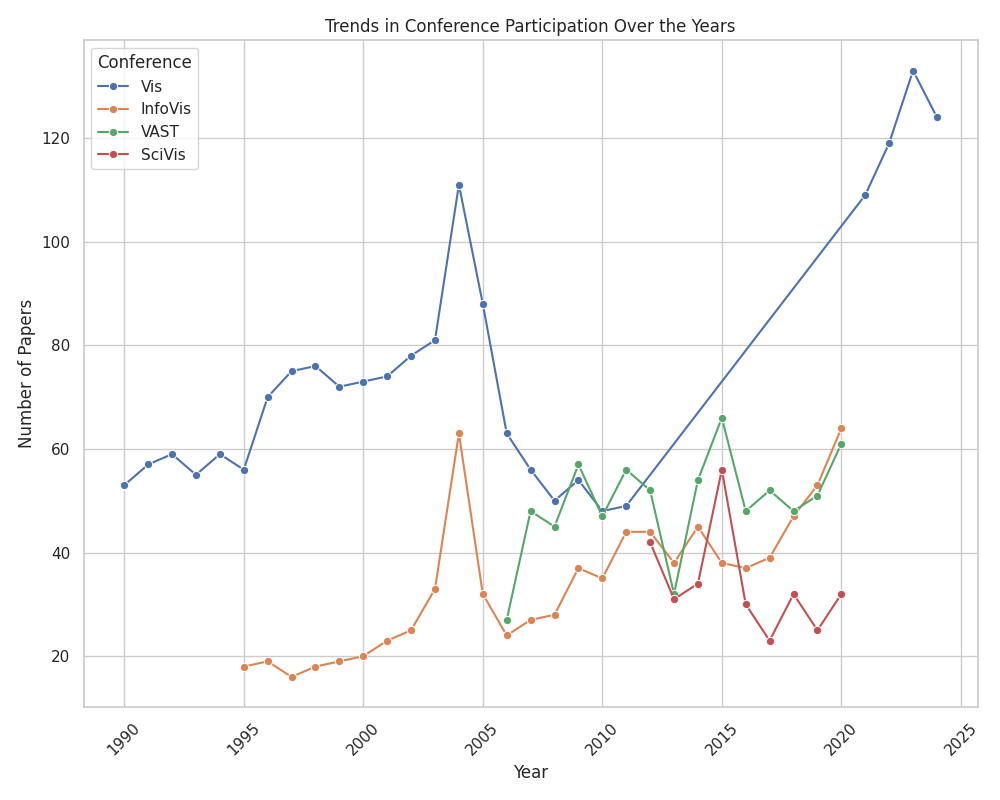}\vspace{-5mm}
    \label{fig:generated-charts}
\end{tcolorbox}

\paragraph{(3) Insight Generation.} For each vetted chart, the \textit{Insight Generation} stage samples textual insights under specific instructions that emphasizes (i) observation fidelity and evidential completeness, (ii) traceability to specific chart elements, (iii) nontriviality and novelty, and (iv) actionability. An example insight corresponding to the above figure is shown below.
\begin{tcolorbox}[colback=gray!5, colframe=gray!50, 
                  boxrule=0.5pt, arc=2mm, left=4pt, right=4pt, top=3pt, bottom=3pt,
                  title={An example generated insight}]
\noindent 
\small
Between 2005 and 2010, both the Vis and InfoVis conferences experienced a clear decline in the number of accepted papers, with Vis decreasing from around 110 papers in 2005 to roughly 60 in 2010, while InfoVis dropped from about 60 to around 40 during the same period. … This shift may reflect the field’s evolving structure, … Understanding these dynamics can help …
\end{tcolorbox}

\paragraph{Scaling and Ranking with Judges}
\label{sec:scaling_ranking}
Each (chart, insight) pair is summarized into a \textit{final report}. 
Because each stage produces multiple candidates, the total number of end-to-end report candidates grows combinatorially. 
Fixing the branching factor of each stage to $k$ yields up to $k^3$ possible report candidates.
A \emph{Judger} agent then assigns a scalar score $s\in[0,100]$ by averaging its per-dimension ratings (e.g., correctness, traceability, nontriviality, actionability). 
We instantiate three strictness levels—\emph{easy}, \emph{moderate}, and \emph{harsh}—and rank final reports by their scores for downstream selection.
Table~\ref{tab:judger-prompts-snippets} provides prompt snippets for these judgers, designed with support from domain experts and inspired by prior work on insight evaluation and characterization~\citep{law2020datainsightsprofessionalvisualization, He_2021, LIAN2025100271}.
To improve robustness and reduce stochastic judgment noise, each report is evaluated by the same judger \emph{three} independent times, and the final score is obtained via  averaging the three repeated scores. This stabilizes ranking under test-time branching and ensures consistency across evaluation runs.
\paragraph{Human Alignment of Judgers}
While judge-based ranking enables scalable selection under combinatorial growth, 
prompting-based judgment remains inherently imperfect, even when prompts are carefully 
designed using grounded literature and domain expertise. 
Different judgers may exhibit systematic biases or preference shifts, making it unclear 
whether judge-guided improvements after scaling genuinely reflect human-perceived quality.

To ensure that judge-guided scaling aligns with human expert preferences, we explicitly 
validate and select a judger whose rankings are consistent with human judgments. 
We therefore conduct a human evaluation using pairwise preference rankings over sampled 
final reports, and measure the agreement between human rankings and each candidate judger. 
This process allows us to identify a judger that reliably reflects human quality assessments, 
which is then used as the proxy signal for evaluating report quality under test-time scaling. Experimental Details are in \S\ref{sec:setup-alignment}

\subsection{Scaling Up with Selective Pruning}\label{sec:pruning}
Having established an end-to-end pipeline for visual insight generation and a human-aligned judging framework in \S\ref{sec:prelim}, 
we now test how far we can scale insight quality under a fixed compute budget (i.e., finding the best outcome out of a $k^3$ combination of $k$ profiles, $k$ charts, and $k$ reports in Fig.~\ref{fig:pipline}).

\begin{figure}[t]
  \centering
    \centering
    \includegraphics[width=1\linewidth, trim=3.6cm 1.2cm 5cm 1.2cm, clip]{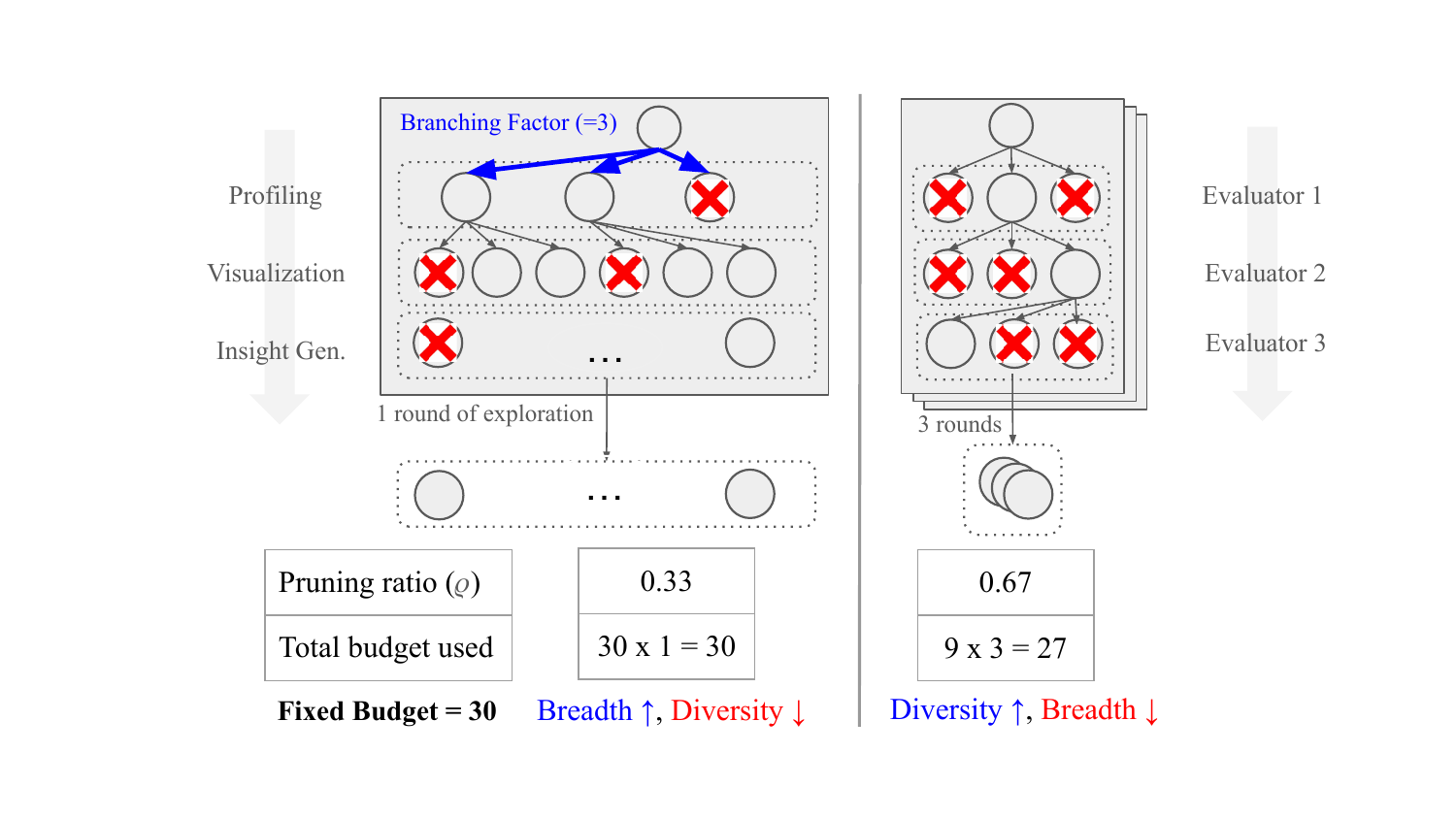}
    \vspace{-5mm}
\caption{
Overview of \textbf{Selective Test-Time Scaling} under a fixed compute budget (e.g., 30).
Stage-specific evaluators perform \textcolor{red}{pruning} across the multi-agent pipeline while controlling the \textcolor{blue}{branching} factor at each stage.
Higher pruning ratios ($\rho$) promote more rounds of exploration and diversity across runs, while reducing within-round breadth.
Details on budget accounting and branching control appear in \S\ref{sec:budget}.
}
\label{fig:prune}
\end{figure}

This section introduces \emph{Selective Test-Time Scaling} (Selective TTS), 
a process-based pruning strategy that allocates compute more efficiently across multi-agent stages. 
Unlike naïve test-time scaling that exhaustively expands all branches, Selective TTS prunes low-quality candidates early based on stage-local evaluations, allowing compute to be concentrated on more promising paths (Fig. \ref{fig:prune}).

\subsubsection{Stage-Wise Selective TTS}
Selective TTS reformulates test-time scaling along the \textit{process} dimension rather than the \textit{temporal} one.
Each round of runs proceeds through the same three core stages, data profiling, visualization, and insight generation, but with \textit{stage-local evaluators} that prune weak candidates before passing results downstream.
For instance, the data profiling stage produces multiple metadata reports;
the visualization stage expands each report into several visualization specifications (topic, variable, chart type) and render the charts; and the insight generation stage drafts multiple textual insights per chart.
After every stage, a dedicated LLM-based stage-local evaluator ranks candidates and discards a fixed fraction of low-quality outputs, enabling more compute to be allocated to high-potential branches.

\paragraph{Pruning Schedule.}
Let $b_s$ denote the branching factor at stage $s \in \{\text{data profiling}, \text{visualization}, \text{insight generation}\}$. 
Given a pruning ratio $\rho$, we retain
\[
n’_s \;=\; \max\!\big(1,\; \lceil (1-\rho)\, b_s \rceil \big)
\]
candidates to pass forward.\footnote{We enforce $\max(1,\cdot)$ to avoid pathological collapse at high pruning rates.}
Thus $\rho=0$ corresponds to the baseline (no pruning, full branching at every stage), while larger $\rho$ increases selectivity by trimming more aggressively.
We apply uniform $\rho$ across stages for simplicity, though state-specific pruning schedules are possible.

\paragraph{Stage-local Evaluators.}
Each pruning decision is guided by an LLM-based judger tailored to its stage: 
a metadata report judger ranks metadata reports, 
a visualization judger ranks directions of visual charts, 
and an insight judger ranks drafted insights. 
This process-level allocation reduces the accumulation of judging noise across time and yields more stable scaling behavior under fixed budget.


\subsubsection{Budget Accounting and Control}\label{sec:budget}
To ensure fair comparison, we hold the total compute constant across all pruning configurations
Compute is measured by the \emph{number of LLM calls} (generation + stage-local judging + final evaluation) across all agents.
Since all runs share the same model and environment, this call-based measure reliably tracks inference cost without requiring token-level accounting. For each run, the budget is computed as follows:
\begin{itemize}[noitemsep, topsep=2pt, leftmargin=1.5em]
\item \textit{Profiling:} $+b_s$ call for metadata reports generation, and $+1$ for pruning (if applied).
\item \textit{Visualization:} for each metadata report (profile), $+1$ for visualization direction generation, and $+1$ call for pruning (if applied).
\item \textit{Chart Rendering:} $+2$ calls per visualization direction (code generation + chart quality verification).
\item \textit{Insight generation:} For each verified chart, $+1$ for insight drafting, $+1$ for pruning (if applied).
\item \textit{Judge Verification:} $+1$ call per chart-insight pair (final report) for quality evaluation
\end{itemize}


Given a baseline budget $B$ ($\rho{=}0$), we adjust the number of runs such that the total LLM calls under each pruning ratio $\rho$ closely matches $B$. 
Our implementation dynamically monitors the accumulated budget during execution and incrementally supplements additional runs until the target budget is reached; the detailed procedure is described in Appendix~\S\ref{app:budget}. 
This normalization ensures that improvements observed under Selective TTS reflect more effective compute allocation rather than increased compute. 
We further validate in \S\ref{sec:selective_reliability} that LLM-call–based budgeting is a reliable proxy for token-level budgets and exhibits consistent performance trends.



\subsubsection{Budget Complexity}\label{sec:complexity}
The run-level budget complexity can be approximated as:
\[
\scalebox{0.85}{$
\mathbbm{E}[B_{\text{run}}(\rho)]
\;\approx\;
\mathcal{O}\!\big(p_v n_s'^3\big)
+\mathcal{O}\!\big(\mathbbm{I}[\rho>0]\, p_v n_s'^2\big),
$}
\]
where $n'_s = \max(1,\lceil (1-\rho)b_s\rceil)$ denotes the number of surviving branches after pruning at stage~$s$,
and $p_v \in [0,1]$ is the expected verification pass rate of chart candidates.
Specifically, $p_v n_s'$ corresponds to the expected number of charts that pass the visualization-stage
quality filter among the $n_s'$ attempted visualizations.

The first term captures the dominant generation and evaluation cost, which scales cubically with the
effective branching factor due to the combinatorial expansion across stages.
The second term represents the additional overhead introduced by pruning, which scales quadratically
in the retained branch size and is incurred only when pruning is enabled.
A full derivation appears in Appendix~\S\ref{app:complexity}.


\begin{table*}[!t]
\centering
\small
\begin{tabular}{l|ccc|ccc}
\hline
\multirow{2}{*}{\textbf{Judger}} 
& \multicolumn{3}{c|}{\textbf{VIS Publication Dataset}} 
& \multicolumn{3}{c}{\textbf{Medical Insurance Dataset}} \\
\cline{2-7}
& Kendall's $\tau$ ($\uparrow$) 
& Spearman's $\rho$ ($\uparrow$)
& Kendall's $W$ ($\uparrow$)
& Kendall's $\tau$ ($\uparrow$)
& Spearman's $\rho$ ($\uparrow$)
& Kendall's $W$ ($\uparrow$) \\
\hline
Easy     
& 0.40$\pm$0.24 & 0.53$\pm$0.24 & \textbf{0.64}
& 0.55$\pm$0.30 & 0.62$\pm$0.26 & 0.54 \\
Moderate 
& 0.45$\pm$0.17 & 0.55$\pm$0.23 & 0.51
& 0.40$\pm$0.00 & 0.55$\pm$0.08 & \textbf{0.65} \\
\textbf{Harsh} 
& \textbf{0.55$\pm$0.30} & \textbf{0.60$\pm$0.37} & 0.59
& \textbf{0.65$\pm$0.30} & \textbf{0.72$\pm$0.27} & \textbf{0.64} \\
\hline
\end{tabular}
\caption{Alignment metrics (mean $\pm$ std across 4 human annotators) between each judger and human annotations on two datasets. Higher values indicate stronger agreement (1 denotes perfect correlation).}
\label{tab:judger-metrics}
\end{table*}

\begin{figure}[h]
  \centering
    \centering
    \includegraphics[width=\linewidth]{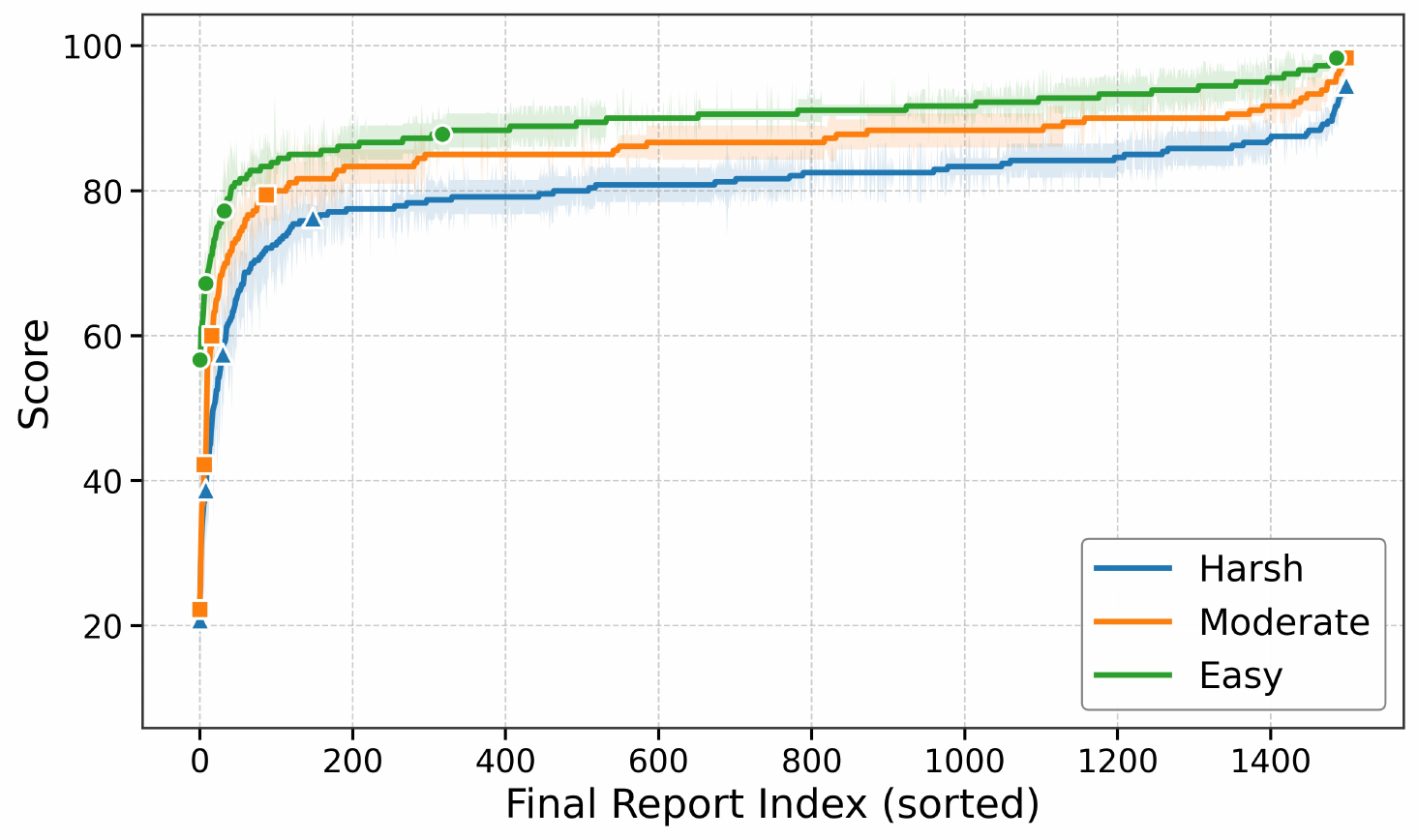}
    \vspace{-5mm}
  \caption{Sorted overall score curves under three judgers (easy, moderate, harsh) on VIS Publication dataset. See Appendix~\S\ref{app:medical} for the Medical Insurance dataset.
  }
  \label{fig:sorted-curves}
\end{figure}

\section{Experimental Setup}
\label{sec:setup}


We design two complementary experimental setups corresponding to our two research questions:  
(1) the \emph{small-sized human alignment experiment} to validate our judge selection~(\S\ref{sec:pipeline}),  
and (2) the \emph{large-scale scaling experiment} to show effectiveness of selective TTS~(\S\ref{sec:pruning}).  
\subsection{Small-Sized Human Alignment Experiment}
\label{sec:setup-alignment}

\paragraph{Sorted Index Curves.}
We make use of the Sorted Index Curves to sample at regular intervals for each judge. This is because, in contrast to refinement based TTS methods \cite{madaan2023selfrefineiterativerefinementselffeedback}, we may parallelize the production of each \emph{final report} separately given there are no dependencies between the reports. So, rather than induce a time dimension through arbitrary sampling (which would allow a more direct comparison to TTS with refinement), we instead present the results of many reports sorted in order of the judge scores. We then vertically stratify-sample reports at the 0\%, 25\%, 50\%, 75\%, and 100\% score quantiles for each judger. 
Pairwise comparisons among these five quantile-selected reports yield $3 \times \binom{5}{2} = 30$ representative evaluation points for human preference assessment.
For instance, Fig.~\ref{fig:sorted-curves} shows an example sorted score curves by three judgers in \S\ref{sec:scaling_ranking} and five vertical quantiles chosen for human alignment evaluation.
\paragraph{Human Preference Annotation and Agreement.}
The sample points are used to collect pairwise preferences from \emph{four} annotators who have extensive experience in data science and visualization.  
From these annotations, we first form a consensus human ranking $\hat{r}$ and evaluate the alignment between each judger’s ranking $r_j$ and $\hat{r}$ using Kendall’s $\tau$ and Spearman’s $\rho$ .  
To ensure that the consensus $\hat{r}$ is itself reliable, we also compute Kendall’s coefficient of concordance $W$ to assess the \emph{inter-rater agreement} among annotators. A sufficiently high $W$ indicates that human preferences are stable and do not exhibit severe internal misalignment. Definitions of all three metrics are provided in the 
Appendix~\S\ref{app:alignment-metrics}.  
We then identify the pseudo ground-truth judger $\tilde{j}$ by selecting the judger with the strongest agreement with human rankings:
\[
\tilde{j}=\arg\max_j\big(\tau_j+\rho_j\big).
\]
The resulting pseudo ground-truth corresponds to the judger most consistent with human judgment (our alignment evaluation are described in \S\ref{sec:human_judge}).
\paragraph{Report Generation and Judging Configuration.}
With \texttt{Qwen2.5-VL-32B-Instruct} as the generation backbone, we run the full pipeline with test-time branching to produce \emph{final reports}, each containing a visualization paired with its corresponding textual insight.  
To ensure that our judger design is robust, we use \texttt{GPT-4o} as a strong judging backbone throughout the human-alignment study, which helps minimize annotation bias arising from model-level limitations.  
For stability, each report is evaluated by the same judger three independent times, and the final score is obtained by averaging these repeated evaluations (majority-voting). 
We evaluate two representative tabular datasets—\emph{VIS Publication Data}~\citep{Isenberg:2017:VMC} and \emph{Medical Insurance}~\citep{dataset_insurance}\footnote{We use two diverse, multi-attribute tabular datasets to enable varied charts/insights and to test cross-domain generality.}—each yielding approximately 1{,}500 final reports.  
From these results, we obtain the \textbf{sorted index curves} of report scores, which visualize the score distribution across pruning ratios.  
These curves then serve as the basis for score-based sampling.
Detailed dataset information is provided in Appendix~\S\ref{app:datasets}.
\begin{figure*}[!htbp]
  \centering
  \begin{subfigure}[t]{0.32\textwidth}
    \centering
    \includegraphics[width=\linewidth]{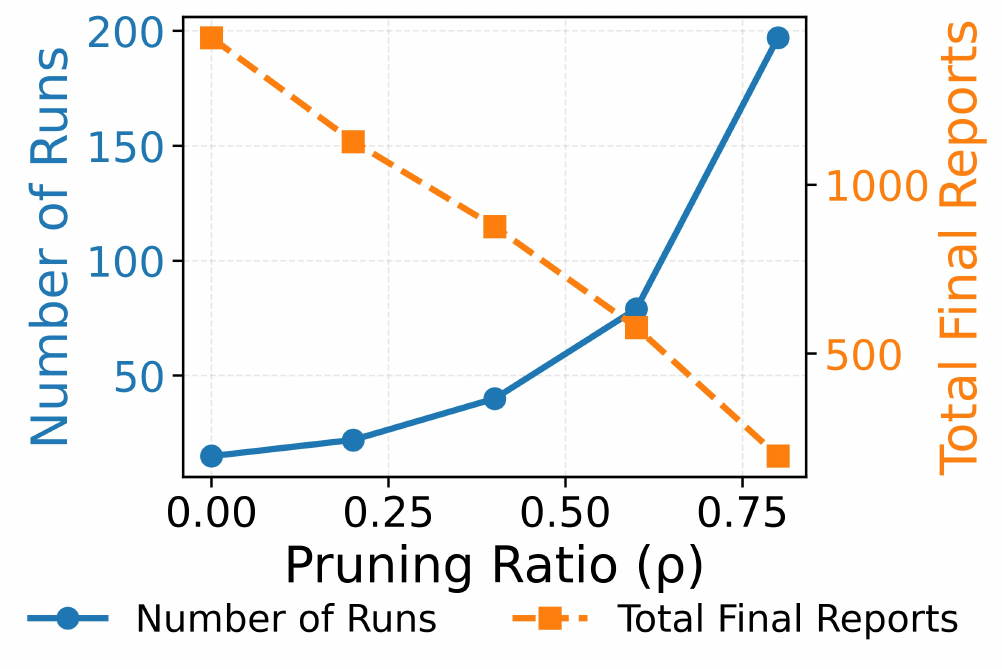}
    \caption{Runs (left) and total final reports (right) vs. pruning ratio $\rho$}
    \label{fig:runs-vips-vs-rho}
  \end{subfigure}
  \hfill
  \begin{subfigure}[t]{0.32\textwidth}
    \centering
    \includegraphics[width=\linewidth]{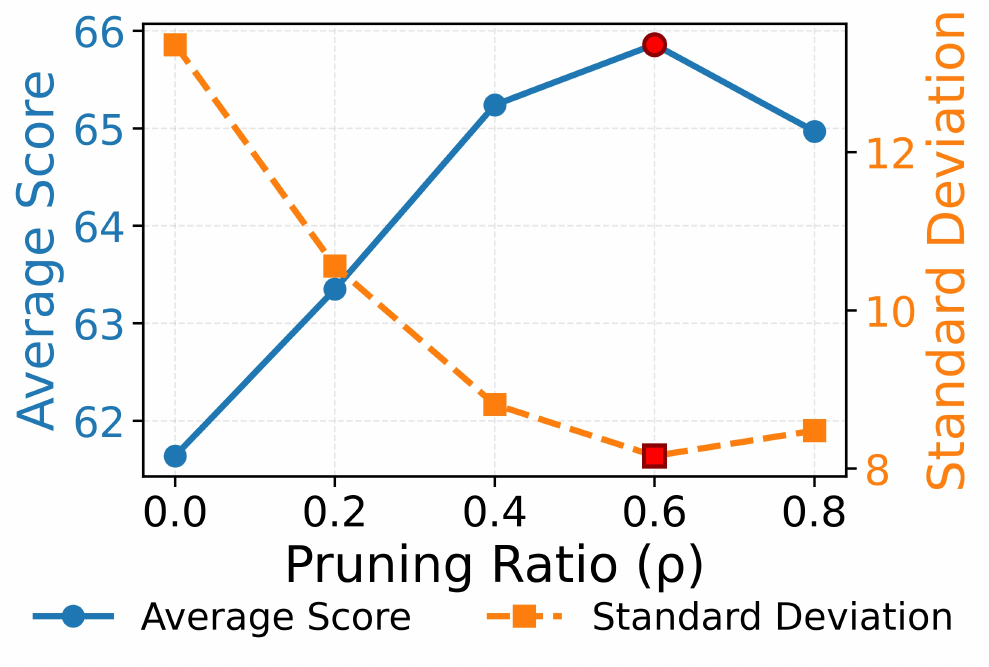}
    \caption{Average score (left) and standard deviation (right) vs. pruning ratio $\rho$}
    \label{fig:score-vs-rho}
  \end{subfigure}
  \hfill
  \begin{subfigure}[t]{0.32\textwidth}
      \includegraphics[width=\linewidth]{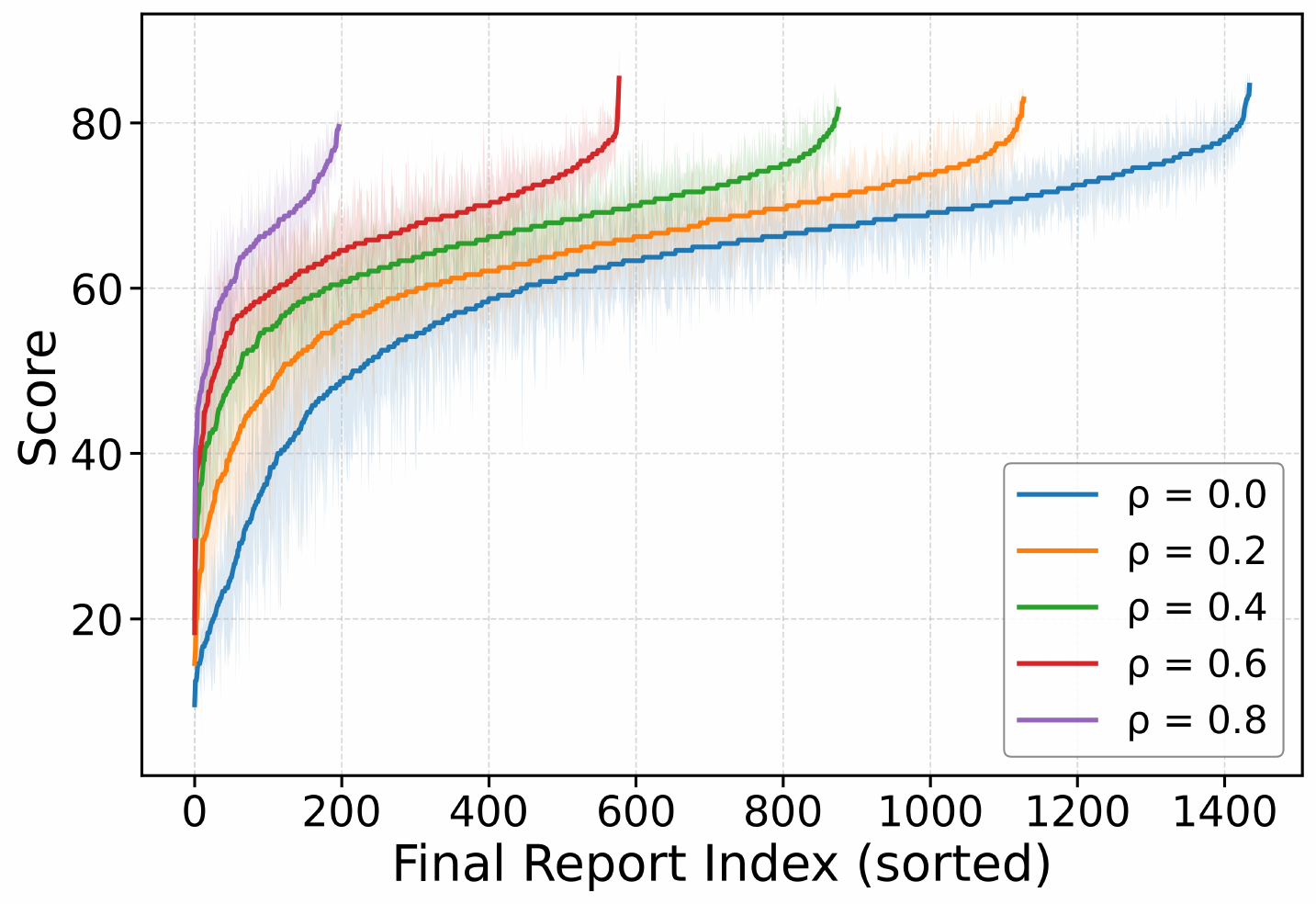}
    \caption{Sorted score curves under different $\rho$.}
    \label{fig:sorted-curves-combined}
  \end{subfigure}
  \caption{\textbf{Effects of pruning ratio $\rho$ on compute allocation and quality.}
  (a) Higher $\rho$ shifts compute from within-run breadth (fewer reports per run) to cross-run exploration (more runs).
  (b) The mean score increases with stronger pruning and peaks at $\rho = 0.6$, while the standard deviation generally decreases, indicating improved stability in report quality.
  (c) Increasing $\rho$ removes the low-quality tail and shifts the score distribution upward.
  }
  \label{fig:rho-score-run}
\end{figure*}

\subsection{Large-Scale Scaling Experiment with Selective TTS}
\label{sec:setup-scaling}
\paragraph{Model, Dataset, and Inference Configuration.}
All main experiments are conducted on the \textit{VIS Publication Dataset}~\citep{Isenberg:2017:VMC}, and we adopt \texttt{Qwen2.5-VL-32B-Instruct} as the generation backbone for all agents. 
To obtain more accurate and stable evaluations, we employ \texttt{GPT-4.1-nano} as the judging backbone(stage-local evaluators and final judger) throughout the scaling experiments. 
To ensure sufficient diversity in test-time scaling, we set the decoding parameters to top-$p=0.9$, temperature $=1.0$, and a maximum generation length of $1500$ tokens.
We fix the stage-local branching factor to $b_s = 5$, ensuring consistent diversity across all stages. 
Pruning applies a single ratio $\rho$ \emph{uniformly} across stages within a run; we sweep $\rho \in \{0.2, 0.4, 0.6, 0.8\}$, while all other settings are held fixed across runs. 
In addition, we later conduct robustness analyses examining cross-model variation, decoding parameter choices, and dataset generalization in \S\ref{sec:selective_reliability}

\paragraph{Baseline budget.}
We establish a reference budget by running the full pipeline with no pruning ($\rho{=}0$) for $15$ independent runs.
All Selective TTS conditions are matched to this budget (within a small tolerance) to isolate \emph{allocation} effects from total compute.

\section{Results}\label{sec:result}
We like to answer the two key research questions:\vspace{1mm}
\\
\noindent
\textbf{RQ1}: Do judge-guided, test-time–scaled reports align with human expert preferences? (\S\ref{sec:human_judge})

\noindent
\textbf{RQ2}: Given alignment, how far can we scale quality under a fixed compute budget?  (\S\ref{sec:selective_main}--\S\ref{sec:qualitative_human_annotation})


\subsection{Judge Selection via Human Alignment}\label{sec:human_judge}
Table~\ref{tab:judger-metrics} shows that the \textbf{harsh} judger best aligns with human preferences across both datasets (highest $\tau$ and $\rho$), with a relatively strong inter-rater agreement reflected by $W$. We adopt it as the pseudo ground-truth objective for all subsequent \textbf{Selective TTS} experiments. This establishes (i) a modular, auditable pipeline with exposed intermediate artifacts, and (ii) a human-calibrated scalar reward that is sufficiently discriminative to guide stage-wise pruning. 
We demonstrate that selective, stage-local pruning improves mean insight quality and reduces variance under a fixed compute budget, providing a practical recipe for scaling visual insight generation when rewards are unavailable.

A representative example of the human preference comparison, shown together with the corresponding judger scores, is provided in Appendix~\S\ref{app:comparison}. Further details on the annotation interface, protocol, and rater calibration procedure can be found in Appendix~\S\ref{app:annotation-interface} and Appendix~\S\ref{app:rater-calibration}.

\subsection{Selective Pruning for Scaling}\label{sec:selective_main}
We test whether Selective TTS improves overall report quality under a fixed compute budget.
Figures ~\ref{fig:runs-vips-vs-rho}-~\ref{fig:sorted-curves-combined} summarize results. 



\paragraph{Compute allocation under pruning.}
As $\rho$ increases, the total number of final reports (i.e., chart-insight pairs) decreases substantially (Fig.~\ref{fig:runs-vips-vs-rho}, right axis and Tab.~\ref{tab:main-results}), since weak branches are discarded early.
Under the same overall budget, this enables \textit{more} independent runs (left axis), \textit{trading within-run breadth for cross-run exploration}.
The total number of reports is not simply $\text{Runs}\cdot n_s'^3$ with $n'_s= \max (1, \lceil (1-\rho)\, b_s \rceil )$, because some branches fail quality gates or execution and are dropped; when $n_s'=1$ (e.g.,  $\rho=0.8$), each surviving run deterministically yields one report.
\emph{Pruning adds lightweight judging overhead but recovers budget by avoiding low-quality generations downstream}, netting a better allocation under the same total calls.

\paragraph{Quality and stability.}
Mean scores increase steadily from $\rho=0$ to $\rho=0.6$ (from $61.64$ to $65.86$), accompanied by a clear reduction in variance (from $13.36$ to $8.16$).
However, at $\rho=0.8$ (best-of-$N$ at each stage) the average score drops and the variance rises. 
This degradation may stem from over-pruning: stage-local evaluators are not always aligned with the final judger, and stronger pruning can amplify these misalignments, potentially removing a larger number of high-quality candidates.
Sorted curves (Fig.~\ref{fig:sorted-curves-combined}) show that the lower tail is pruned away and the distribution shifts upwards, while top scores are preserved.
Together, these trends indicate that Selective TTS concentrates compute on promising directions, yielding higher mean quality and lower variance under a matched budget.
Detailed numbers appear in Appendix~\S\ref{app:main_results}, Table~\ref{tab:main-results}.
\begin{figure}[!t]
  \centering
    \includegraphics[width=0.8\linewidth,trim=0 2cm 0 0,clip]{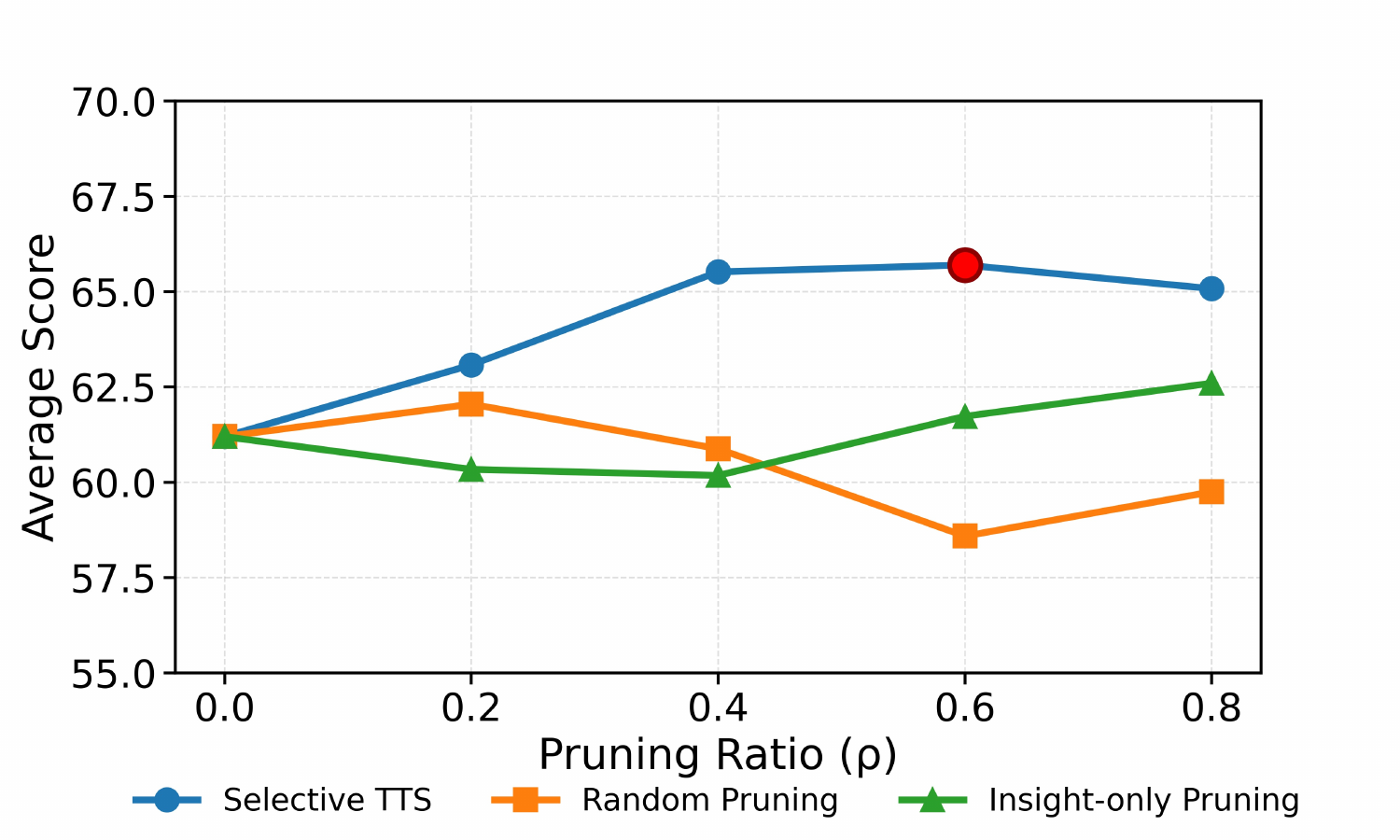}\vspace{0mm}
    \includegraphics[width=0.8\linewidth]{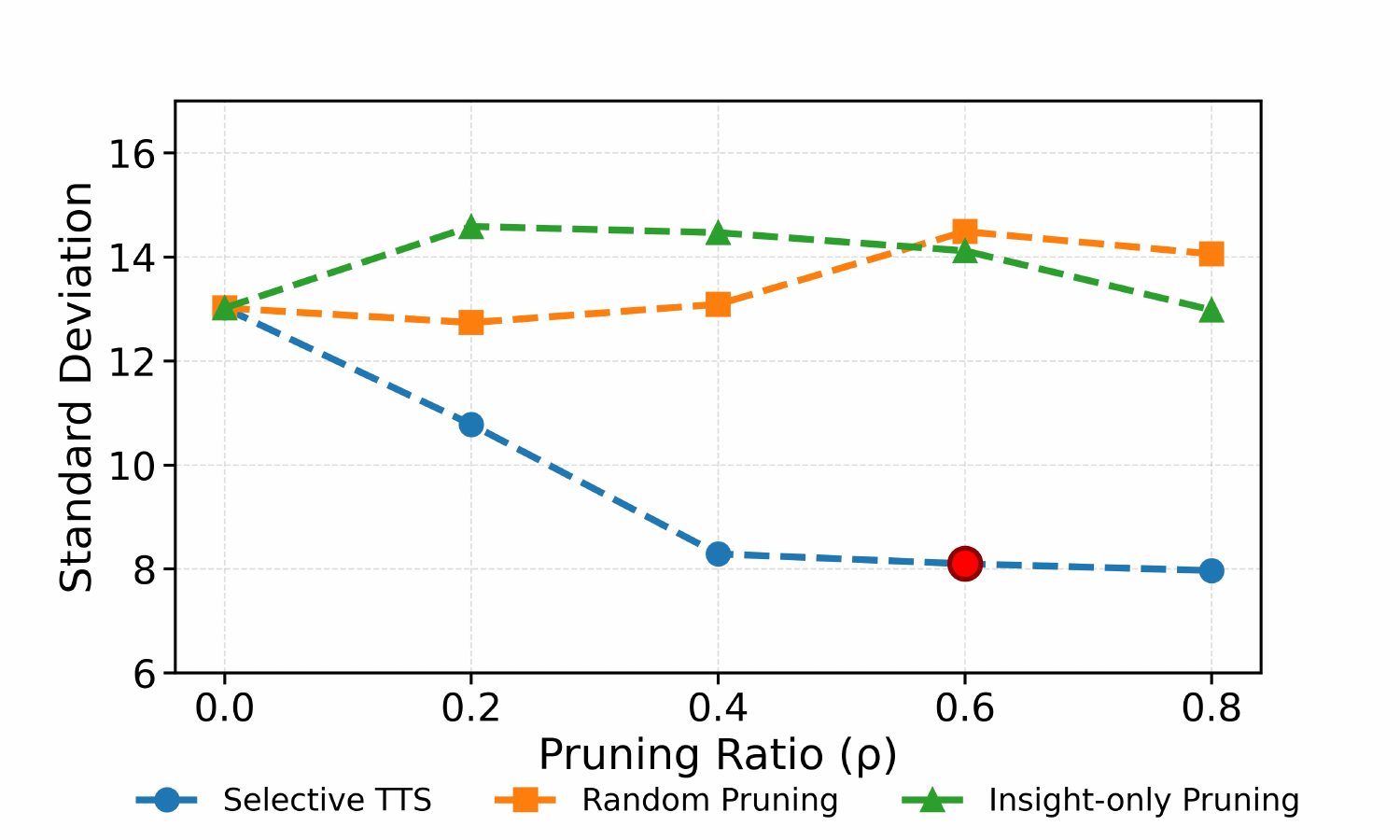}
  \caption{Ablation results comparing Selective TTS with Random Pruning and Final Stage only (Insight-Only) Pruning across varying pruning ratios $\rho$: (a) Average scores and (b) Standard deviation
  \vspace{-3mm}
} 
  \label{fig:ablation}
\end{figure}

\subsection{Ablation Study}
\label{sec:ablation}
To clarify the role of stage local pruning in Selective TTS, we conduct two ablation studies: replacing all stage local evaluators with random sampling, and applying pruning only at the final insight stage. These ablations disentangle whether the gains stem from the full selective mechanism or from late stage ranking or random exploration. For fair comparison, all experiments are run in a small scale setting with eight runs for the Selective TTS baseline, and the LLM call budget is matched across all conditions. Detailed are reported in Appendix \S\ref{app:ablation}.

\paragraph{Random Sampling.}
In this setting, we remove all stage-local evaluators and sample candidates at each stage purely at random according to the pruning ratio~$\rho$. 
Based on Fig.~\ref{fig:ablation}, 
results show that random pruning yields substantially worse performance, with mean scores decreasing as $\rho$ increases, despite similar or larger computation.  
This indicates that unguided pruning harms overall quality, suggesting that stage-local evaluators provide meaningful global benefits by filtering unpromising branches early.

\paragraph{Final Stage only Pruning.}
We analyze whether the gains arise solely from ranking at the final stage. In this setting, pruning is applied only during insight generation, while all earlier stages keep full branching. This isolates the effect of downstream ranking, since the final judge evaluates chart and insight pairs. Performance remains largely unchanged at small pruning ratios and shows only modest gains under aggressive pruning, which are substantially smaller than those achieved by full Selective TTS. Even at $\rho = 0.8$, where insight only pruning effectively becomes a best of N selection over final reports, performance still lags behind full Selective TTS. Because earlier stages are not pruned, many low quality branches persist to the final stage, limiting the impact of late stage selection alone.
\begin{figure}[!t]
\hspace*{-0.5cm}
  \centering
  \begin{subfigure}[t]{0.25\textwidth}
    \centering
    \includegraphics[width=\linewidth,trim=0.6cm 0 0.55cm 0,clip]{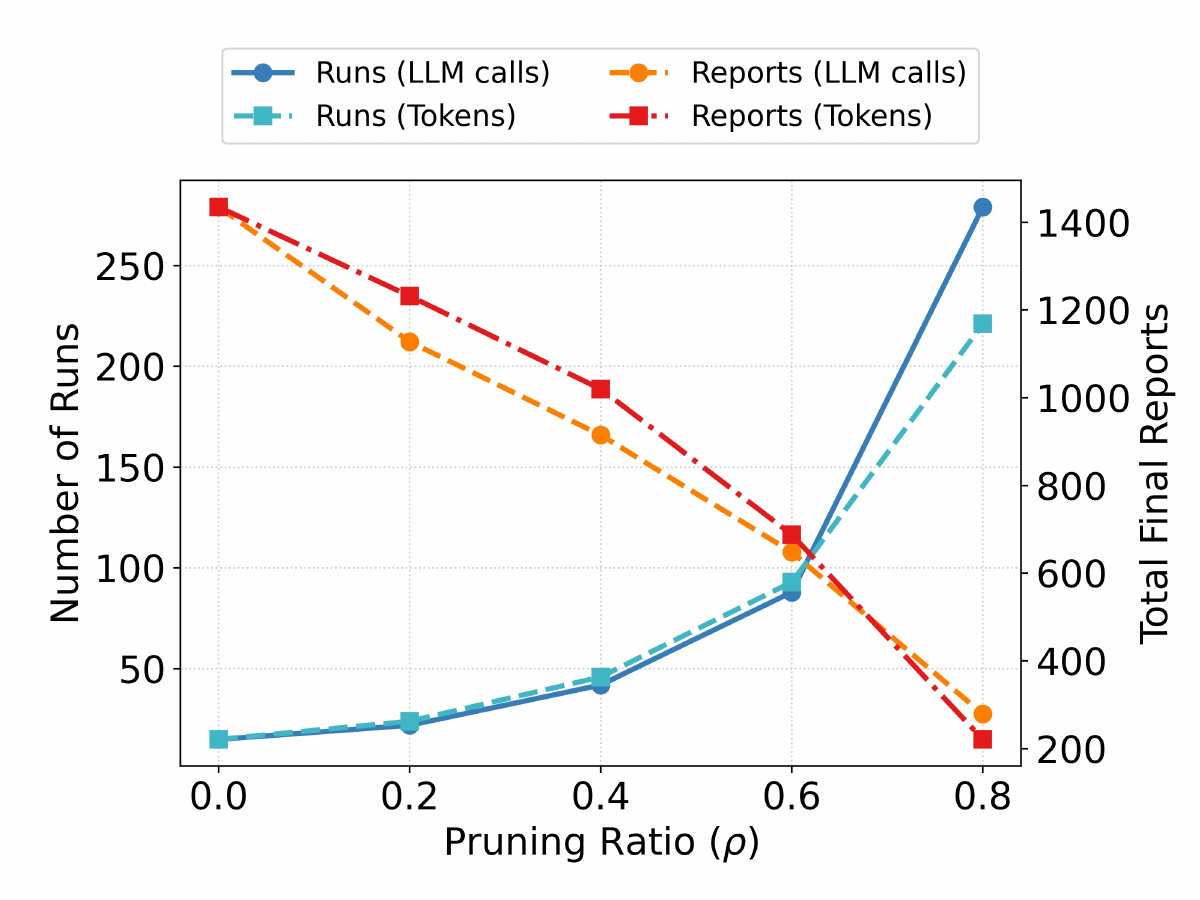}
    \caption{Runs (left) and \# reports (right) vs. pruning ratio $\rho$}
    \label{fig:ablation-score}
  \end{subfigure}
  \begin{subfigure}[t]{0.25\textwidth}
    \centering
    \includegraphics[width=\linewidth,trim=0.6cm 0 0.6cm 0,clip]{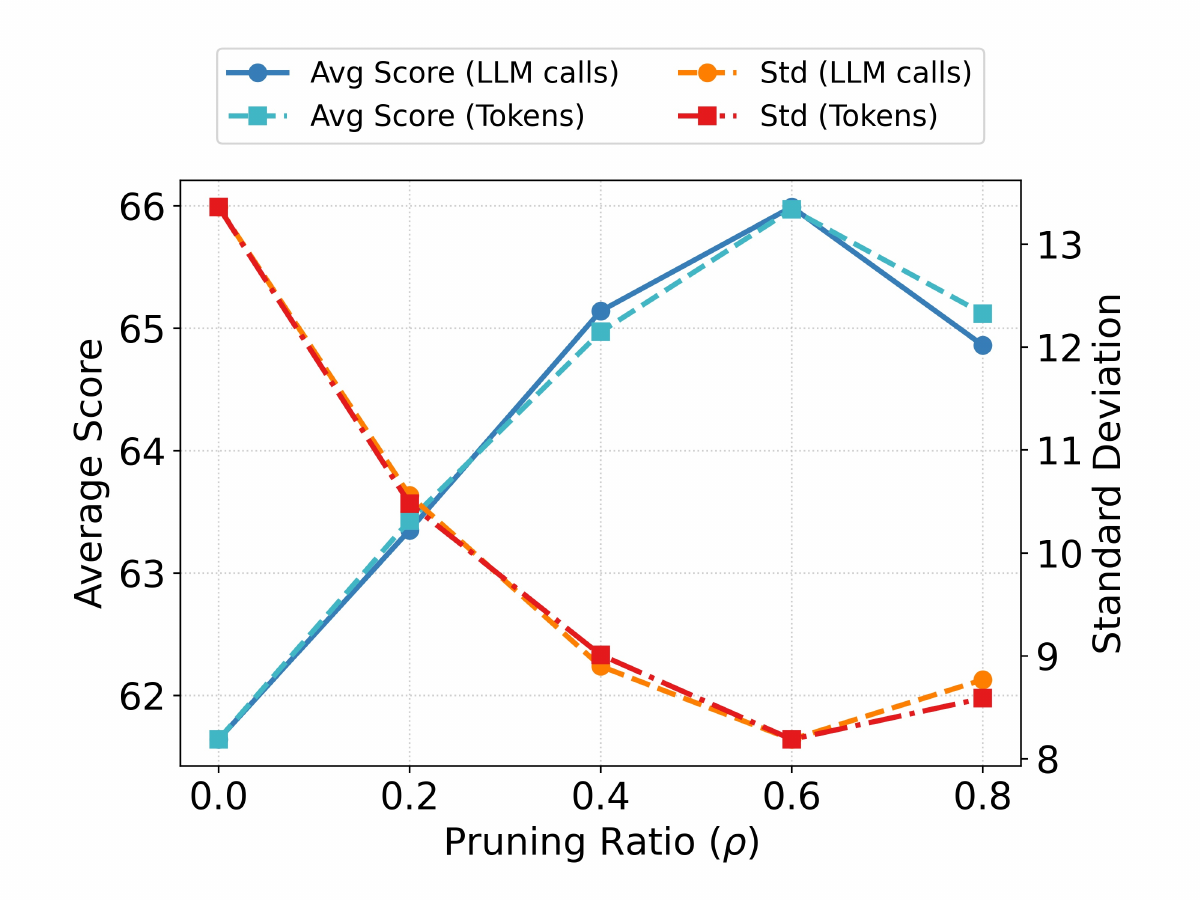}
    \caption{Score (left) and deviation
(right) vs. pruning ratio $\rho$}
    \label{fig:ablation-std}
  \end{subfigure}
  \hspace*{0.5cm}
  \caption{Effects of Selective TTS with LLM-calls as budget and
Token budget
} 
  \label{fig:token-budget}
\end{figure}

\subsection{Robustness of Scaling}
\label{sec:selective_reliability}
To evaluate the reliability and generality of Selective TTS, we perform additional analyses: comparing LLM call and token level budget accounting, examining cross-model variation, analyzing sensitivity to decoding parameters such as temperature, evaluating transfer to a different dataset, and testing robustness to output length variation. Detailed results are provided in Appendix \S\ref{app:robustness}.

\paragraph{Token-Level vs.\ LLM-Call Budget.}
To ensure that using LLM calls as the compute budget does not distort scaling behavior, we rerun the full experiment while controlling the number of output tokens instead of model invocations. As shown in Fig.~\ref{fig:token-budget}, performance curves under both budgeting schemes are highly similar. The number of runs and total generated tokens exhibit nearly identical trends across pruning ratios, indicating comparable exploration exploitation trade offs.
Since the qualitative behavior of Selective TTS is preserved under token level budgeting, LLM call budgets are a reliable proxy. In addition, LLM call budgeting is easier to control and reproduce, as token counts vary with prompt verbosity, chart complexity, and model specific generation behavior, whereas model calls provide a stable and environment agnostic compute unit.
\begin{figure}[t]
  \centering
    \centering
    \includegraphics[width=\linewidth]{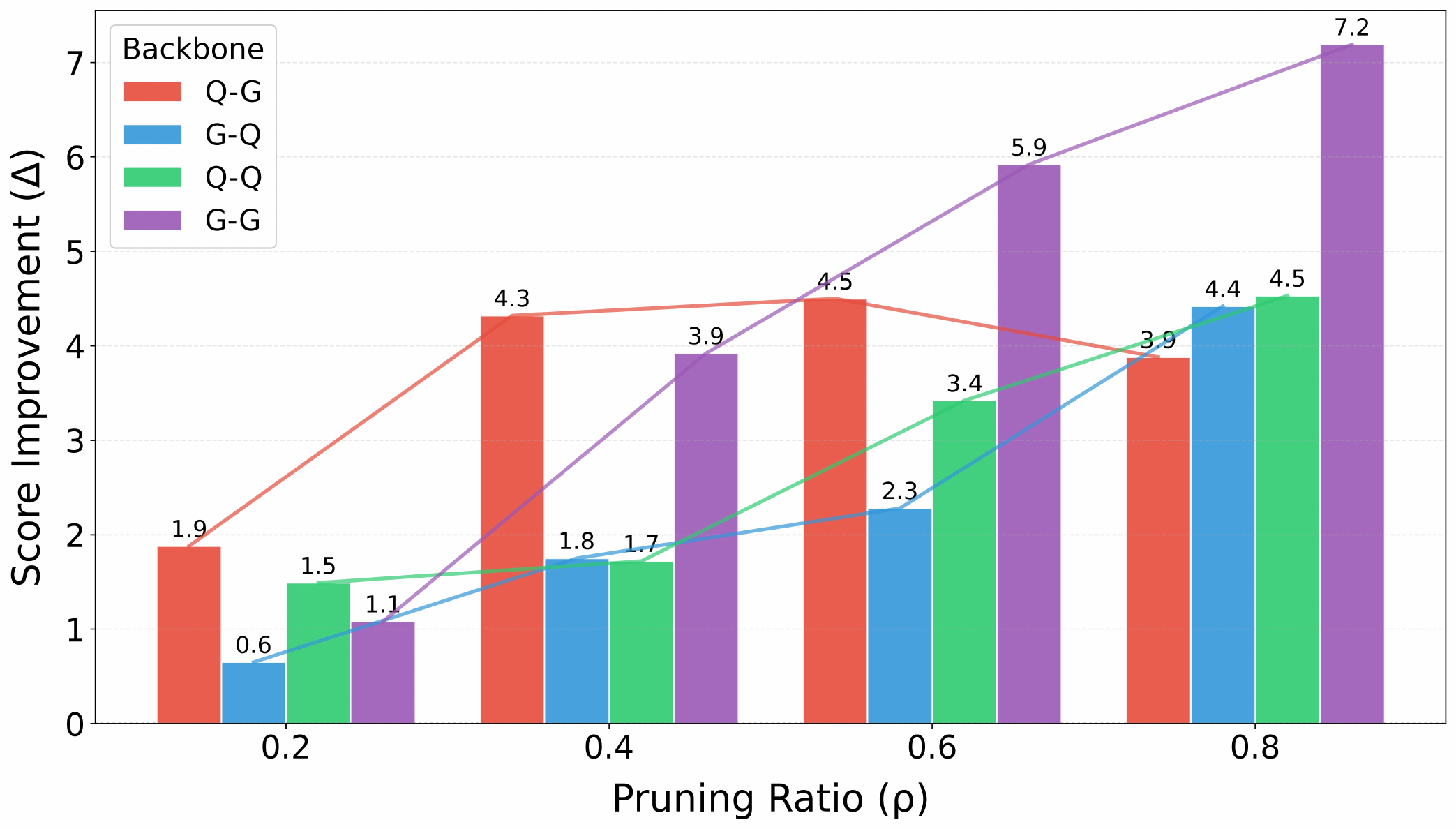}
    \vspace{-5mm}
  \caption{Score improvements ($\Delta$) achieved by Selective TTS across four generator–judger backbone combinations (Q-G, G-Q, Q-Q, and G-G). 
  }
  \label{fig:backbone}
\end{figure}

\begin{figure}[t]
  \centering
    \centering
    \includegraphics[width=\linewidth]{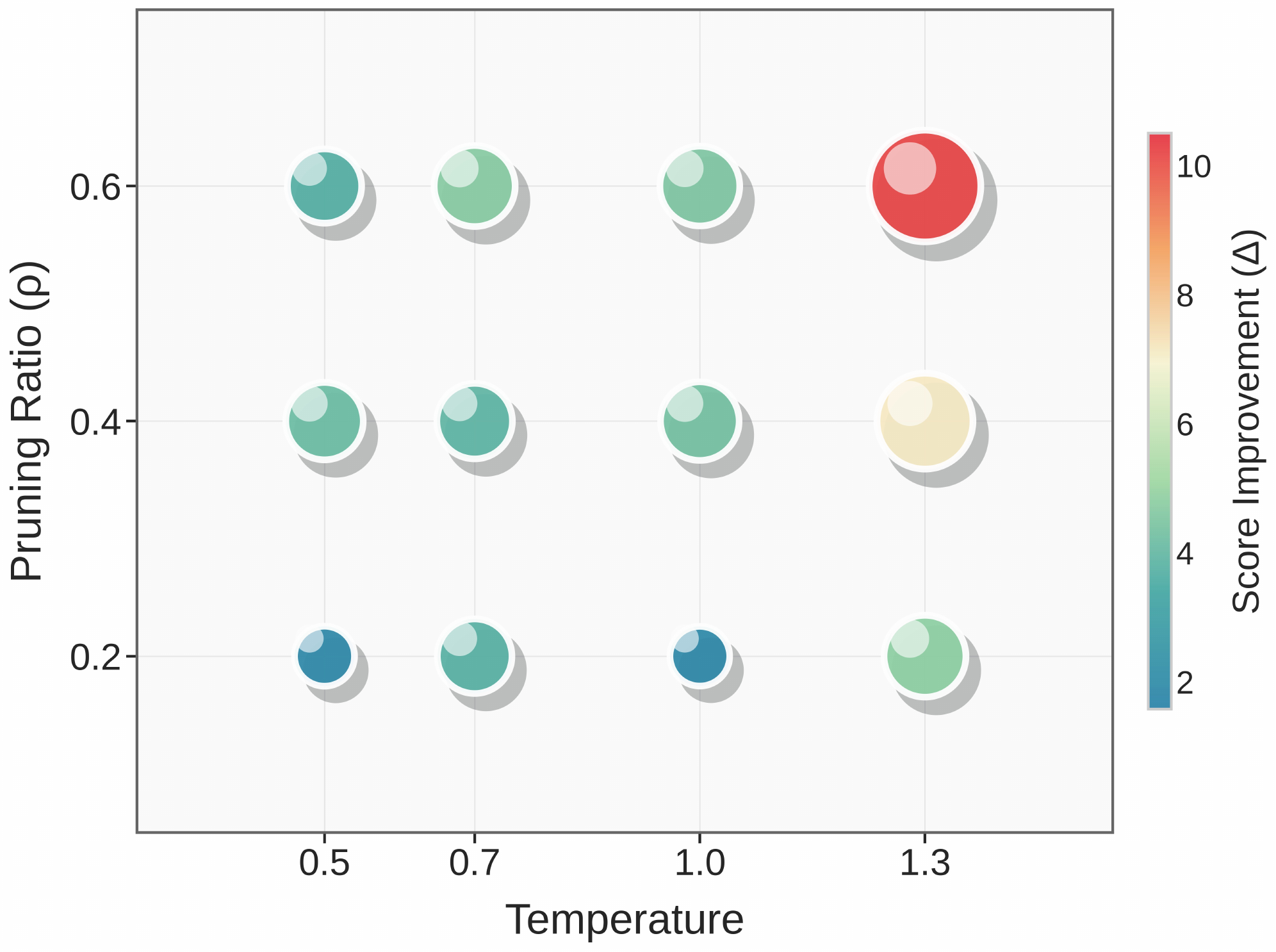}
    \vspace{-5mm}
  \caption{Score improvements of Selective TTS across different decoding temperatures and pruning ratios~$\rho$. 
Each bubble represents the improvement over the baseline ($\rho{=}0$), with color and size indicating the magnitude of the gain.}
  \label{fig:temperature}
\end{figure}

\begin{figure}[t]
  \centering
    \centering
    \includegraphics[width=\linewidth]{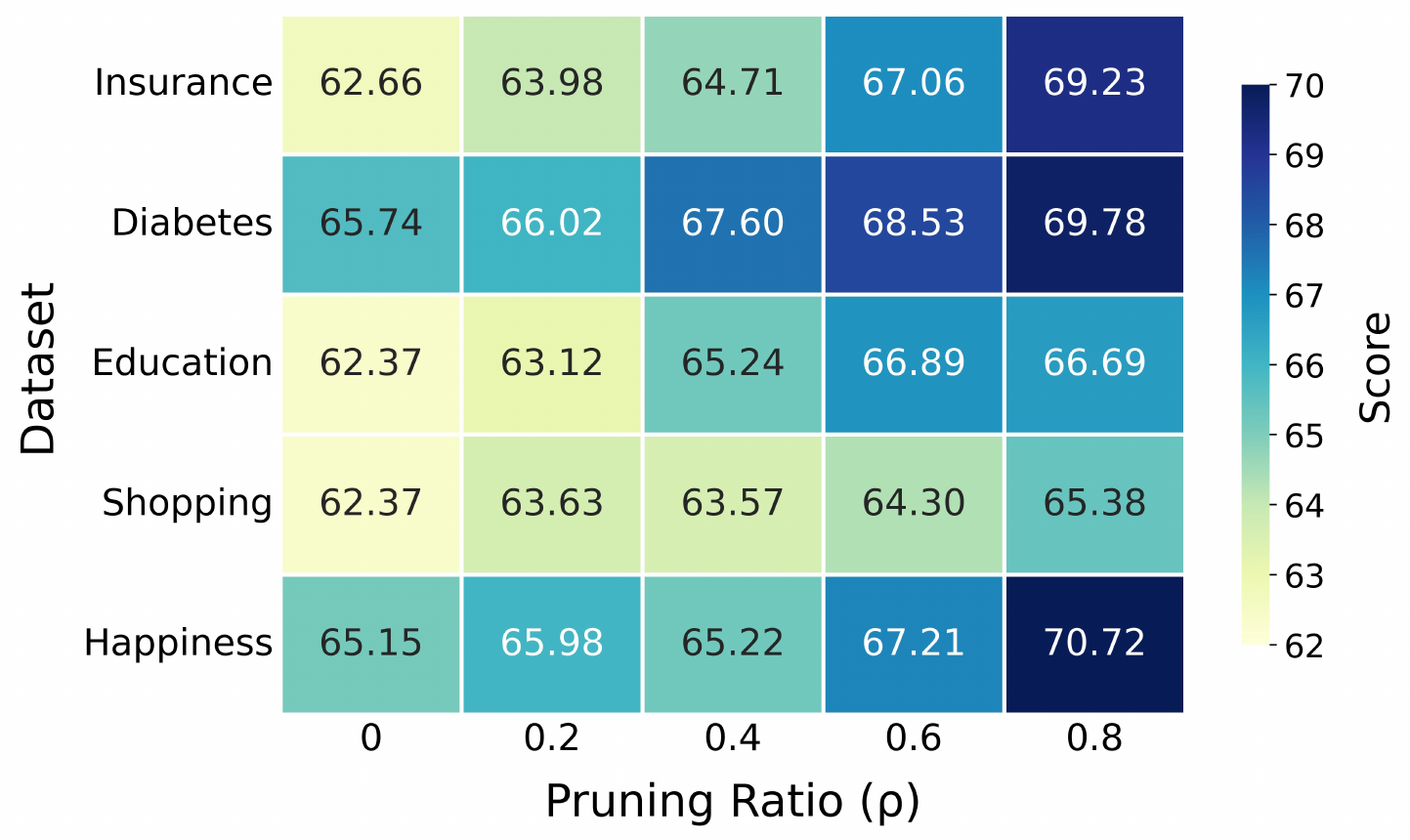}
    \vspace{-5mm}
  \caption{Score heatmap across datasets and pruning ratios~$\rho$. 
Selective~TTS shows consistent performance gains across all datasets.}
  \label{fig:generalization}
\end{figure}

\paragraph{Cross-Model Variation.}
To assess sensitivity to the model backbone, we evaluate four generator-judger pairings across two model families: Q-Q, Q-G, G-Q, and G-G, where Q denotes \texttt{Qwen2.5-VL-32B-Instruct} and G denotes \texttt{GPT-4.1-nano}. For each pairing, we run the baseline system at $\rho=0$ for eight runs to estimate a matched compute budget, then apply Selective TTS at higher pruning ratios. As shown in Fig.~\ref{fig:backbone}, all configurations show consistent improvements as $\rho$ increases. While gain magnitudes vary slightly across backbones, stronger pruning reliably improves quality in every case, indicating that Selective TTS generalizes across model families and is not driven by family specific biases.

\paragraph{Sensitivity to Decoding Parameters.}
Selective TTS benefits from diverse candidates, as pruning is effective only when the search space contains meaningful variation. We therefore use high diversity decoding by default (temperature = 1.0, top-$p$ = 0.9) to avoid overly deterministic generation.

To test sensitivity to decoding choices, we vary the generator temperature while fixing top-$p$, and apply Selective TTS with the same pruning ratios. For each temperature, we first run the baseline system ($\rho=0$) for eight runs to estimate a matched compute budget, then apply Selective TTS at higher pruning ratios under this fixed budget. As shown in Fig.~\ref{fig:temperature}, Selective TTS consistently outperforms the baseline across all temperatures, even as generation becomes more deterministic at lower values. The modest dip around 0.4 to 0.6 likely reflects reduced diversity, which weakens stage local discrimination. Nonetheless, higher pruning ratios continue to yield larger gains, and the overall improvement pattern remains stable. Extremely high temperatures introduce execution errors and reduce usable runs, as detailed in Appendix \S\ref{app:robustness} Table~\ref{fig:temperature}.

Overall, these results indicate that Selective TTS does not depend on a specific decoding regime and remains robust as the candidate distribution shifts from high to low diversity generation.

\paragraph{Generalization Across Datasets.}
To test generalization beyond the VIS dataset, we evaluate Selective TTS on five additional tabular datasets spanning diverse semantic and statistical characteristics:
\emph{Medical Insurance}~\cite{dataset_insurance}, 
\emph{Diabetes}~\cite{dataset_diabetes}, 
\emph{Education Performance}~\cite{dataset_education}, 
\emph{Customer Shopping Behavior}~\cite{dataset_shopping}, 
and the \emph{World Happiness Report}~\cite{dataset_happiness}. Dataset details are provided in Appendix \S\ref{app:datasets}. For each dataset, we run the baseline system ($\rho=0$) eight times to estimate a matched compute budget, then apply Selective TTS at higher pruning ratios.

\begin{figure}[t]
  \centering
    \centering
    \includegraphics[width=\linewidth]{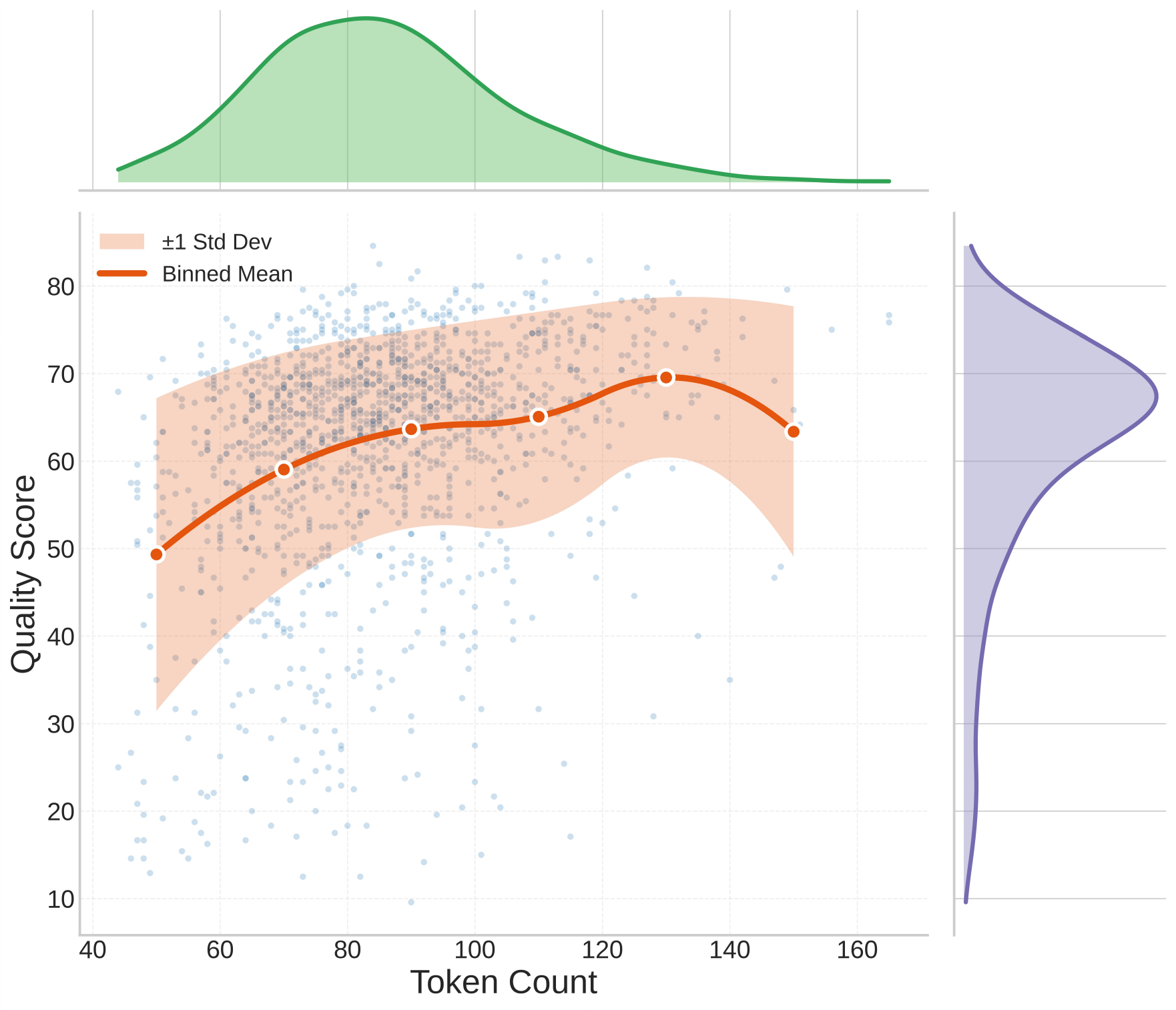}
    \vspace{-5mm}
  \caption{Insight length vs quality score.}
  \label{fig:token_distribution}
\end{figure}

\begin{figure}[t]
  \centering
    \centering
    \includegraphics[width=0.8\linewidth,trim=0cm 0 0 3.0cm,clip]{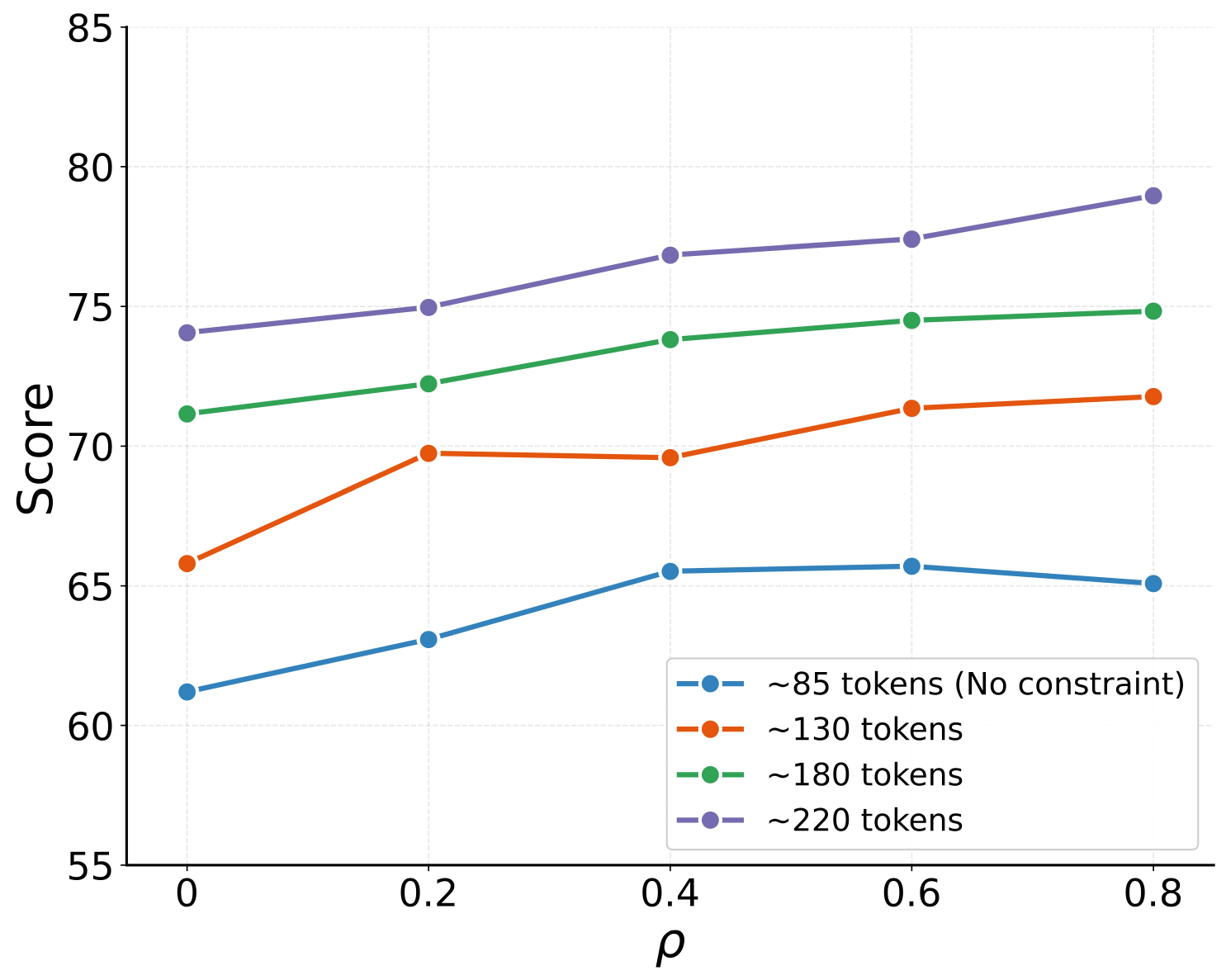}
    \vspace{-3mm}
  \caption{Scaling behavior of Selective~TTS under different insight-length regimes.}
  \label{fig:robust_tokens}
\end{figure}

\begin{figure}[t]
    \centering
    \includegraphics[width=0.8\linewidth]{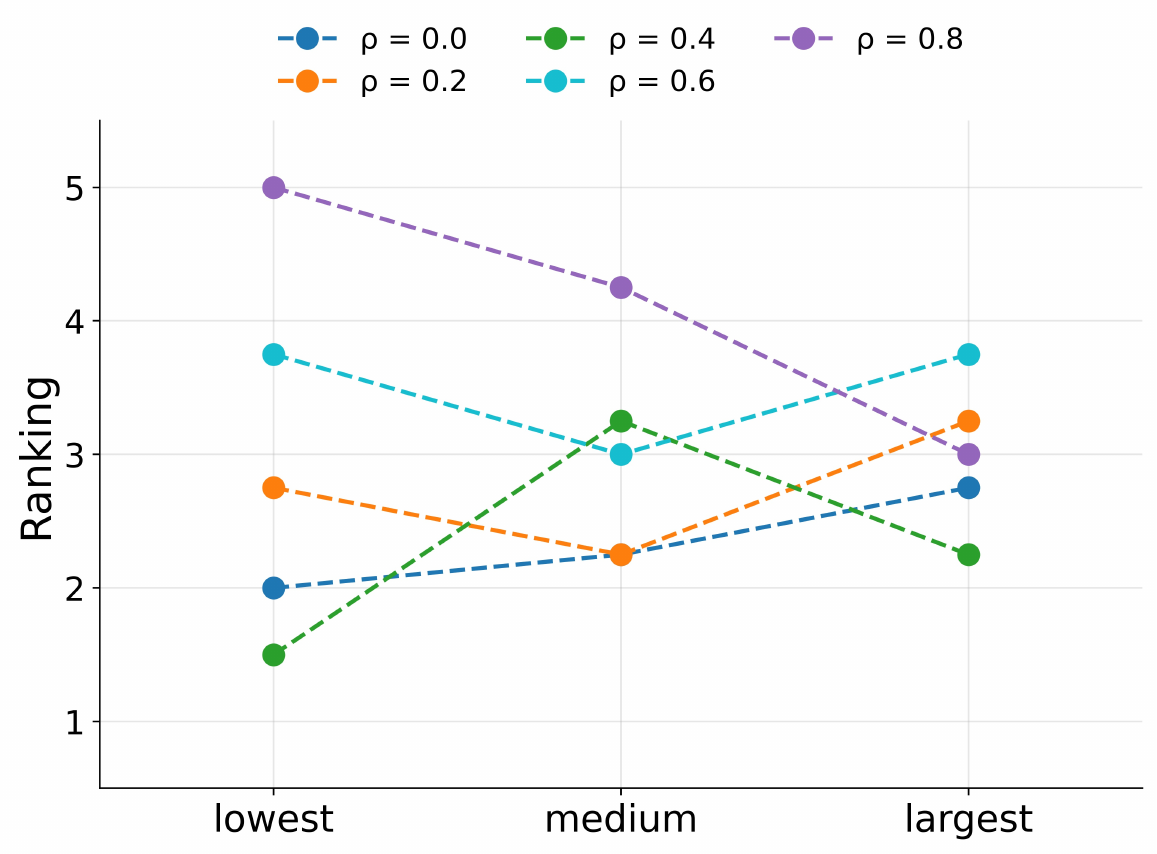}
    \caption{Mean annotator rankings across pruning ratios. Rankings converge as high-scoring report become comparably strong. (higher is better)}
    \label{fig:rankings}
\end{figure}

As shown in Fig.~\ref{fig:generalization}, all datasets show consistent score improvements as the pruning ratio $\rho$ increases, indicating that the scaling behavior is not dataset specific. Moderate pruning ($\rho \in {0.2, 0.4, 0.6}$) yields steady gains across all datasets, while the most aggressive setting ($\rho=0.8$) produces mixed results, with weaker improvements on some datasets. This suggests that excessive pruning can over eliminate promising branches depending on dataset complexity and noise. Overall, Selective TTS generalizes well across heterogeneous domains, with moderate pruning offering the most reliable trade off between exploration and pruning depth.

\paragraph{Sensitivity to Insight Length.}
Under the baseline setting ($\rho=0$), we observe a positive correlation between insight length and evaluation score: longer insights generally receive higher scores, likely because they provide richer context, clearer reasoning, and more supporting detail. Fig.~\ref{fig:token_distribution} illustrates this trend at the instance level.

Motivated by this, we test whether Selective TTS continues to yield gains as output length increases, and whether pruning remains effective beyond regimes already favored by the evaluator. To separate length effects from scaling behavior, we group insights into coarse token length regimes and evaluate Selective TTS within each group. We enforce length variation through prompt level constraints and, for each regime, run the baseline system ($\rho=0$) for eight runs to estimate the average output length.
As shown in Fig.~\ref{fig:robust_tokens}, increasing the pruning ratio $\rho$ consistently improves performance across all length regimes, with highly similar scaling trends. Meanwhile, baseline scores rise with output length, confirming that longer insights are generally rated more highly.

\begin{figure*}[t]
  \centering

  \begin{minipage}[t]{0.31\linewidth}
    \centering
    \includegraphics[trim=0cm 0cm 0cm 0, clip, 
    width=\linewidth]{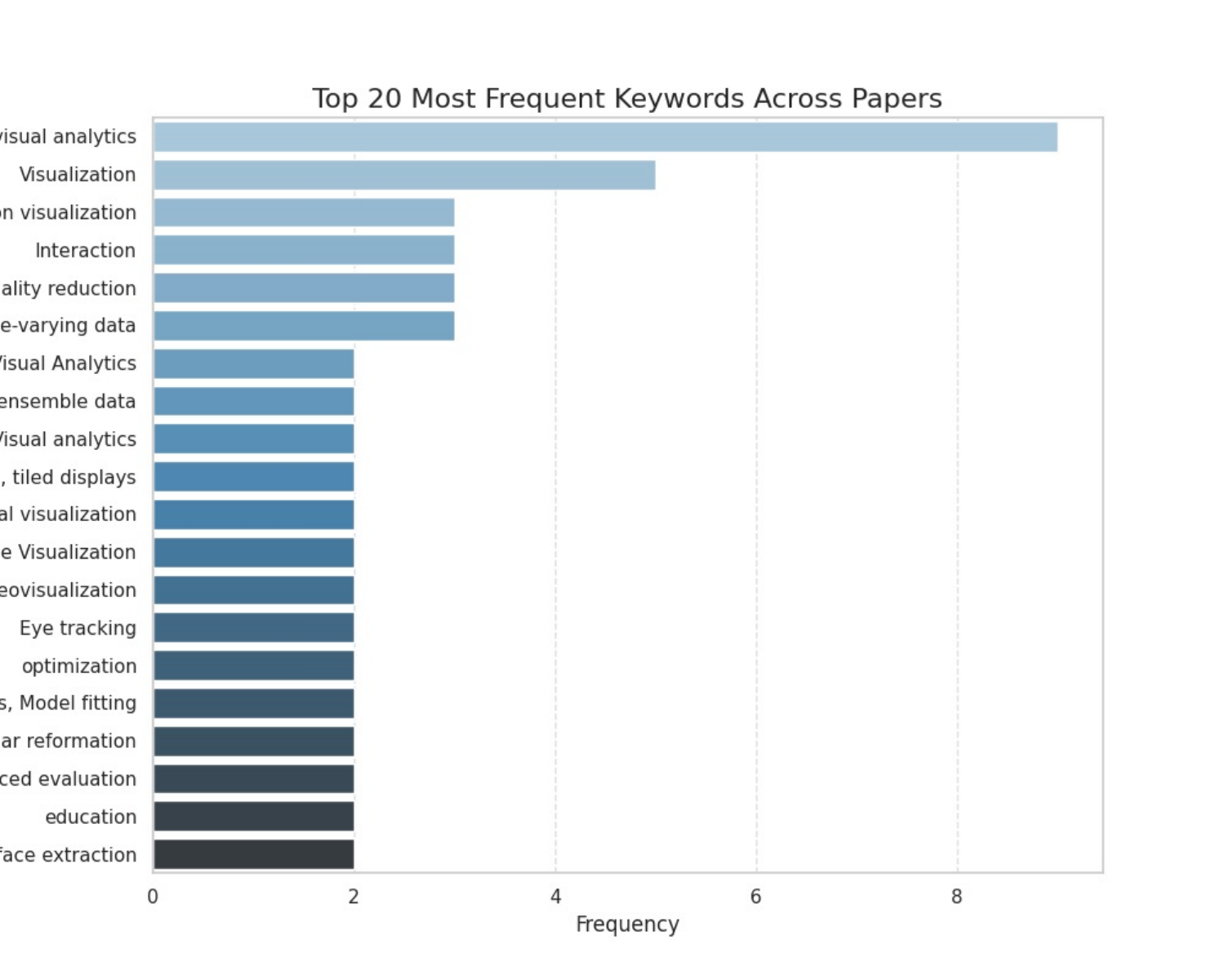}
\vspace{-0.1em}
    \tiny
    Several keywords, such as 'Dimensionality reduction,' 'Time-varying data,' and 'Ensemble data,' have similar frequencies (\~3), indicating that these topics are equally important but distinct areas of inquiry. This suggests a diverse set of challenges and methods being explored in handling large and dynamic datasets. Researchers looking to innovate might explore novel combinations of dimensionality reduction techniques with ensemble data analysis, particularly when dealing with temporal data. Conference organizers or professional associations in the field could plan panel discussions or workshops focusing on these interconnected areas to facilitate knowledge sharing and collaboration among experts.
    
  \end{minipage}\hspace{-0em}
  \begin{minipage}[t]{0.31\linewidth}
    \centering
    \includegraphics[trim=0cm 0cm 0cm 0, clip, 
    width=\linewidth]{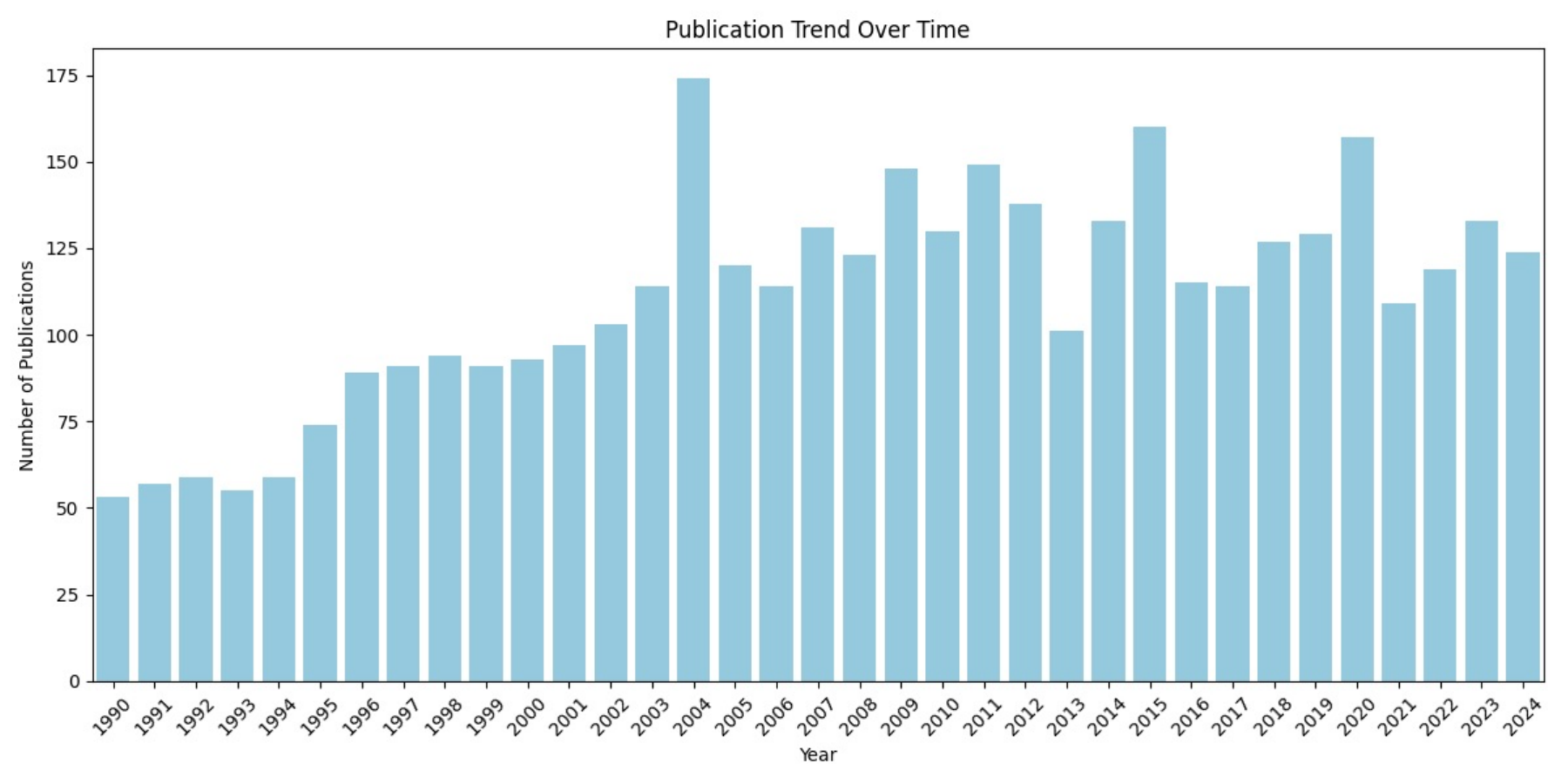}
\vspace{-0.1em}
    \tiny
    There appears to be a cyclical pattern in the publication trend, with periods of growth followed by stabilization or slight decline. For instance, after the initial growth phase ending in 2004, the number of publications shows stability or minor declines in subsequent years. However, there is a resurgence of growth starting around 2015, leading to another peak in 2020. This cycle may reflect periodic cycles in funding, attention, or major discoveries within the field. Understanding these cycles could help stakeholders anticipate periods of increased activity and plan resources accordingly. For example, research institutions might increase their focus on grant proposals and collaboration efforts ahead of expected growth phases, as seen leading up to the peaks in 2004 and 2020.

  \end{minipage}\hspace{-0em}
  \begin{minipage}[t]{0.36\linewidth}
    \centering
    \includegraphics[trim=0cm 0cm 0cm 0, clip, 
    width=\linewidth]{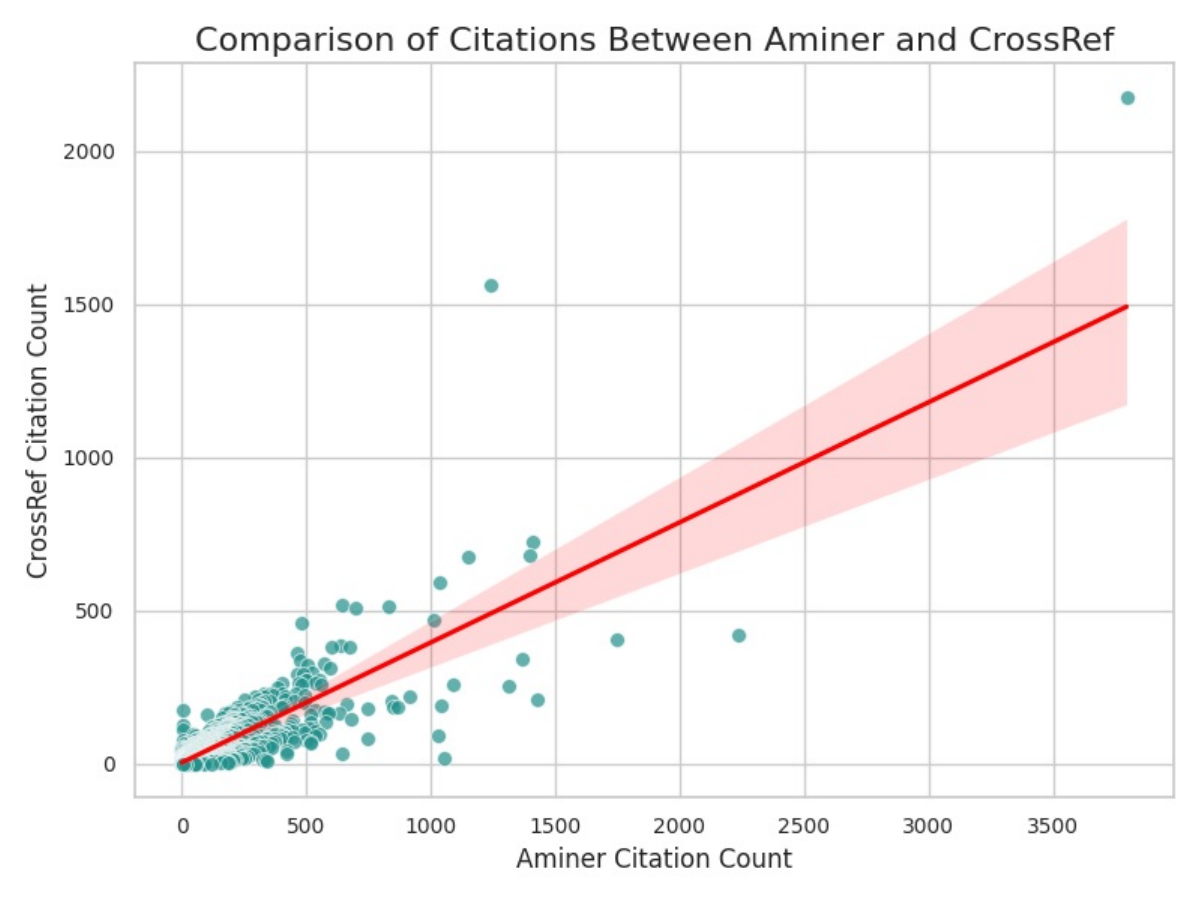}
\vspace{-0.1em}
    \tiny
    The scatter plot reveals a positive correlation between Aminer Citation Count and CrossRef Citation Count, as indicated by the upward-sloping red regression line. This suggests that as the citation count in Aminer increases, there is a corresponding increase in the citation count reported by CrossRef. However, the scatter of data points around the regression line implies variability in this relationship, meaning some papers may have citation counts that deviate significantly from the expected trend. For example, while most points are near the line, there are outliers such as the paper with approximately 3700 Aminer citations but only \~400 CrossRef citations. This could indicate discrepancies in how these platforms count citations or differences in their coverage of certain publication types or fields. Given this variability, actors like researchers or publishers might need to consider both citation databases and cross-reference results to ensure a more accurate understanding of a paper's impact.
    
  \end{minipage}

  \caption{Generated reports for $\rho = 0.6$ at the lowest (32.92), median (58.3), and highest scores (84.2).}
  \label{fig:rho_0.6_all}
\end{figure*}


\subsection{Human Evaluation of Generated Reports}\label{sec:qualitative_human_annotation}
We sample reports with low, medium, and high automatic scores across all pruning ratios and collect pairwise preferences from four annotators.

\paragraph{Qualitative Evaluation}
Within each pruning ratio, higher scoring reports are consistently more focused, evidence grounded, and reliable. For example, at $\rho=0.6$ (Fig.~\ref{fig:rho_0.6_all}), the lowest scoring report contains only surface level keywords, the mid scoring report summarizes overall publication trends, and the highest scoring report explains why certain outliers show different citation counts across two citation systems.

Increasing $\rho$ generally improves overall report quality. Even low scoring reports at $\rho=0.8$ are more coherent and analytical than those at smaller $\rho$ (see Fig.~\ref{fig:low_all} in Appendix \S\ref{app:qualitative}). High scoring reports become similarly informative while remaining diverse in perspective; for instance, $\rho=0.8$ focuses on citation discrepancies and outlier points, whereas $\rho=0.4$ highlights anomalies in paper length (Fig.~\ref{fig:high_all} in Appendix \S\ref{app:qualitative}).

\paragraph{Pairwise Quality Ranking}
Annotators judged high-quality reports to be comparably strong, as reflected by the low variance in preferences and largely indistinguishable ratings reported in Table~\ref{tab:report-correlation} in Appendix~\S\ref{app:qualitative}.
This trend is further illustrated in Fig.~\ref{fig:rankings}, where mean annotator rankings across pruning ratios increasingly converge for the highest-ranked reports, indicating that strong reports become difficult to distinguish as their quality improves.
Overall, these results suggest that SelectiveTTS raises the quality floor while simultaneously reducing variability among top-ranked outputs, leading to more consistent and reliable insights. Additional qualitative examples and discussion are provided in Appendix\S\ref{app:qualitative}.

\section{Related Work}

\paragraph{Test-Time Scaling in LLM Agents.}
Test-Time Scaling (TTS) enhances LLM performance by allocating more inference-time compute~\citep{snell2024scalingllmtesttimecompute, liu20251bllmsurpass405b}.  
Existing approaches are either \emph{sequential}—performing iterative self-refinement or reasoning loops~\citep{madaan2023selfrefineiterativerefinementselffeedback, gou2024criticlargelanguagemodels, zhu2025scalingtesttimecomputellm, muennighoff2025s1simpletesttimescaling}—or \emph{parallel}, sampling multiple candidates for reward-based aggregation (e.g., Best-of-$N$~\citep{sun2024fastbestofndecodingspeculative}, tree search~\citep{yao2023treethoughtsdeliberateproblem}).  
Recent studies extend TTS to multi-agent or interactive settings, redistributing compute across reasoning, planning, and verification modules~\citep{zhu2025scalingtesttimecomputellm, wang2025agentttslargelanguagemodel, yang2025gta1guitesttimescaling, jin2025headsbetteronetesttime}.  
However, these methods mainly emphasize \emph{temporal refinement} within a single reasoning loop.  
Our work instead formulates a \emph{stage-wise} TTS that reallocates compute across agents, mitigating judge error propagation and enabling interpretable scaling under fixed budgets.

\paragraph{LLM Agents for Data Science and Visual Insight Generation.}
LLM-based agents increasingly automate data-centric workflows such as exploratory analysis, code generation, and visualization.  
Early systems like Google’s \emph{Data Science Agent}~\citep{google_dsa_labs_2025}, \emph{DS-Agent}~\citep{guo2024dsagent}, and \emph{DataInterpreter}~\citep{hong2024datainterpreter} demonstrated end-to-end automation but remained largely template-driven and descriptive.  
Later variants extend to domains such as genomics~\citep{liu2024genotex}, ML benchmarking~\citep{liu2025hypobench}, and healthcare~\citep{merrill2024wearable_llm_agents}.  

Recent work further explores \emph{visual insight generation}, where LLMs interpret and contextualize charts~\citep{wang2025chartinsighterapproachmitigatinghallucination, Zhao_2025_LEVA, Zhao_2025_LightVA}, informed by studies on human insight evaluation~\citep{law2020datainsightsprofessionalvisualization, He_2021, LIAN2025100271}.  
However, most systems remain single-pass without inference-time optimization or stage coordination.  
We instead treat this task as a \emph{testbed for selective, stage-wise TTS}, enabling structured multi-agent reasoning and adaptive compute allocation across the pipeline.


\section{Conclusion}
In this work, we reframed unverifiable insight generation as a 
\emph{test-time scaling} problem and proposed \emph{Selective TTS}, a 
process-based, stage-wise pruning strategy guided by lightweight LLM-based evaluators. 
Building on an end-to-end multi-agent pipeline for chart-grounded insight 
generation, we introduced judger selection via human alignment and 
formulated compute usage in terms of LLM calls. Our experiments 
demonstrated that selective pruning under a fixed budget 
progressively reduces variance, improves average final report quality, and 
enables broader exploration across runs without requiring additional 
resources. 

This study closes a gap between generic reasoning agents and 
data-centric decision support, showing how LLM-as-Judgers can provide a principled mechanism for scaling unverifiable reasoning. 
\section{Limitations}
Our study has several limitations.
First, we evaluated only five datasets, which may not fully capture the diversity of real-world data science tasks.
Second, although we explored larger compute budgets and an alternative dataset, the overall compute scale and dataset coverage remain moderate.

Future work could address these limitations in several directions.
In particular, expanding dataset coverage and model diversity would enable a more comprehensive evaluation.
It would also be valuable to explore a wider range of parameter configurations, such as varying the branching factor and pruning ratio, to better understand their effects on selective scaling behavior.
Beyond visual insight generation, an important direction is to extend the proposed Selective TTS framework to other unverifiable generative tasks, such as story generation and video generation, where similar challenges of scalability and evaluation arise.

\paragraph{Potential risks.}
As our framework automates unverifiable insight generation, there remains a risk of producing biased or misleading conclusions if model judgments are misaligned with human reasoning.
We therefore emphasize that such systems should be used to assist, rather than replace, human analysts, with appropriate calibration, transparency, and human oversight.
\section{Ethical Considerations}
The proposed \emph{SelectTTS} framework does not involve the collection of sensitive user data, nor does it rely on any personally identifiable information.  
All datasets used are publicly available and contain only aggregated, non-sensitive tabular records.  
Human annotations were conducted voluntarily by researchers with prior experience in data visualization, without financial incentives, and no personally identifying information was collected.  

The goal of this study is methodological, aiming to understand how inference-time compute allocation affects reasoning quality, rather than to deploy autonomous data science systems.  
Nevertheless, automated insight generation carries potential societal and epistemic risks, including over-reliance on synthetic or unverifiable insights, amplification of bias from training data, and misinterpretation of automatically generated visual explanations.

We encourage responsible deployment by maintaining human oversight in critical decision-making contexts, transparently reporting model provenance and parameters, and discouraging the use of automatically generated insights for policy or medical decision-making without expert verification.  

Finally, all experiments were conducted under controlled research environments, and no additional environmental or privacy risks were introduced beyond standard LLM inference workloads.

\bibliography{custom}

\appendix
\section{Use of LLMs}
We used ChatGPT(chatgpt.com) to generate structured sentences as placeholders then paraphrased and refined in our own words.
\section{Datasets}
\label{app:datasets}
We conduct all experiments on six public datasets designed for visualization-based reasoning and data analysis tasks: \textit{VIS Publication}~\citep{Isenberg:2017:VMC},  \textit{Medical Insurance}~\citep{dataset_insurance}, \emph{Diabetes}~\cite{dataset_diabetes}, 
\emph{Education Performance}~\cite{dataset_education}, 
\emph{Customer Shopping Behavior}~\cite{dataset_shopping}, 
and the \emph{World Happiness Report}~\cite{dataset_happiness}. All datasets are publicly available and released under open research licenses (e.g., CC-BY 4.0). They contain structured tabular data paired with visualizations, enabling evaluation of chart-grounded insight generation.

\paragraph{VIS Publication.}
The VIS Publication dataset is curated from data tables used in published IEEE VIS conference papers. Each sample consists of a structured data table with multiple numerical and categorical attributes, from which candidate visualizations and corresponding descriptive insights can be derived. The dataset covers a diverse range of visualization types (e.g., bar, line, scatter, stacked area) and real-world analytic topics such as demographics, performance metrics, and experimental results.

\paragraph{Medical Insurance.}
The Medical Insurance dataset contains synthetic records generated from an open statistical dataset describing personal attributes (e.g., age, BMI, smoking status, number of children, and insurance charges). It is widely used in data-science education and benchmarking for visualization and regression analysis. 

\paragraph{Diabetes.}
The Diabetes dataset contains clinical measurement records collected from patient examinations. 
Each row includes numerical health indicators such as glucose concentration, blood pressure, 
skin thickness, insulin level, BMI, and pedigree function, alongside a binary attribute indicating 
diabetes status. 
The dataset consists of structured samples combining numerical and binary categorical variables.

\paragraph{Education Performance.}
The Education dataset provides student-level records containing demographic descriptors, study 
behavior variables, and academic performance measurements. 
Attributes include gender, parental education, study time, attendance, and exam scores across 
multiple subjects, represented through both categorical and numerical fields.

\paragraph{Shopping Behavior.}
The Shopping Behavior dataset consists of customer transaction entries from a retail setting. 
Each record includes product category, unit price, purchase quantity, customer segment, and 
transaction date. 
The dataset integrates numerical purchase information with categorical descriptors of customers 
and products.

\paragraph{Happiness.}
The Happiness dataset aggregates annual country-level indicators reported in the World Happiness Report. 
Each entry describes socioeconomic and well-being metrics such as GDP per capita, social support, 
healthy life expectancy, freedom, generosity, and corruption perception. 
The dataset is composed of numerical indicators organized by country and year.

\paragraph{Usage and Licensing.}
Both datasets are used solely for academic research. No personally identifiable information is included in any record. All visualizations and insights used in our experiments were automatically generated by our pipeline or verified through human calibration.
\begin{table*}[!htbp]
\centering
\small
\setlength{\tabcolsep}{6pt}
\begin{tabular}{lcccccc}
\hline
\textbf{Setting} & \textbf{Runs} & \textbf{Total Final Reports} & \textbf{Score (Avg. $\pm$ Std.)} & \textbf{Gen. Budget} & \textbf{Prune Budget} \\
\hline
Baseline ($\rho=0$)   & 15  & 1435 & 61.64 $\pm$ 13.36 & 2567 & 0   \\
$\rho=0.2$            & 22  & 1128 & 63.35 $\pm$ 10.56 & 2240 & 384 \\
$\rho=0.4$            & 40  & 876  & 65.24 $\pm$ 8.81  & 2117 & 439 \\
$\rho=0.6$            & 79  & 578  & \textbf{65.86 $\pm$ 8.16} & 2032 & 522 \\
$\rho=0.8$            & 197 & 197  & 64.97 $\pm$ 8.48  & 1970 & 591 \\
\hline
\end{tabular}
\caption{Main results under different pruning ratios $\rho$ on the VIS dataset.
Selective TTS progressively reduces the number of final reports while improving average quality under a comparable LLM-call budget.}
\label{tab:main-results}
\end{table*}
\section{Additional Results on Medical Insurance Dataset}
\label{app:medical}
\begin{figure}[t]
    \centering
    \centering
    \includegraphics[width=\linewidth]{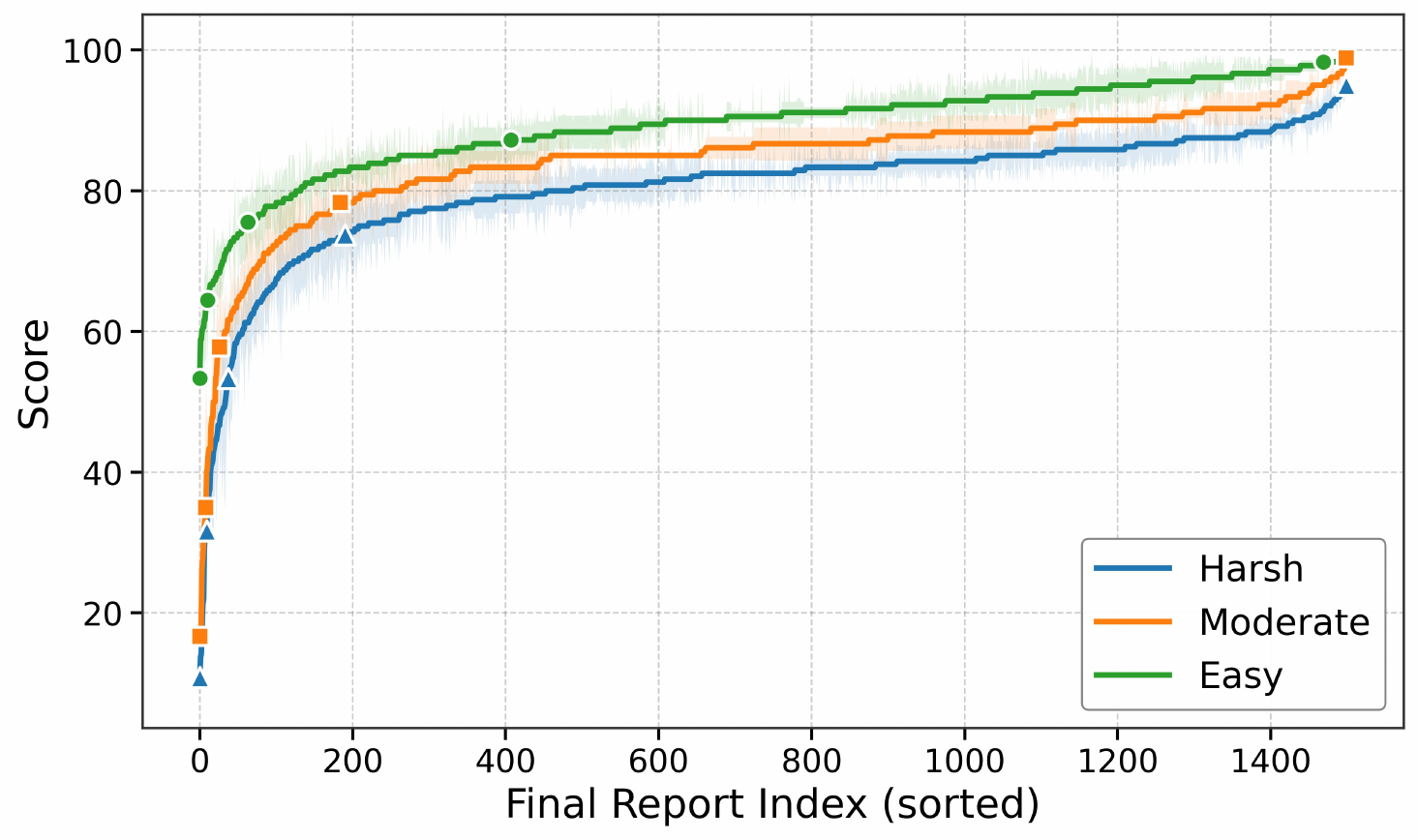}
    \vspace{-5mm}
  \caption{Sorted overall score curves under three judgers (easy, moderate, harsh) on Medical Insurance dataset.
  }
  \label{fig:sorted-curves-insurance}
\end{figure}

\paragraph{Observation from Scaling Graph.}
Fig.~\ref{fig:sorted-curves-insurance} shows the sorted overall score curves on the \textit{Medical Insurance} dataset under the three judgers (\textbf{easy}, \textbf{moderate}, and \textbf{harsh}).  

\section{Human Annotation}
To assess how well the automatic judger aligns with human reasoning, we conducted a structured human annotation study. 
This section outlines the overall evaluation framework, including (1) the rank-based correlation metrics used to quantify alignment, 
(2) the annotation interface and protocol for pairwise comparison, 
(3) the calibration procedures ensuring annotation consistency, 
and (4) illustrative examples comparing human and model preferences.
\subsection{Kendall’s $\tau$, Spearman’s $\rho$, and Kendall’s $W$}
\label{app:alignment-metrics}

To quantify the rank-based alignment between the model-judged ordering $r_J$ of sampled final reports and the consensus human ranking $\hat{r}$, we employ three classical non-parametric statistics: \textbf{Kendall’s $\tau$}, \textbf{Spearman’s $\rho$}, and \textbf{Kendall’s $W$}.  

\paragraph{Kendall’s $\tau$.}
Given two rankings $r_J$ and $\hat{r}$ over $n$ items, Kendall’s $\tau$ evaluates their agreement by comparing all $\tfrac{1}{2}n(n-1)$ item pairs:
\[
\tau_J = \frac{C - D}{\tfrac{1}{2}n(n-1)},
\]
where $C$ and $D$ are the numbers of \textit{concordant} and \textit{discordant} pairs, respectively.  
A pair $(i,j)$ is concordant when the two rankings order $i$ and $j$ in the same direction:
\[
(r_J(i)-r_J(j))(\hat{r}(i)-\hat{r}(j)) > 0.
\]
$\tau_J$ ranges from $-1$ (perfect inversion) to $1$ (perfect agreement), with $0$ indicating no systematic correlation.  
Because it focuses solely on pairwise orderings, $\tau_J$ is relatively insensitive to large but localized rank shifts.

\paragraph{Spearman’s $\rho$.}
Spearman’s rank correlation coefficient instead measures global monotonic agreement by comparing the squared distances between the assigned ranks:
\[
\rho_J = 1 - \frac{6 \sum_{i=1}^{n} (r_J(i) - \hat{r}(i))^2}{n(n^2 - 1)}.
\]
Equivalently, it is the Pearson correlation computed over the rank vectors themselves. Like $\tau_J$, its value lies in $[-1,1]$.  
Compared to Kendall’s $\tau$, Spearman’s $\rho$ penalizes large rank deviations more heavily and therefore captures broader structural differences between the two orderings.

\paragraph{Kendall’s $W$.}
While $\tau$ and $\rho$ describe pairwise agreement between two rankings, Kendall’s coefficient of concordance $W$ measures the \textit{overall} consistency across $m$ annotators ranking the same $n$ items.  
Let $R_{\cdot i}$ denote the sum of ranks assigned to item $i$, and let $\bar{R}$ be their mean. Kendall’s $W$ is defined as:
\[
W = \frac{12 \sum_{i=1}^{n} (R_{\cdot i} - \bar{R})^2}{m^2 (n^3 - n)}.
\]
$W$ ranges from $0$ (no agreement beyond chance) to $1$ (perfect concordance among annotators).  
Higher values indicate that human annotators produce closely aligned rankings, while lower values reveal substantial disagreement.  
In our setting, $W$ establishes the intrinsic consistency of human judgments themselves, contextualizing how well each Judger aligns with human preferences.

\paragraph{Interpretation.}
Kendall’s $\tau$ emphasizes fine-grained pairwise ordering consistency; Spearman’s $\rho$ captures global monotonic structure and penalizes large rank deviations; Kendall’s $W$ summarizes multi-annotator agreement.  
Together, these metrics offer complementary perspectives on how closely a Judger’s ranking $r_J$ aligns with the human consensus $\hat{r}$ and how much inherent variability exists within human annotations.
\begin{figure*}[t]
  \centering
  \begin{subfigure}[t]{0.48\linewidth}
      \centering
      \includegraphics[width=\linewidth]{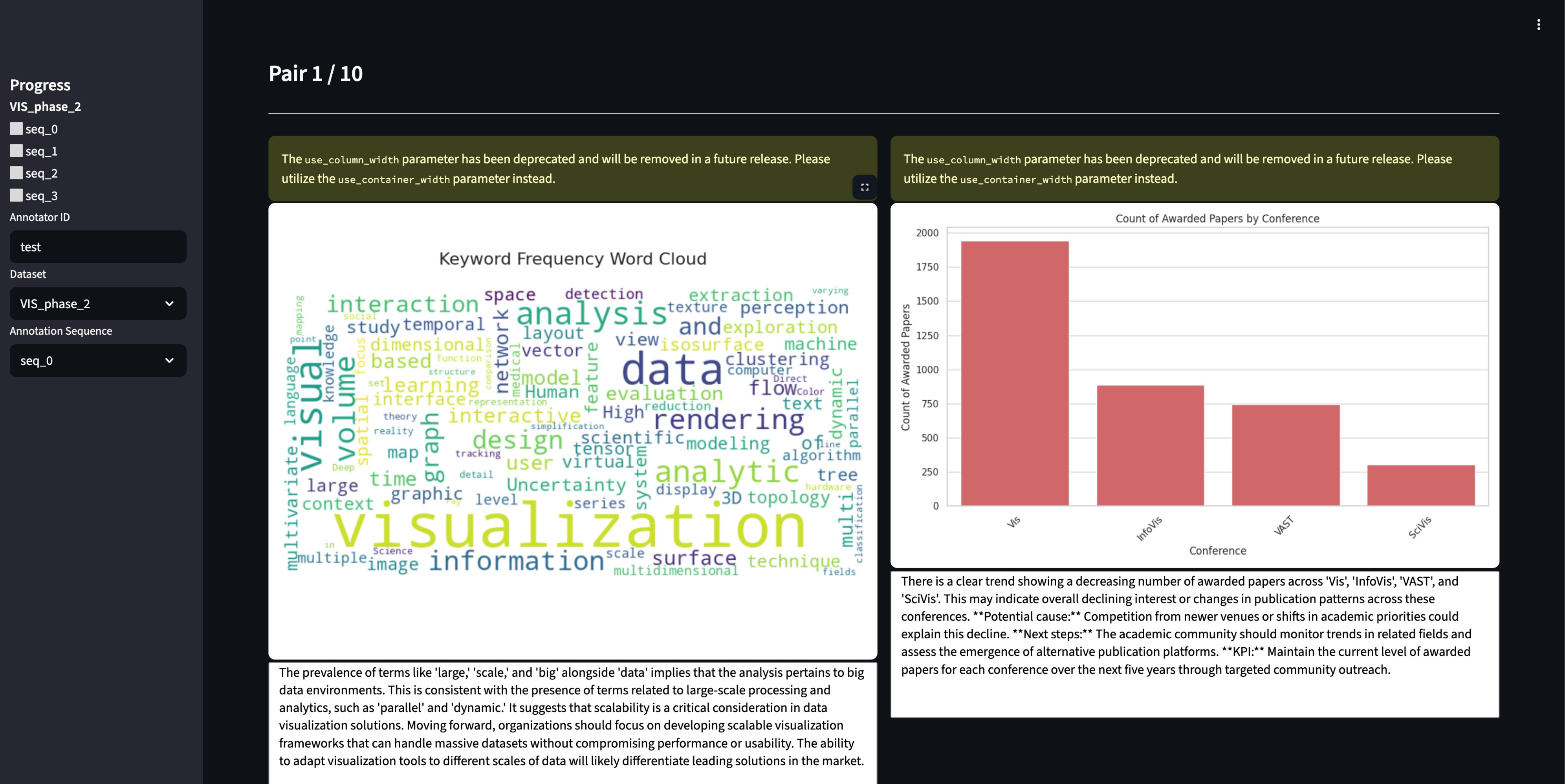}
      \caption{Upper panel of the interface displaying two chart–insight reports for side-by-side comparison.}
      \label{fig:interface_upper}
  \end{subfigure}
  \hfill
  \begin{subfigure}[t]{0.48\linewidth}
      \centering
      \includegraphics[width=\linewidth]{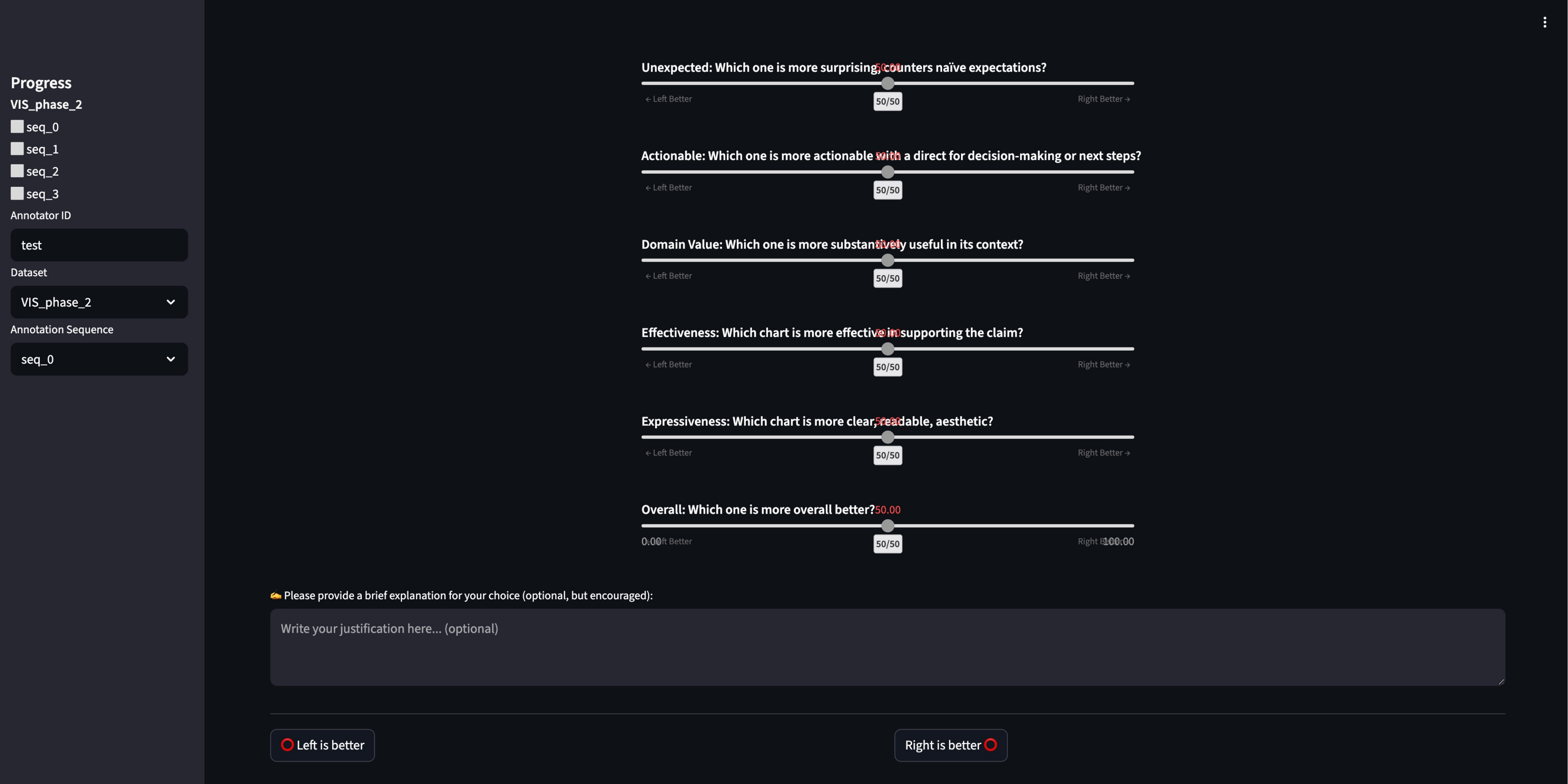}
      \caption{Lower panel showing rubric sliders and overall selection interface.}
      \label{fig:interface_lower}
  \end{subfigure}
  \caption{Human annotation interface used for pairwise comparison of visual–insight reports. 
  The interface consists of an upper comparison panel and a lower evaluation panel.}
  \label{fig:interface}
\end{figure*}
\subsection{Annotation Interface and Protocol}
\label{app:annotation-interface}
Fig.~\ref{fig:interface} illustrates the human annotation interface used in our study.
The interface consists of two main sections: the \textbf{upper panel} (Fig.~\ref{fig:interface_upper}) and the \textbf{lower panel} (Fig.~\ref{fig:interface_lower}).
\paragraph{Interface Layout.}
The left sidebar (shared across both panels) displays annotator metadata, including the \emph{Annotator ID}, the selected \emph{Dataset} (e.g., \texttt{VIS\_phase\_2}), and the corresponding \emph{Annotation Sequence}.
Each annotation session sequentially presents a set of \textbf{paired final reports}, where each pair consists of two chart–insight final reports to be compared.

The \textbf{upper panel} shows the visual comparison interface: two reports (left and right) are displayed side by side, each containing one visualization and its associated insight text.
For a sequence of length $n$, all possible $\binom{n}{2}$ pairs are constructed for pairwise comparison, with both the pair order and the left–right positioning randomized to prevent bias.

The \textbf{lower panel} presents the interactive rating sliders used for detailed comparison along multiple rubric dimensions:
\textit{Trustworthy / Plausible}, \textit{Complex}, \textit{Unexpectedness}, \textit{Actionability}, \textit{Domain Value}, \textit{Effectiveness}, and \textit{Expressiveness}. Note that, for illustration purposes, not all dimensions are shown in the figure due to layout constraints.
Raters are asked to assess which report performs better on each dimension before making an overall decision.
This multi-dimensional evaluation encourages more deliberate reasoning, leading to more consistent and reliable judgments.

\paragraph{Annotation Protocol.}
Annotators are also encouraged to provide a short textual justification (“\emph{Explain your choice}”) to record their reasoning process.
Finally, they indicate the overall better report (\textit{Left is better} or \textit{Right is better}).
Across all $\binom{n}{2}$ comparisons, the number of “wins” for each report is aggregated to obtain a global ranking within the sequence.
All results are automatically logged and synchronized to our central server for aggregation and further analysis.

\subsection{Rater Calibration and Adjudication}
\label{app:rater-calibration}
Before the formal annotation process, all raters underwent a brief \textbf{calibration phase} to ensure a consistent understanding of the task and rubric.
Each rater first reviewed the corresponding dataset and several representative examples to familiarize themselves with the data characteristics.

All annotators were \emph{graduate students with prior experience in data visualization and analytic reasoning}, ensuring familiarity with interpreting charts and evaluating insights.

For each dataset, a designated \textbf{pilot annotation sequence} was jointly reviewed by all raters.
During this session, raters discussed their interpretations of each rubric dimension and aligned their criteria for scoring and pairwise comparison.
This collaborative calibration helped reduce ambiguity and improve inter-rater reliability in the subsequent individual annotations.

After formal annotation, all results were checked for potential anomalies such as inconsistent ranking cycles or missing comparisons.
Whenever such irregularities were detected, the affected rater was asked to re-annotate the corresponding sequence to ensure validity and consistency across the final results.

\subsection{Example of Human–Judger Preference Comparison}
\label{app:comparison}
\begin{figure*}[t]
  \includegraphics[width=\linewidth]{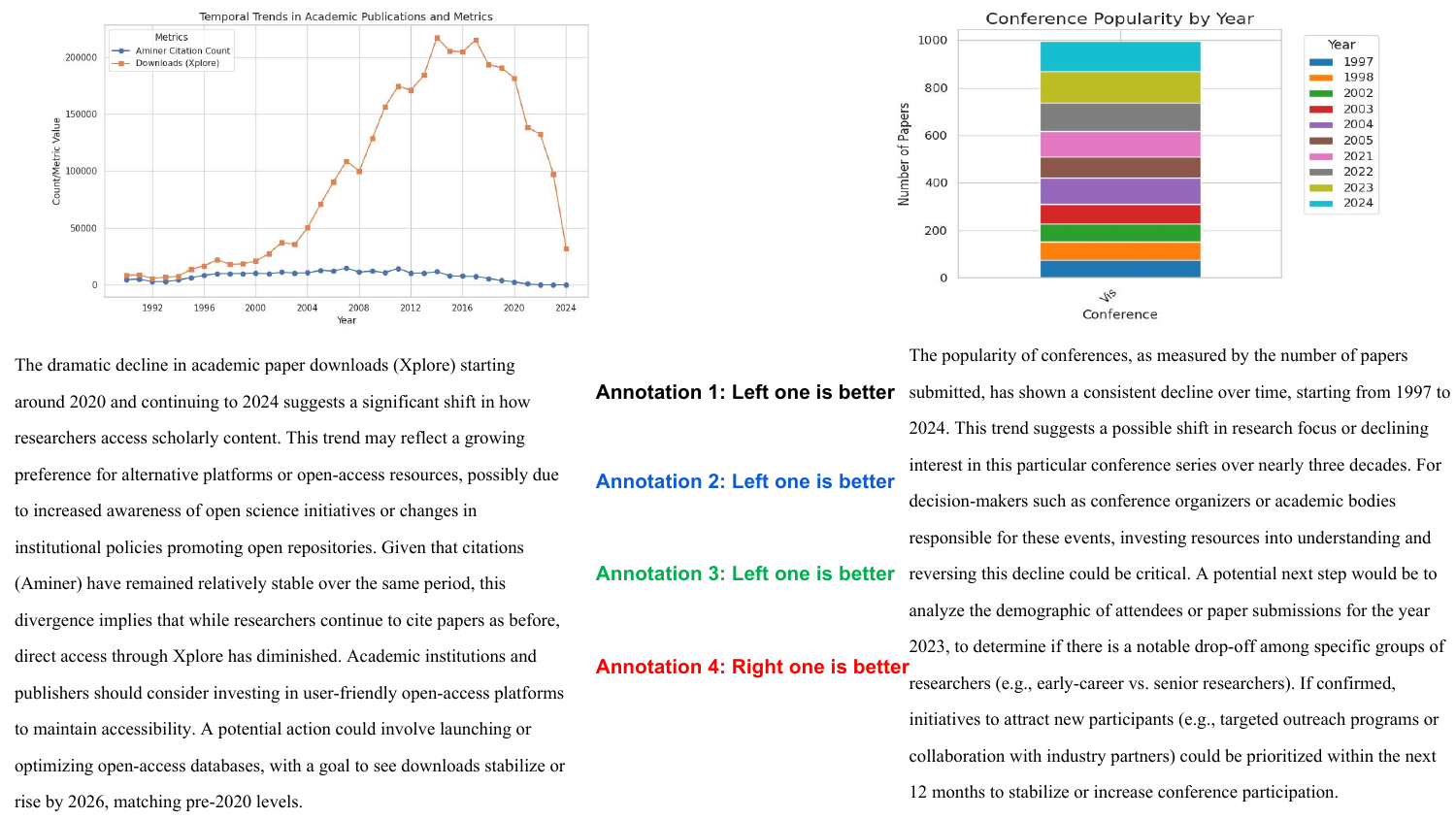}
  \caption{Representative example of a human–judger preference comparison. }
  \label{fig:preference_example}
\end{figure*}

Fig.~\ref{fig:preference_example} illustrates one representative pairwise comparison from the human annotation interface, showing two visual–insight reports displayed side by side along with their corresponding judger scores and human annotator preferences. 
Each annotator independently selected which report they considered better overall, and the aggregated preferences are shown for illustration. 
Note that in the actual annotation process, the judger scores were \emph{not visible} to annotators; they were only added here for post-hoc comparison and visualization purposes.

\section{Supplementary Analysis and Results for Experiments}
This section provides extended analyses and supplementary experiments that complement the main findings presented in the paper. 
We begin by deriving the theoretical compute complexity of a single pipeline run and elucidating how pruning affects the overall budget (\S\ref{app:complexity}). 
We then describe the budget–matching procedure used in our experiments to ensure fair comparisons across pruning ratios~$\rho$ (\S\ref{app:budget}).  
Following these methodological foundations, we report detailed quantitative results for the main experiments (\S\ref{app:main_results}), including run counts, final report statistics, and score distributions under matched compute conditions.  
We further present comprehensive ablation studies (\S\ref{app:ablation}) that examine the contribution of each stage’s pruning strategy and compare Selective~TTS to several simplified baselines.  
Finally, we provide robustness analyses across budget formulations, backbone models, and decoding configurations (\S\ref{app:robustness}), offering a broader evaluation of the consistency and generality of Selective~TTS.
\subsection{Derivation of Budget Complexity Formula}
\label{app:complexity}

We derive the approximate compute complexity of a single pipeline run under pruning ratio~$\rho$.  
For each stage $s \in \{\text{data profiling}, \text{visualization}, \text{insight generation}\}$, 
let $b_s$ denote its branching factor, and let
\[
n_s' = \max\!\big(1, \lceil (1-\rho)b_s \rceil\big)
\]
denote the number of retained candidates after pruning at stage~$s$.  

\paragraph{Exact Budget Decomposition.}
The total compute budget of a single run can be decomposed by enumerating all major LLM calls
performed across stages.  
Specifically, the run-level budget can be written as
\[
\scalebox{0.72}{$
\begin{aligned}
B_{\text{run}}(\rho)
&= b_s + \mathbbm{I}[\rho>0] \\
&\quad + \sum_{i=1}^{n_s'}
\Big(
1 + \mathbbm{I}[\rho>0]
+ 2n_s'
+ |V_i|
+ |V_i|\mathbbm{I}[\rho>0]
+ |V_i|n_s'
\Big).
\end{aligned}
$}
\]
where $|V_i|$ denotes the number of visualization candidates generated from the $i$-th retained
meta-report that pass chart verification.  
By construction, $|V_i| \in \{0,1,\dots,n_s'\}$ and may vary across different branches.

The terms correspond to the following operations:
\begin{itemize}[noitemsep, topsep=2pt, leftmargin=1.5em]
    \item $b_s + \mathbbm{I}[\rho>0]$: metadata report generation ($b_s$ LLM calls), with optional pruning;
    \item $1 + \mathbbm{I}[\rho>0]$: for each metadata report, visualization direction generation (1 call) and optional pruning;
    \item $2n_s'$: for each retained visualization direction, chart code generation (1 call) followed by chart quality verification (1 call);
    \item $|V_i|$: for each verified chart, insight generation (1 call);
    \item $|V_i|\mathbbm{I}[\rho>0]$: optional pruning applied to generated insights;
    \item $|V_i|n_s'$: for each verified chart, $n_s'$ candidate reports are retained and evaluated by the stage-specific judger.
\end{itemize}
\paragraph{Expectation-Based Simplification.}
Since $|V_i|$ varies across branches, we analyze the expected budget.
Let $p_v \in [0,1]$ denote the \emph{chart verification pass probability} for an individual
visualization candidate within a branch (i.e., the probability that a generated chart passes
the verification check and is kept for downstream insight generation).  
Under this formulation, $|V_i|$ denotes the number of verified charts among the $n_s'$ attempted
visualizations generated from the $i$-th retained meta-report.  
Since each chart passes verification independently with probability $p_v$, we model
$|V_i|$ as a binomial random variable with $n_s'$ trials and success probability $p_v$,
which implies $\mathbbm{E}[|V_i|] = p_v n_s'$.
Taking expectation over all branches yields
\[
\scalebox{0.52}{$
\begin{aligned}
\mathbbm{E}\!\left[B_{\text{run}}(\rho)\right]
&= b_s + \mathbbm{I}[\rho>0]
+ \sum_{i=1}^{n_s'}
\mathbbm{E}\!\Big[
1 + \mathbbm{I}[\rho>0] + 2n_s' + |V_i| + |V_i|\mathbbm{I}[\rho>0] + |V_i|n_s'
\Big] \\
&= b_s + \mathbbm{I}[\rho>0]
+ n_s'\Big(
1 + \mathbbm{I}[\rho>0] + 2n_s' + \mathbbm{E}[|V_i|]
+ \mathbbm{E}[|V_i|]\mathbbm{I}[\rho>0] + \mathbbm{E}[|V_i|]n_s'
\Big) \\
&= b_s + \mathbbm{I}[\rho>0]
+ n_s'\Big(
1 + \mathbbm{I}[\rho>0] + 2n_s' + p_v n_s'
+ p_v n_s'\mathbbm{I}[\rho>0] + p_v {n_s'}^{2}
\Big).
\end{aligned}
$}
\]
Expanding and regrouping terms by order, we obtain
\[
\scalebox{0.60}{$
\begin{aligned}
\mathbbm{E}\!\left[B_{\text{run}}(\rho)\right]
&=
\underbrace{\mathcal{O}\!\Big(b_s + n_s' + (2 + p_v) {n_s'}^{2} + p_v {n_s'}^{3}\Big)}_{\text{Expected Generation Budget}}
+
\underbrace{\mathcal{O}\!\Big(\mathbbm{I}[\rho>0]\,(1 + n_s' + p_v {n_s'}^{2})\Big)}_{\text{Expected Pruning Overhead}} .
\end{aligned}
$}
\]

\paragraph{Asymptotic Form.}
Dropping constants and lower-order terms, the dominant budget complexity becomes
\[
\scalebox{0.85}{$
\mathbbm{E}[B_{\text{run}}(\rho)]
\;\approx\;
\mathcal{O}\!\big(p_v n_s'^3\big)
+\mathcal{O}\!\big(\mathbbm{I}[\rho>0]\, p_v n_s'^2\big).
$}
\]

\begin{table*}[!htbp]
\centering
\small
\setlength{\tabcolsep}{6pt}
\begin{tabular}{lccccc}
\hline
\textbf{Setting} & \textbf{Runs} & \textbf{Total Final Reports} & \textbf{Score (Avg. $\pm$ Std.)} & \textbf{Gen. Budget} & \textbf{Prune Budget} \\
\hline
Baseline ($\rho=0$) & 8   & 805 & 61.20 $\pm$ 13.02 & 1435 & 0   \\
$\rho=0.2$          & 12  & 644 & 63.08 $\pm$ 10.78 & 1270 & 218 \\
$\rho=0.4$          & 24  & 498 & 65.52 $\pm$ 8.29  & 1211 & 251 \\
$\rho=0.6$          & 45  & 324 & \textbf{65.70 $\pm$ 8.10} & 1146 & 294 \\
$\rho=0.8$          & 110 & 110 & 65.08 $\pm$ 7.97 & 1100 & 330 \\
\hline
\end{tabular}
\caption{Selective TTS under different pruning ratios $\rho$ on the VIS dataset (small-scale setting).}
\label{tab:ablation-selective}
\end{table*}

\begin{table*}[!htbp]
\centering
\small
\setlength{\tabcolsep}{6pt}
\begin{tabular}{lccccc}
\hline
\textbf{Setting} & \textbf{Runs} & \textbf{Total Final Reports} & \textbf{Score (Avg. $\pm$ Std.)} & \textbf{Gen. Budget} & \textbf{Prune Budget} \\
\hline
Baseline ($\rho=0$) & 8   & 805 & 61.20 $\pm$ 13.02 & 1435 & 0 \\
$\rho=0.2$          & 16  & 696 & 62.05 $\pm$ 12.74 & 1418 & 0 \\
$\rho=0.4$          & 28  & 573 & 60.88 $\pm$ 13.09 & 1443 & 0 \\
$\rho=0.6$          & 60  & 382 & 58.59 $\pm$ 14.49 & 1438 & 0 \\
$\rho=0.8$          & 144 & 144 & 59.75 $\pm$ 14.06 & 1440 & 0 \\
\hline
\end{tabular}
\caption{Random pruning baseline under different pruning ratios $\rho$ (small-scale setting).}
\label{tab:ablation-random}
\end{table*}

\begin{table*}[!htbp]
\centering
\small
\setlength{\tabcolsep}{6pt}
\begin{tabular}{lccccc}
\hline
\textbf{Setting} & \textbf{Runs} & \textbf{Total Final Reports} & \textbf{Score (Avg. $\pm$ Std.)} & \textbf{Gen. Budget} & \textbf{Prune Budget} \\
\hline
Baseline ($\rho=0$) & 8  & 805 & 61.20 $\pm$ 13.02 & 1435 & 0 \\
$\rho=0.2$          & 9  & 608 & 60.34 $\pm$ 14.59 & 1223 & 152 \\
$\rho=0.4$          & 13 & 564 & 60.18 $\pm$ 14.47 & 1323 & 188 \\
$\rho=0.6$          & 12 & 410 & 61.73 $\pm$ 14.12 & 1225 & 205 \\
$\rho=0.8$          & 14 & 244 & 62.60 $\pm$ 12.98 & 1207 & 244 \\
\hline
\end{tabular}
\caption{Insight-only pruning under different pruning ratios $\rho$ (small-scale setting).}
\label{tab:ablation-insight}
\end{table*}

\subsection{Budget Controlling}
\label{app:budget}

For each pruning ratio $\rho$, we would like the total compute used by Selective TTS to be comparable to a baseline system with no pruning.  
Let $B^\ast$ denote the target baseline budget at $\rho = 0$ (either measured in LLM calls or output tokens, depending on the experiment).  
For a fixed $\rho$, let $b_{\rho,i}$ be the budget consumed by the $i$-th run, and let
\[
B_{\rho}^{(n)} \;=\; \sum_{i=1}^{n} b_{\rho,i}.
\]
denote the cumulative budget after $n$ runs.  
Our goal is to choose a number of runs $n_\rho^\star$ such that $B_\rho^{(n_\rho^\star)}$ is as close as possible to $B^\ast$.

\vspace{0.5ex}\noindent
\textbf{Step 1: Initial approximation.}
For each pruning ratio $\rho$, we first obtain a rough estimate of the typical per–run cost $b_{\rho,i}$ from a small number of pilot runs.\footnote{In practice, this can be as simple as running a few preliminary runs and taking their empirical statistics.}  
Based on this estimate, we select an initial number of runs $n_0$ (smaller than the baseline run count) and execute these runs, yielding an initial cumulative budget
\[
B_\rho^{(n_0)} = \sum_{i=1}^{n_0} b_{\rho,i},
\]
which is chosen such that it under-approximates but remains close to the target budget:
\[
B_\rho^{(n_0)} < B^\ast \quad\text{and}\quad B_\rho^{(n_0)} \approx B^\ast.
\]
\vspace{0.5ex}\noindent
\textbf{Step 2: Incremental refinement.}
We initialize the refinement phase with $n = n_0$ and track the remaining budget gap
\[
\Delta^{(n)} = B^\ast - B_\rho^{(n)}.
\]
We then add runs one at a time:
\[
B_\rho^{(n+1)} = B_\rho^{(n)} + b_{\rho,n+1},
\]
continuing this process until we reach the first index $n^+$ for which $B_\rho^{(n^+)} \ge B^\ast$.  
Let $n^- = n^+ - 1$ denote the previous index, for which $B_\rho^{(n^-)} < B^\ast$.
\vspace{0.5ex}\noindent
\textbf{Step 3: Choosing the final run count.}
We then select the run count whose cumulative budget is closest to the target:
\[
n_\rho^\star \;=\; 
\arg\min_{n \in \{n^-,\,n^+\}} \bigl| B_\rho^{(n)} - B^\ast \bigr|.
\]
In other words, we compare $B_\rho^{(n^-)}$ and $B_\rho^{(n^+)}$ and keep whichever is nearer to $B^\ast$.  
This procedure ensures that, for each pruning ratio $\rho$, the total compute $B_\rho^{(n_\rho^\star)}$ is tightly matched to the baseline budget $B^\ast$, while remaining deterministic and reproducible.  
We apply the same procedure both when budgeting in terms of LLM calls and when budgeting in terms of output tokens (cf.\ Fig.~\ref{fig:token-budget}).

\subsection{Detailed Results for Main Experiments}
\label{app:main_results}
Table~\ref{tab:main-results} reports the full quantitative results for Selective~TTS under different pruning ratios~$\rho$ on the VIS dataset.  
These values correspond to the performance trends visualized in Fig.~\ref{fig:rho-score-run}, and include the number of runs executed, the total number of final reports generated, the average score and standard deviation, and the compute budgets allocated to generation and pruning.

As pruning becomes stronger, the number of runs increases while the number of final reports decreases, reflecting the shift from within-run breadth to cross-run exploration under a matched compute budget.  
Mean scores improve steadily from $\rho{=}0$ to $\rho{=}0.6$ (from $61.64$ to $65.86$), accompanied by a substantial reduction in variance (from $13.36$ to $8.16$), indicating both higher quality and greater stability.  
At $\rho{=}0.8$, average scores decline slightly and variance rises, consistent with the risk of over-pruning discussed in the main text.  
\begin{table*}[!htbp]
\centering
\small
\setlength{\tabcolsep}{6pt}
\begin{tabular}{lccccc}
\hline
\textbf{Setting} & \textbf{Runs} & \textbf{Total Final Reports} & \textbf{Score (Avg. $\pm$ Std.)} & \textbf{Gen. Tokens} & \textbf{Prune Tokens} \\
\hline
Baseline ($\rho=0$) & 15 & 1435 & 61.64 $\pm$ 13.36 & 2545971 & 0 \\
$\rho=0.2$          & 24 & 1232 & 63.43 $\pm$ 10.48 & 2357895 & 215323 \\
$\rho=0.4$          & 46 & 1020 & 64.97 $\pm$ 9.01  & 2309491 & 260884 \\
$\rho=0.6$          & 93 & 688  & \textbf{65.97 $\pm$ 8.19}  & 2209659 & 321396 \\
$\rho=0.8$          & 221 & 221 & 65.12 $\pm$ 8.59 & 2190862 & 353504 \\
\hline
\end{tabular}
\caption{Robustness to token-level budgeting: results under matched output-token budgets.}
\label{tab:robust-token}
\end{table*}
\begin{table*}[!htbp]
\centering
\small
\setlength{\tabcolsep}{6pt}
\begin{tabular}{lccccc}
\hline
\textbf{Setting} & \textbf{Runs} & \textbf{Total Final Reports} & \textbf{Score (Avg. $\pm$ Std.)} & \textbf{Gen. Budget} & \textbf{Prune Budget} \\
\hline
\multicolumn{6}{l}{\textbf{Q--G Backbone}} \\
Baseline ($\rho=0$) & 8 & 805 & 61.20 $\pm$ 13.02 & 1435 & 0 \\
$\rho=0.2$          & 12 & 644 & 63.08 $\pm$ 10.78 & 1270 & 218 \\
$\rho=0.4$          & 24 & 498 & 65.52 $\pm$ 8.29  & 1211 & 251 \\
$\rho=0.6$          & 45 & 324 & \textbf{65.70 $\pm$ 8.10}  & 1146 & 294 \\
$\rho=0.8$          & 110 & 110 & 65.08 $\pm$ 7.97 & 1100 & 330 \\
\hline
\multicolumn{6}{l}{\textbf{G--Q Backbone}} \\
Baseline ($\rho=0$) & 8 & 788 & 70.18 $\pm$ 6.87 & 1404 & 0 \\
$\rho=0.2$          & 12 & 586 & 70.83 $\pm$ 6.19 & 1180 & 202 \\
$\rho=0.4$          & 22 & 450 & 71.93 $\pm$ 5.91 & 1144 & 234 \\
$\rho=0.6$          & 47 & 302 & 72.46 $\pm$ 6.64 & 1124 & 286 \\
$\rho=0.8$          & 108 & 108 & \textbf{74.60 $\pm$ 5.76} & 1080 & 324 \\
\hline
\multicolumn{6}{l}{\textbf{Q--Q Backbone}} \\
Baseline ($\rho=0$) & 8 & 650 & 71.53 $\pm$ 6.08 & 1172 & 0 \\
$\rho=0.2$          & 10 & 476 & 73.02 $\pm$ 6.24 & 969  & 165 \\
$\rho=0.4$          & 20 & 375 & 73.25 $\pm$ 5.51 & 957  & 196 \\
$\rho=0.6$          & 38 & 250 & 74.95 $\pm$ 5.78 & 930  & 236 \\
$\rho=0.8$          & 90 & 90  & \textbf{76.06 $\pm$ 5.38} & 900  & 270 \\
\hline
\multicolumn{6}{l}{\textbf{G--G Backbone}} \\
Baseline ($\rho=0$) & 8 & 747 & 59.68 $\pm$ 13.67 & 1344 & 0 \\
$\rho=0.2$          & 11 & 590 & 60.76 $\pm$ 13.74 & 1171 & 201 \\
$\rho=0.4$          & 19 & 467 & 63.60 $\pm$ 10.95 & 1118 & 233 \\
$\rho=0.6$          & 41 & 304 & 65.60 $\pm$ 8.99  & 1072 & 276 \\
$\rho=0.8$          & 103 & 103 & \textbf{66.87 $\pm$ 8.14} & 1030 & 309 \\
\hline
\end{tabular}
\caption{Cross-model robustness: generation--judge backbone combinations (Q--G, G--Q, Q--Q, G--G).}
\label{tab:robust-crossmodel}
\end{table*}

\begin{table*}[!htbp]
\centering
\small
\setlength{\tabcolsep}{6pt}
\begin{tabular}{lccccc}
\hline
\textbf{Setting} & \textbf{Runs} & \textbf{Total Final Reports} & \textbf{Score (Avg. $\pm$ Std.)} & \textbf{Gen. Budget} & \textbf{Prune Budget} \\
\hline
\multicolumn{6}{l}{\textbf{Temperature = 0.5}} \\
Baseline ($\rho=0$) & 8  & 675 & 61.38 $\pm$ 14.10 & 1224 & 0 \\
$\rho=0.2$          & 10 & 500 & 63.29 $\pm$ 12.48 & 990  & 170 \\
$\rho=0.4$          & 18 & 438 & \textbf{65.53 $\pm$ 7.95}  & 1024 & 214 \\
$\rho=0.6$          & 37 & 280 & 65.08 $\pm$ 9.22  & 970  & 250 \\
\hline
\multicolumn{6}{l}{\textbf{Temperature = 0.7}} \\
Baseline ($\rho=0$) & 8  & 690 & 60.71 $\pm$ 13.61 & 1253 & 0 \\
$\rho=0.2$          & 12 & 560 & 64.46 $\pm$ 9.88  & 1120 & 192 \\
$\rho=0.4$          & 19 & 417 & 64.59 $\pm$ 10.10 & 1008 & 209 \\
$\rho=0.6$          & 39 & 282 & \textbf{65.39 $\pm$ 8.82}  & 998  & 256 \\
\hline
\multicolumn{6}{l}{\textbf{Temperature = 1.0 (Original)}} \\
Baseline ($\rho=0$) & 8  & 805 & 61.20 $\pm$ 13.02 & 1435 & 0 \\
$\rho=0.2$          & 12 & 644 & 63.08 $\pm$ 10.78 & 1270 & 218 \\
$\rho=0.4$          & 24 & 498 & 65.52 $\pm$ 8.29  & 1211 & 251 \\
$\rho=0.6$          & 45 & 324 & \textbf{65.70 $\pm$ 8.10}  & 1146 & 294 \\
\hline
\multicolumn{6}{l}{\textbf{Temperature = 1.3}} \\
Baseline ($\rho=0$) & 6  & 87 & 55.14 $\pm$ 18.39 & 212 & 0 \\
$\rho=0.2$          & 8  & 54 & 59.95 $\pm$ 16.33 & 182 & 36 \\
$\rho=0.4$          & 7  & 40 & 62.29 $\pm$ 12.85 & 167 & 39 \\
$\rho=0.6$          & 11 & 28 & \textbf{65.34 $\pm$ 8.62}  & 165 & 45 \\
\hline
\end{tabular}
\caption{Robustness to decoding variation under different temperatures. All results exclude the $\rho=0.8$ setting due to instability and high-error execution at extreme temperatures.}
\label{tab:robust-decoding}
\end{table*}
\begin{table*}[!htbp]
\centering
\small
\setlength{\tabcolsep}{6pt}
\begin{tabular}{lccccc}
\hline
\textbf{Setting} & \textbf{Runs} & \textbf{Total Final Reports} & \textbf{Score (Avg. $\pm$ Std.)} & \textbf{Gen. Budget} & \textbf{Prune Budget} \\
\hline
\multicolumn{6}{l}{\textbf{Medical Insurance Dataset}~\citep{dataset_insurance}} \\
Baseline ($\rho=0$) & 8  & 760 & 62.66 $\pm$ 10.72 & 1293 & 0 \\
$\rho=0.2$          & 10 & 556 & 63.98 $\pm$ 10.76 & 1060 & 184 \\
$\rho=0.4$          & 19 & 450 & 64.71 $\pm$ 9.57  & 1059 & 221 \\
$\rho=0.6$          & 43 & 292 & 67.06 $\pm$ 8.07  & 1028 & 264 \\
$\rho=0.8$          & 99 & 99  & \textbf{69.23 $\pm$ 7.49}  & 990  & 297 \\
\hline
\multicolumn{6}{l}{\textbf{Diabetes Dataset}~\citep{dataset_diabetes}} \\
Baseline ($\rho=0$) & 8  & 725 & 65.74 $\pm$ 10.10 & 1240 & 0 \\
$\rho=0.2$          & 10 & 520 & 66.02 $\pm$ 11.03 & 1015 & 175 \\
$\rho=0.4$          & 20 & 417 & 67.60 $\pm$ 9.19  & 999  & 208 \\
$\rho=0.6$          & 40 & 286 & 68.53 $\pm$ 7.34  & 999  & 257 \\
$\rho=0.8$          & 95 & 95  & \textbf{69.78 $\pm$ 6.99}  & 950  & 285 \\
\hline
\multicolumn{6}{l}{\textbf{Education Performance Dataset}~\citep{dataset_education}} \\
Baseline ($\rho=0$) & 8  & 800 & 62.37 $\pm$ 10.97 & 1352 & 0 \\
$\rho=0.2$          & 11 & 584 & 63.12 $\pm$ 10.18 & 1118 & 194 \\
$\rho=0.4$          & 21 & 474 & 65.24 $\pm$ 8.79  & 1108 & 232 \\
$\rho=0.6$          & 42 & 310 & \textbf{66.89 $\pm$ 7.59}  & 1070 & 276 \\
$\rho=0.8$          & 104 & 104 & 66.69 $\pm$ 8.47 & 1040 & 312 \\
\hline
\multicolumn{6}{l}{\textbf{Customer Shopping Behavior Dataset}~\citep{dataset_shopping}} \\
Baseline ($\rho=0$) & 8  & 805 & 62.37 $\pm$ 11.51 & 1380 & 0 \\
$\rho=0.2$          & 13 & 580 & 63.63 $\pm$ 9.99  & 1141 & 197 \\
$\rho=0.4$          & 21 & 474 & 63.57 $\pm$ 10.75 & 1136 & 236 \\
$\rho=0.6$          & 45 & 308 & 64.30 $\pm$ 10.22 & 1102 & 282 \\
$\rho=0.8$          & 106 & 106 & \textbf{65.38 $\pm$ 9.84} & 1060 & 318 \\
\hline
\multicolumn{6}{l}{\textbf{World Happiness Report Dataset}~\citep{dataset_happiness}} \\
Baseline ($\rho=0$) & 8  & 715 & 65.15 $\pm$ 8.98 & 1283 & 0 \\
$\rho=0.2$          & 10 & 552 & 65.98 $\pm$ 8.46 & 1082 & 186 \\
$\rho=0.4$          & 19 & 429 & 65.22 $\pm$ 9.37 & 1038 & 215 \\
$\rho=0.6$          & 41 & 282 & 67.21 $\pm$ 8.16 & 1023 & 261 \\
$\rho=0.8$          & 99 & 99  & \textbf{70.72 $\pm$ 8.51} & 990  & 297 \\
\hline
\end{tabular}
\caption{Robustness of Selective~TTS across five additional datasets under different pruning ratios $\rho$.}
\label{tab:dataset-generalization}
\end{table*}

\begin{table*}[!htbp]
\centering
\small
\setlength{\tabcolsep}{6pt}
\begin{tabular}{lcccccc}
\hline
\textbf{Setting} & \textbf{Runs} & \textbf{Total Final Reports} & \textbf{Score (Avg. $\pm$ Std.)} & \textbf{Gen. Budget} & \textbf{Prune Budget} & \textbf{Avg. Insight Tokens} \\
\hline
\multicolumn{7}{l}{\textbf{Original Prompt}} \\
Baseline ($\rho=0$) & 8   & 805 & 61.20 $\pm$ 13.02 & 1435 & 0   & 84.50 $\pm$ 19.62 \\
$\rho=0.2$          & 12  & 644 & 63.08 $\pm$ 10.78 & 1270 & 218 & 85.74 $\pm$ 19.09 \\
$\rho=0.4$          & 24  & 498 & 65.52 $\pm$ 8.29  & 1211 & 251 & 88.33 $\pm$ 20.15 \\
$\rho=0.6$          & 45  & 324 & \textbf{65.70 $\pm$ 8.10}  & 1146 & 294 & 91.78 $\pm$ 21.01 \\
$\rho=0.8$          & 110 & 110 & 65.08 $\pm$ 7.97  & 1100 & 330 & 94.24 $\pm$ 20.54 \\
\hline
\multicolumn{7}{l}{\textbf{$\sim$130 tokens}} \\
Baseline ($\rho=0$) & 8   & 795 & 65.80 $\pm$ 13.34 & 1401 & 0   & 129.72 $\pm$ 23.08 \\
$\rho=0.2$          & 12  & 608 & 69.74 $\pm$ 8.54  & 1216 & 208 & 134.56 $\pm$ 23.61 \\
$\rho=0.4$          & 20  & 480 & 69.58 $\pm$ 8.81  & 1146 & 238 & 135.11 $\pm$ 24.63 \\
$\rho=0.6$          & 43  & 322 & 71.35 $\pm$ 6.93  & 1118 & 288 & 139.49 $\pm$ 25.58 \\
$\rho=0.8$          & 108 & 108 & \textbf{71.77 $\pm$ 6.73}  & 1080 & 324 & 147.11 $\pm$ 26.55 \\
\hline
\multicolumn{7}{l}{\textbf{$\sim$180 tokens}} \\
Baseline ($\rho=0$) & 8   & 740 & 71.15 $\pm$ 10.51 & 1313 & 0   & 177.74 $\pm$ 35.48 \\
$\rho=0.2$          & 12  & 552 & 72.23 $\pm$ 10.06 & 1110 & 190 & 174.76 $\pm$ 35.72 \\
$\rho=0.4$          & 19  & 465 & 73.81 $\pm$ 7.24  & 1093 & 228 & 176.82 $\pm$ 35.02 \\
$\rho=0.6$          & 41  & 300 & 74.50 $\pm$ 6.69  & 1055 & 271 & 191.45 $\pm$ 40.93 \\
$\rho=0.8$          & 101 & 101 & \textbf{74.83 $\pm$ 7.11}  & 1010 & 303 & 196.65 $\pm$ 37.53 \\
\hline
\multicolumn{7}{l}{\textbf{$\sim$220 tokens}} \\
Baseline ($\rho=0$) & 10  & 790 & 74.06 $\pm$ 10.44 & 1416 & 0   & 222.87 $\pm$ 51.68 \\
$\rho=0.2$          & 15  & 604 & 74.97 $\pm$ 8.78  & 1226 & 210 & 220.49 $\pm$ 54.22 \\
$\rho=0.4$          & 25  & 472 & 76.84 $\pm$ 6.39  & 1154 & 240 & 236.39 $\pm$ 62.93 \\
$\rho=0.6$          & 47  & 314 & 77.41 $\pm$ 5.59  & 1126 & 288 & 242.31 $\pm$ 58.91 \\
$\rho=0.8$          & 109 & 109 & \textbf{78.96 $\pm$ 5.77}  & 1090 & 327 & 245.32 $\pm$ 55.07 \\
\hline
\end{tabular}
\caption{Robustness to output length variation under different insight-generation constraints.}
\label{tab:robust-length}
\end{table*}

\subsection{Detailed Results for Ablation Study}
\label{app:ablation}

Fig.~\ref{fig:ablation} visualizes the comparative behavior of Selective~TTS, Random Pruning,
and Insight-only Pruning across pruning ratios~$\rho$.  
Tables~\ref{tab:ablation-selective}, \ref{tab:ablation-random}, and \ref{tab:ablation-insight}
provide the corresponding numerical results, including the number of executed runs, total final
reports, average scores, and compute budgets.

Across all pruning ratios, Selective~TTS consistently achieves higher mean scores and lower
variance than both ablation baselines, even under the small-scale setting.  
Random Pruning shows no clear improvement trend and exhibits substantially higher variance,
indicating that unguided reduction of candidates is ineffective for quality preservation.
Insight-only Pruning performs slightly better than Random Pruning but remains significantly
weaker than Selective~TTS, demonstrating that pruning applied only at the final stage is
insufficient to realize the benefits of structured test-time selection.
\begin{figure*}[t]
  \centering
  \includegraphics[width=0.32\linewidth]{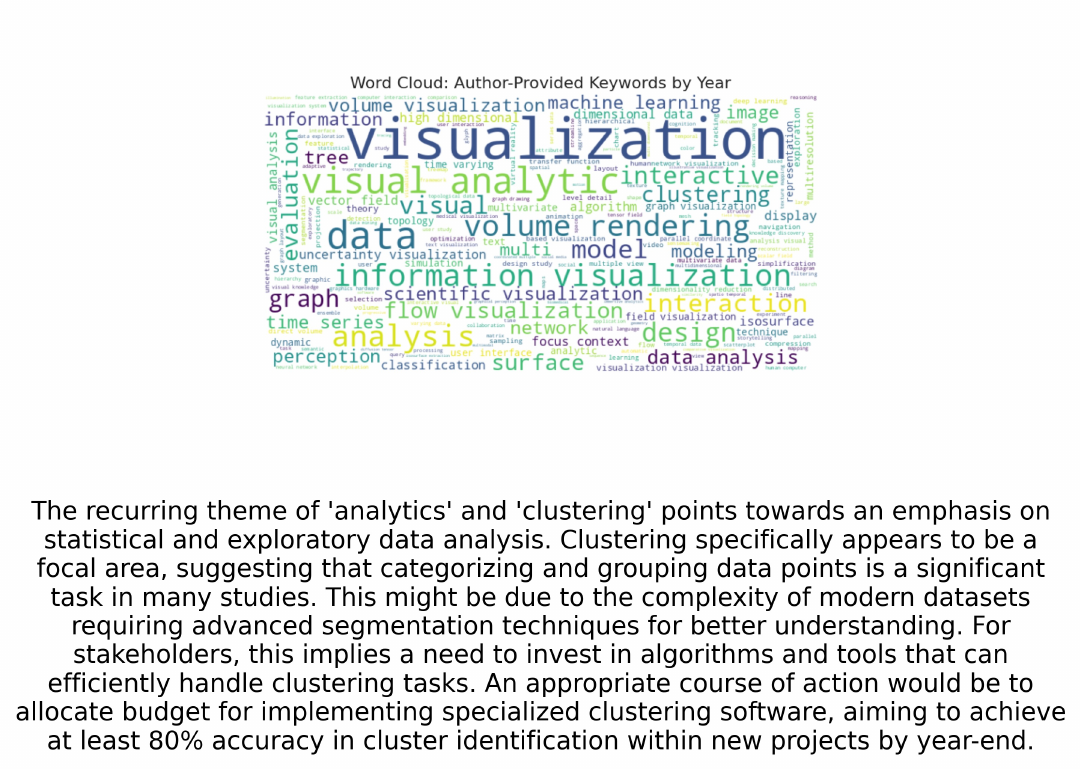}\hfill
  \includegraphics[width=0.32\linewidth]{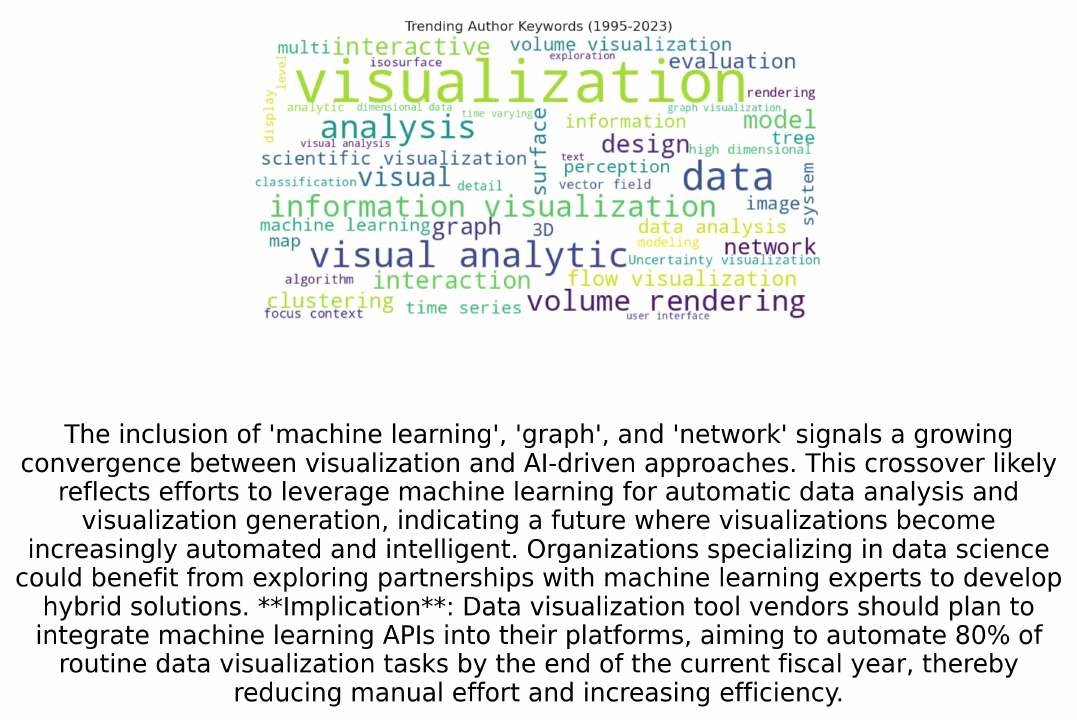}\hfill
  \includegraphics[width=0.32\linewidth]{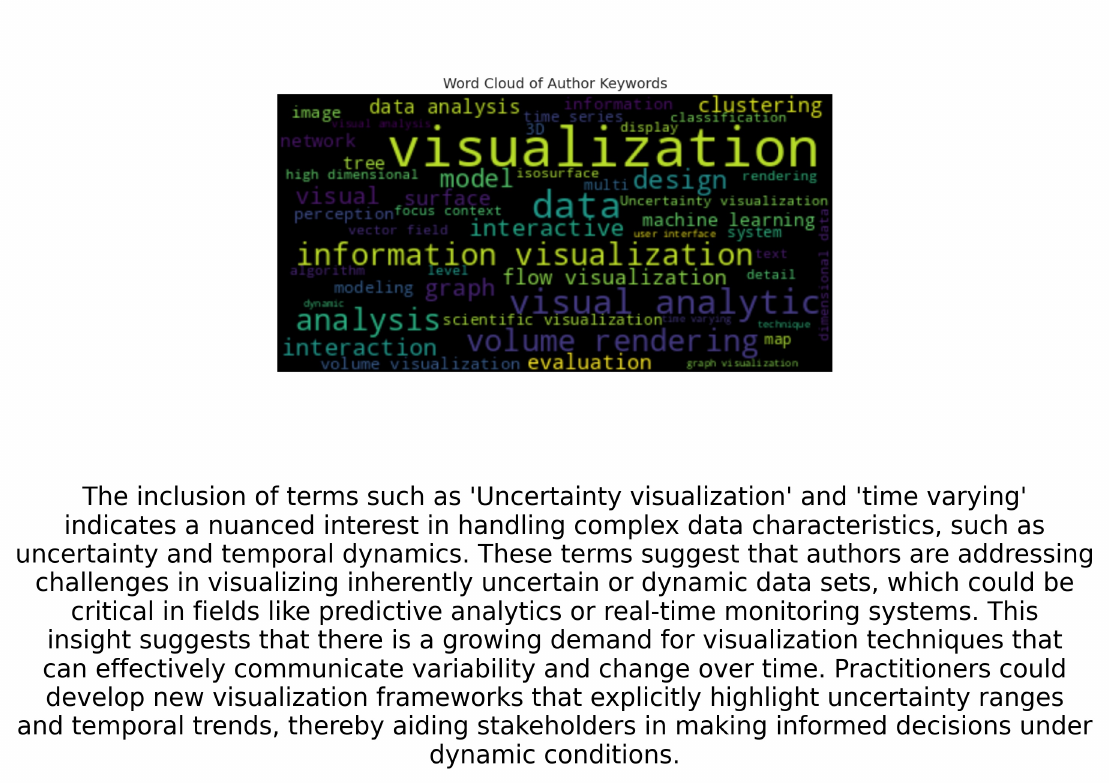}
    \includegraphics[width=0.32\linewidth]{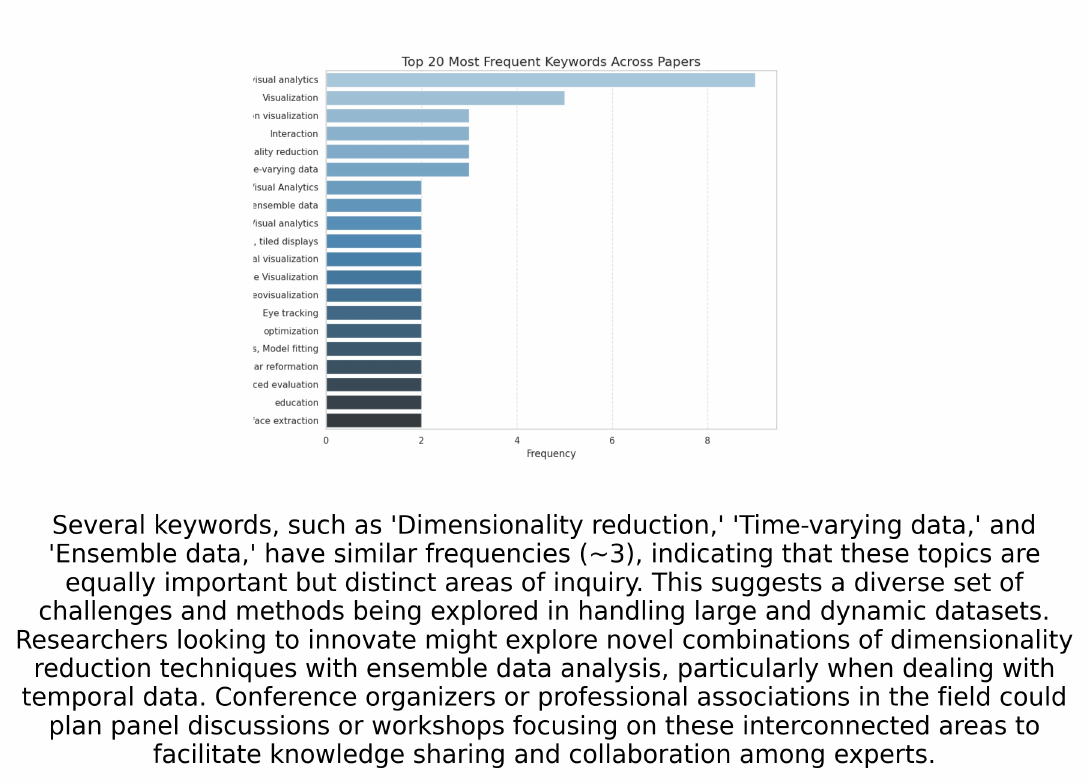}
      \includegraphics[width=0.32\linewidth]{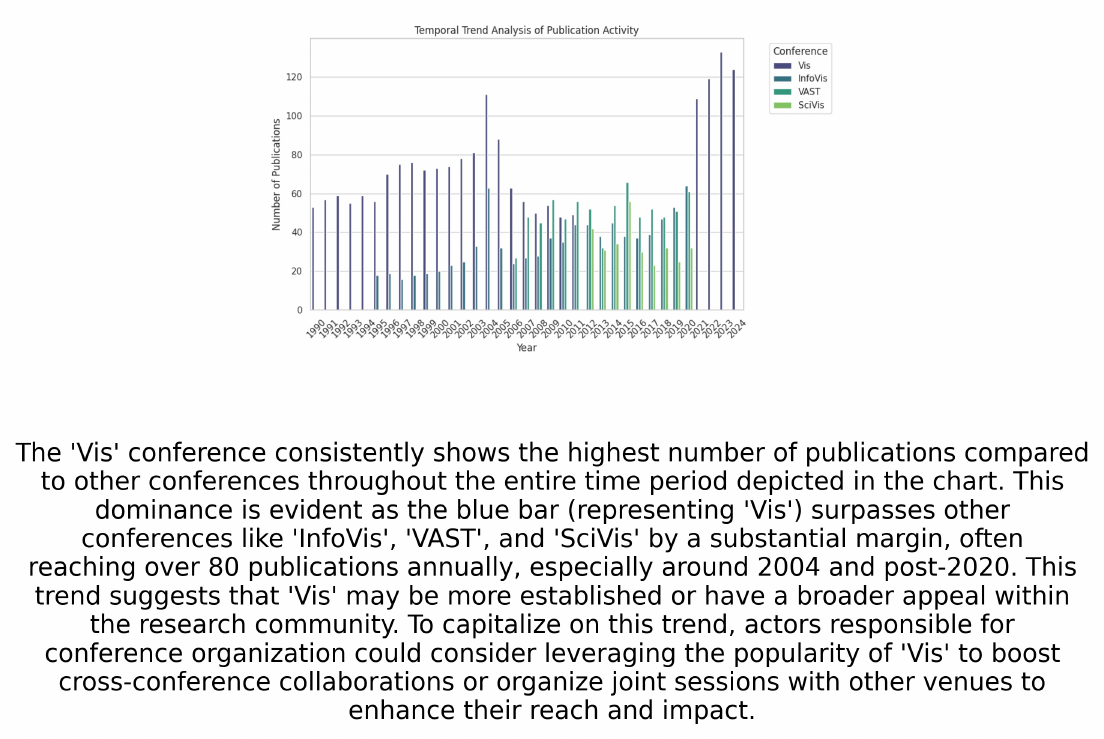}
  \caption{Low-scoring reports across all pruning ratios. The corresponding scores are: $\rho$, with scores of $\rho=0$: 11.3, $\rho=0.2$: 20, $\rho=0.4$: 11.3, $\rho=0.6$: 32.9, $\rho=0.8$: 47.5.}
  \label{fig:low_all}
\end{figure*}

\begin{figure*}[t]
  \centering
  \includegraphics[width=0.32\linewidth]{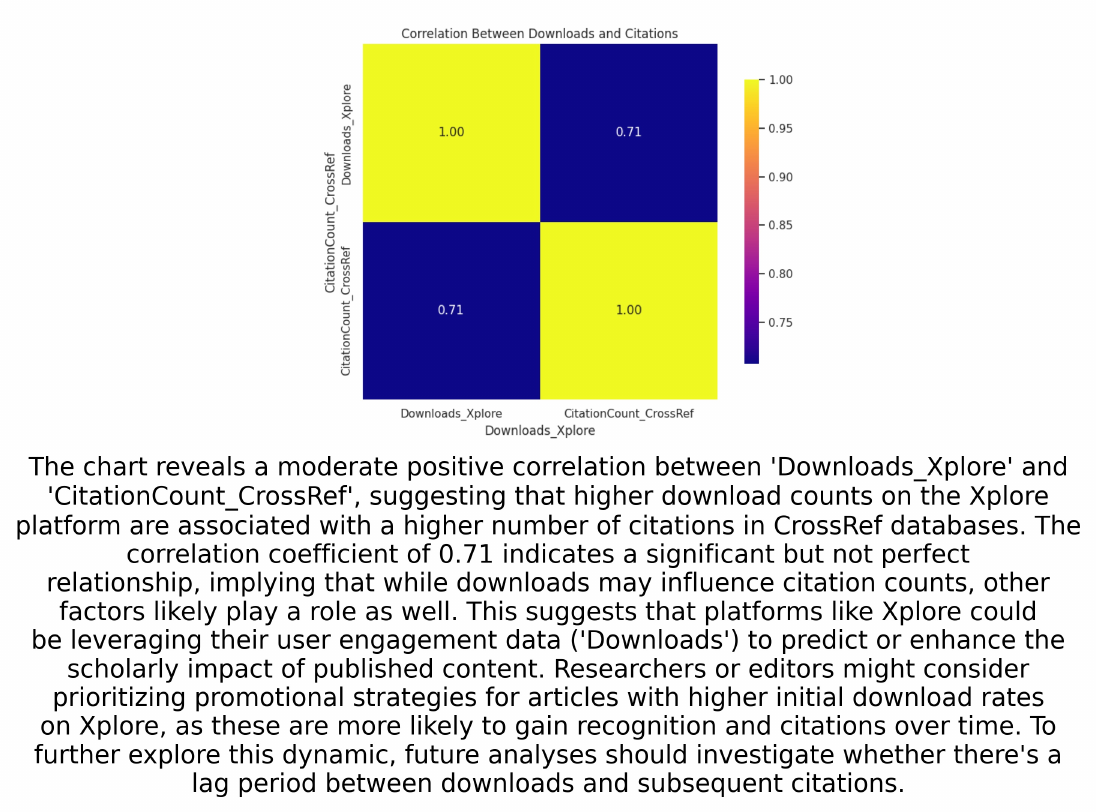}\hfill
  \includegraphics[width=0.32\linewidth]{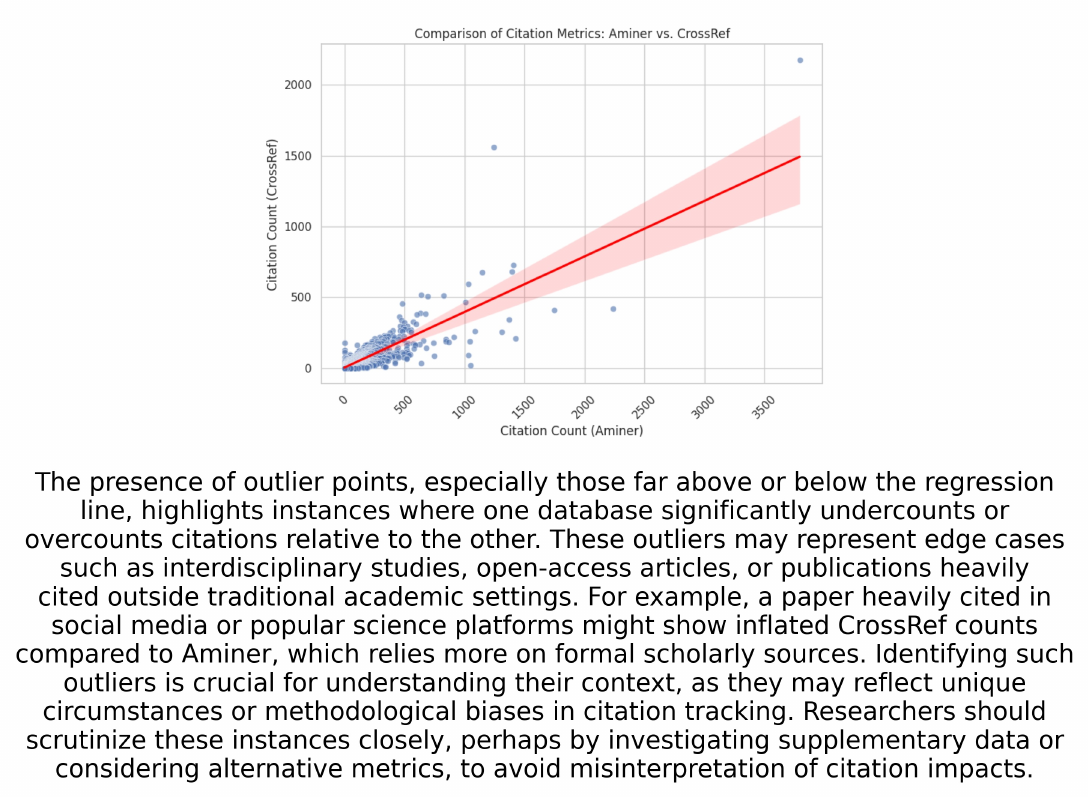}\hfill
  \includegraphics[width=0.32\linewidth]{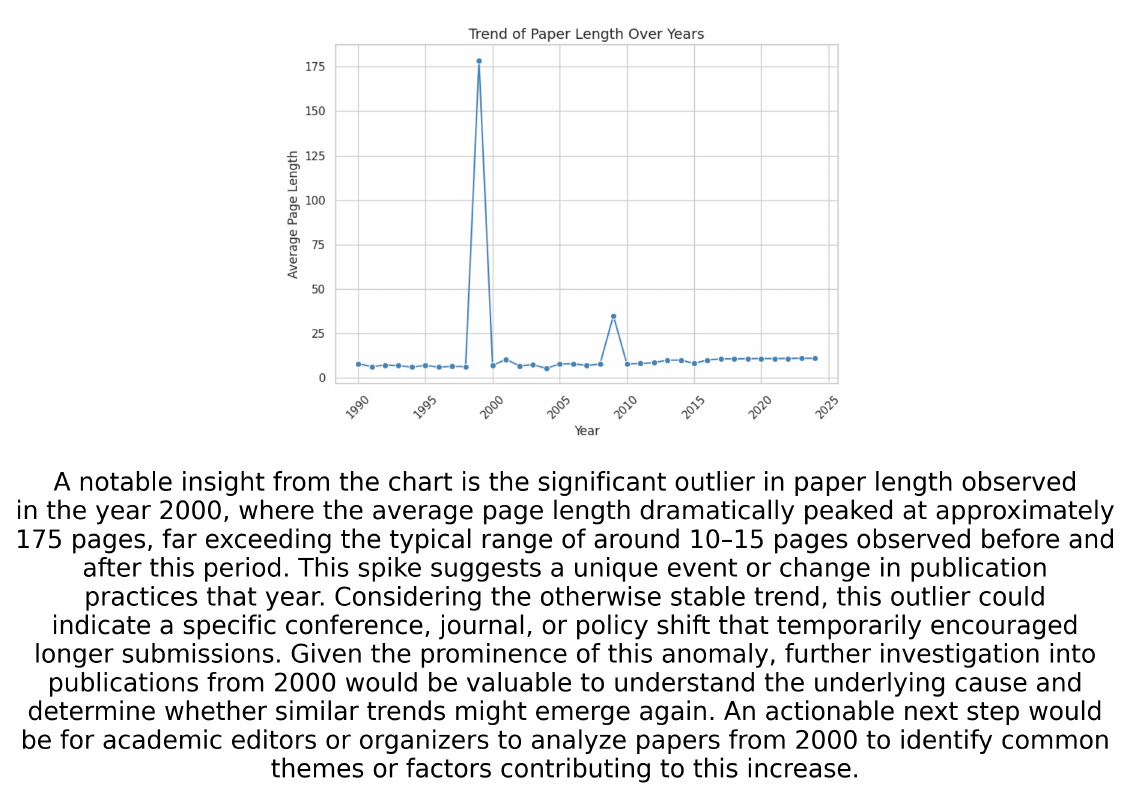}
    \includegraphics[width=0.32\linewidth]{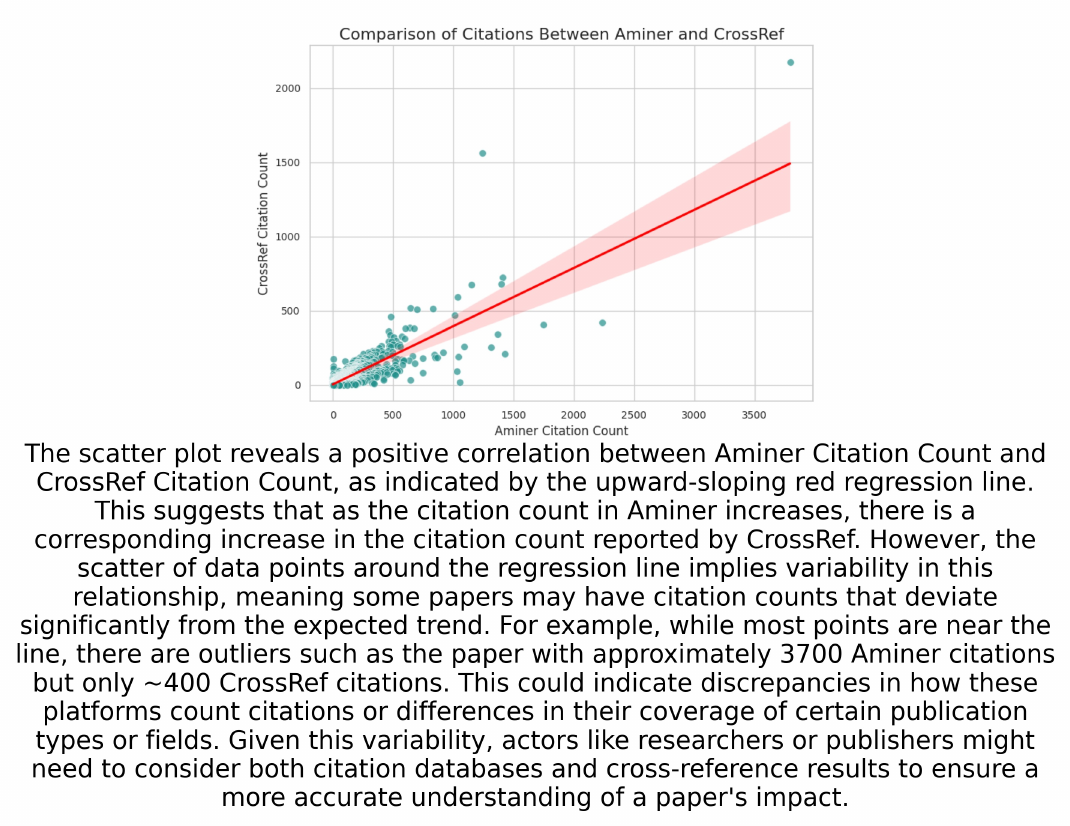}
      \includegraphics[width=0.32\linewidth]{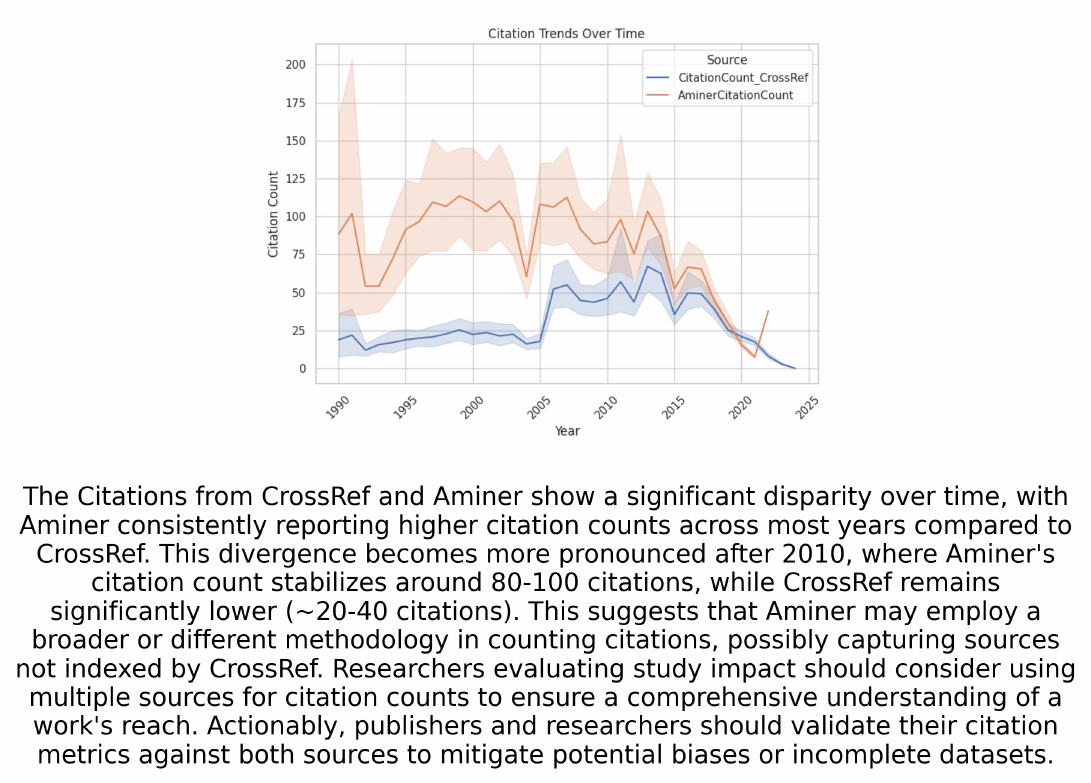}
  \caption{High-scoring reports across all pruning ratios. The corresponding scores are: $\rho=0$: 85.4, $\rho=0.2$: 85, $\rho=0.4$: 84.6, $\rho=0.6$: 84.2, $\rho=0.8$: 84.2.}
  \label{fig:high_all}
\end{figure*}

\subsection{Detailed Results for Robustness Analyses}
\label{app:robustness}

This section provides the full numerical results corresponding to the robustness studies reported 
in \S\ref{sec:selective_reliability}.  
Across three dimensions—budget formulation, including model backbone, and decoding configuration, we 
observe that Selective~TTS produces highly consistent scaling trends, supporting the reliability 
and generality of the method.

\paragraph{Token-Level vs.\ LLM-Call Budget.}
Fig.~\ref{fig:token-budget} visualizes the score and variance trends when the total compute is 
controlled using output tokens rather than LLM calls.  
Table~\ref{tab:robust-token} reports the detailed statistics, including the number of runs, 
final reports, and token-level budgets.  
Across all pruning ratios~$\rho$, the performance curves under token budgeting closely match 
those obtained under LLM-call budgeting: mean scores rise from $\rho=0$ to $\rho=0.6$, before 
slightly declining at $\rho=0.8$, and variance follows the same decreasing-then-increasing 
pattern.  
The number of runs and total generated tokens also exhibit similar trends.  
These observations confirm that LLM calls provide a faithful and stable proxy for 
token-level compute, while additionally offering easier control and higher reproducibility in 
practice.

\paragraph{Cross-Model Variation.}
To examine whether scaling behavior depends on the underlying model family, we evaluate 
Selective~TTS under four generator–judger backbone combinations: Q--G, G--Q, Q--Q, and G--G.  
The grouped-bar comparison in Fig.~\ref{fig:backbone} shows that all four backbones exhibit 
consistent improvement as $\rho$ increases, though the absolute score levels differ across 
models.  
Table~\ref{tab:robust-crossmodel} provides the full numerical breakdown for each pruning ratio.  
The similar upward trends across model families indicate that Selective~TTS does not rely on 
a specific generator or judger, and that no strong family-specific bias is driving the 
performance gains.

\paragraph{Sensitivity to Decoding Parameters.}
We further test robustness to decoding diversity by varying the generation temperature while 
keeping top-$p$ fixed.  
Fig.~\ref{fig:temperature} summarizes the score improvements across temperatures, and 
Table~\ref{tab:robust-decoding} lists the detailed results for each $\rho \in \{0, 0.2, 0.4, 0.6\}$.  
Across temperatures $0.5$, $0.7$, $1.0$, and $1.3$, Selective~TTS consistently improves over 
the baseline, with higher pruning ratios yielding larger gains.  
While extremely high temperatures introduce instability and reduce the number of usable runs 
(as reflected in Appendix tables), the overall scaling pattern remains stable: Selective~TTS 
continues to allocate compute effectively even as output diversity shifts.  
These results demonstrate that Selective~TTS is robust to decoding variability and does not 
depend on a narrow set of generation parameters.
\paragraph{Generalization Across Datasets.}
We further assess robustness by evaluating Selective~TTS on six additional tabular datasets 
(Insurance, Diabetes, Education, Shopping, Happiness, and Student).  
Fig.~\ref{fig:generalization} visualizes the score trends across pruning ratios, and 
Table~\ref{tab:dataset-generalization} reports the full numeric results.

Across all datasets, Selective~TTS consistently improves performance as $\rho$ increases from 
$0$ to $0.6$, mirroring the behavior observed on the VIS dataset.  
The aggressive setting ($\rho=0.8$) produces dataset-dependent outcomes: several datasets 
continue to improve, while others show weaker gains, suggesting that heavy pruning may remove 
useful branches when data are noisier or structurally complex.  
Overall, these results indicate that Selective~TTS generalizes reliably across diverse data 
domains, with moderate pruning ratios offering the most stable improvements.
\paragraph{Sensitivity to Insight Length.}
Fig.~\ref{fig:token_distribution} reveals a clear association between insight length and final evaluation score under the baseline setting ($\rho=0$), motivating an investigation into whether Selective~TTS remains effective when the length of generated insights varies.

Specifically, we control the average insight length through prompt-level sentence constraints rather
than explicit token limits (e.g., requiring at least six complete sentences with fully developed
reasoning), and group the resulting outputs into coarse length categories.
Table~\ref{tab:robust-length} reports the detailed numerical results for each regime, while
Fig.~\ref{fig:robust_tokens} summarizes the corresponding performance trends.
Across all insight-length settings, Selective~TTS consistently improves performance as the pruning
ratio $\rho$ increases, and the relative scaling trends remain stable.
Although higher pruning ratios tend to select slightly longer insights on average, this effect alone
does not explain the observed gains.
First, as shown in \S\ref{sec:ablation}, pruning applied only at the insight stage fails to
achieve comparable or stable improvements, indicating that length-based selection at the final stage
is insufficient.
Second, within each length regime, the variance in average insight length across
$\rho \in \{0, 0.4\}$ is relatively small (e.g., $\sim$4 tokens for the $\sim$85-token setting,
$\sim$5 tokens for $\sim$130 tokens, $\sim$1 token for $\sim$180 tokens, and $\sim$13 tokens for
$\sim$220 tokens), yet meaningful performance improvements are still observed.
\begin{table}[t]
\centering
\small
\resizebox{\columnwidth}{!}{
\begin{tabular}{l|ccc}
\hline
\textbf{Reports} & Kendall's $\tau$ ($\uparrow$) & Spearman's $\rho$ ($\uparrow$) & Kendall's W \\
\hline
Low & 0.70$\pm$0.10 & 0.82$\pm$0.08 & 0.7875 \\
Medium & 0.30$\pm$0.41 & 0.38$\pm$0.45 & 0.2750 \\
High & -0.15$\pm$0.17 & -0.17$\pm$0.19 & 0.1250 \\
\hline
\end{tabular}
}
\caption{Alignment metrics between the harsh judger and human ratings for reports sampled from \textit{low}, \textit{medium}, and \textit{high} scoring across all pruning ratios. The alignment decreases in high-quality reports, as the differences become subtle and indistinguishable and annotators rated these top reports comparably strong.}
\label{tab:report-correlation}
\end{table}
\begin{figure*}[!htbp]
  \centering
  \includegraphics[width=0.32\linewidth]{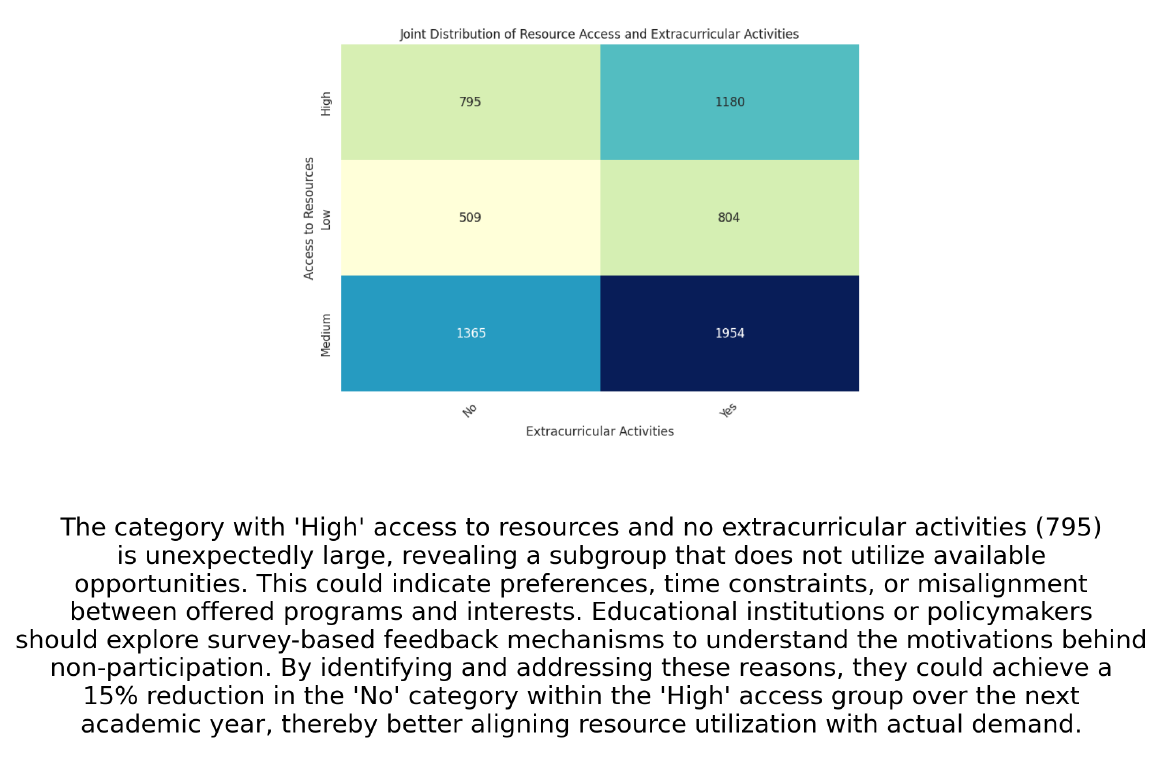}\hfill
  \includegraphics[width=0.32\linewidth]{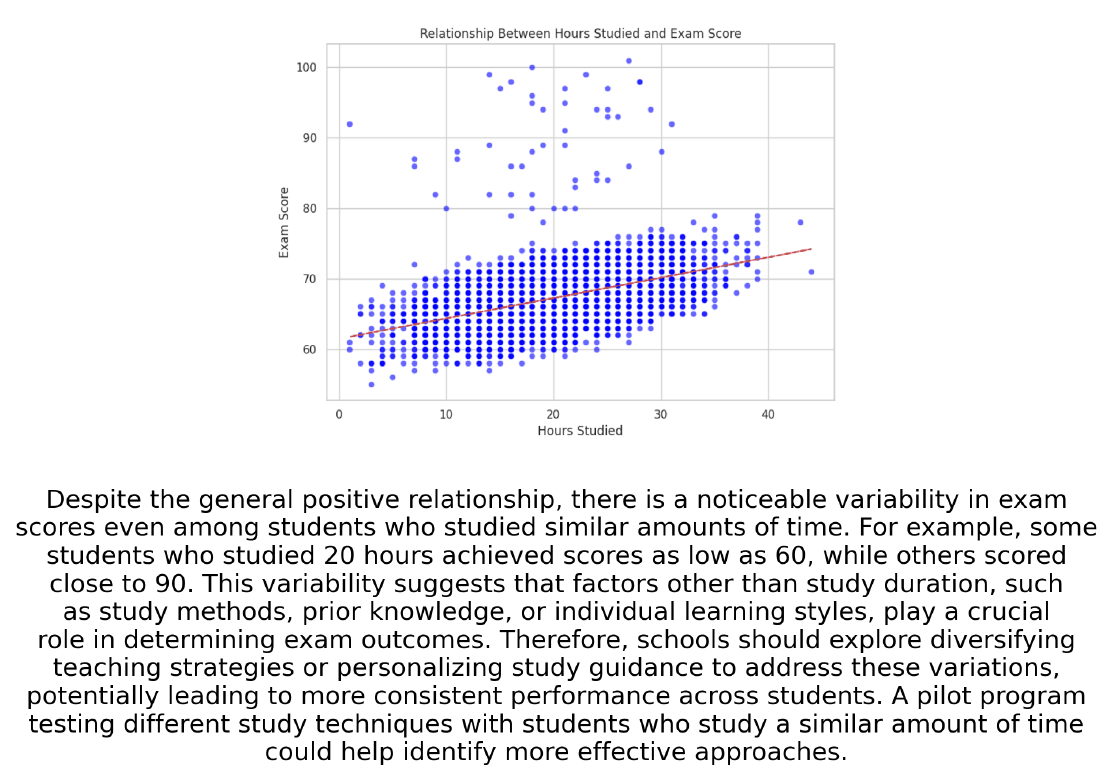}\hfill
  \includegraphics[width=0.32\linewidth]{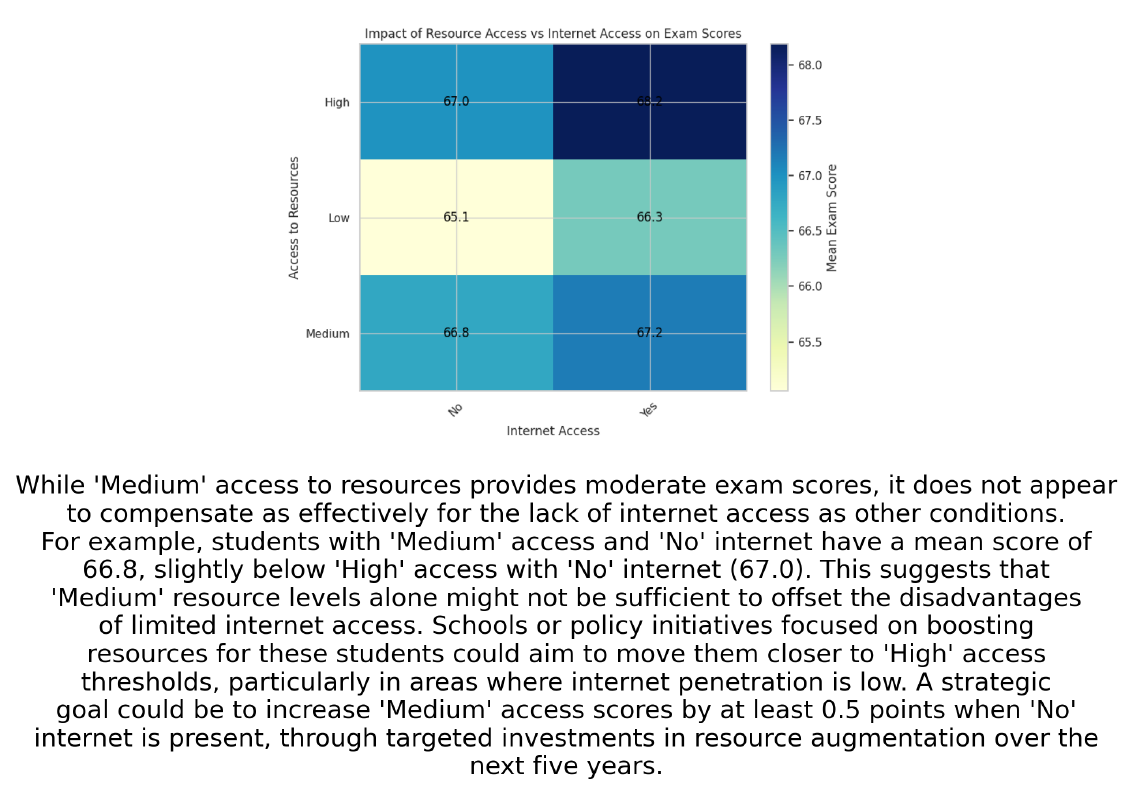}
    \includegraphics[width=0.32\linewidth]{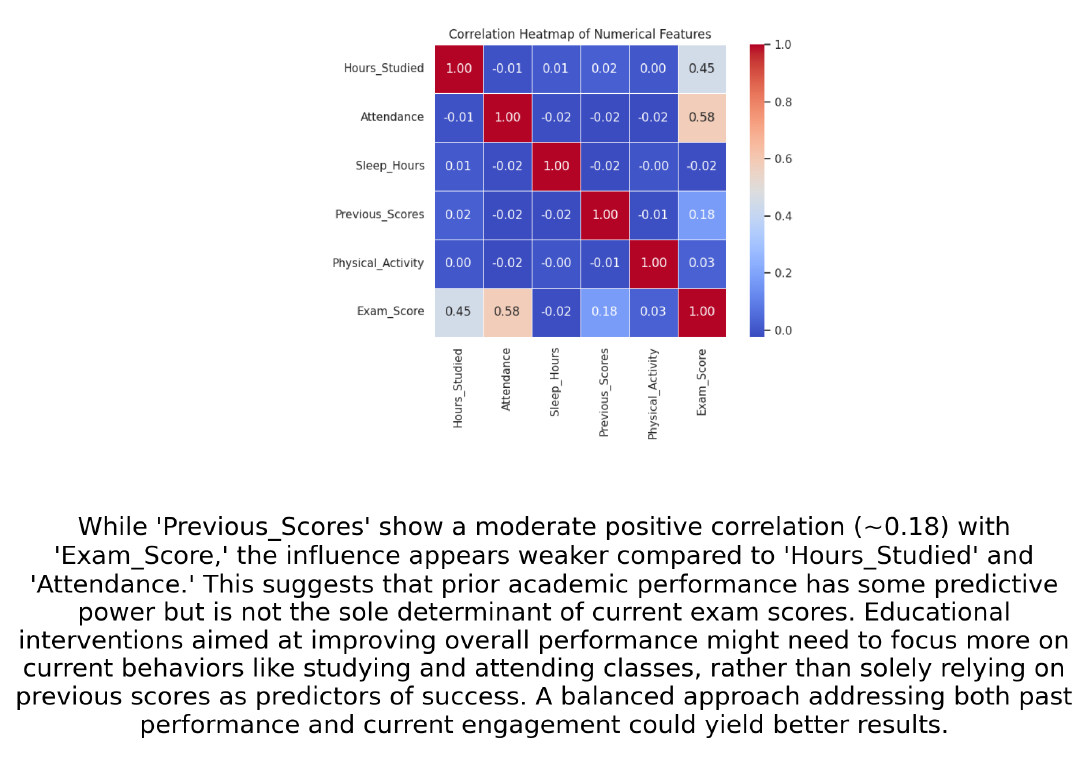}
      \includegraphics[width=0.32\linewidth]{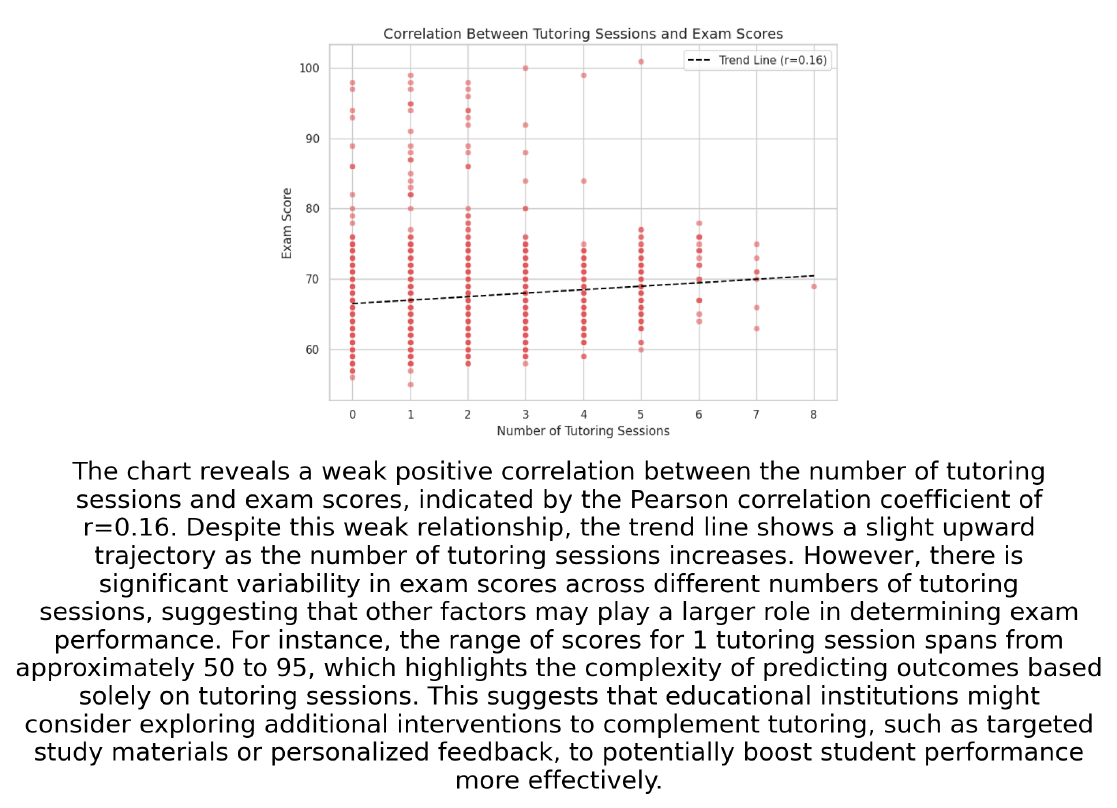}
      \includegraphics[width=0.32\linewidth]{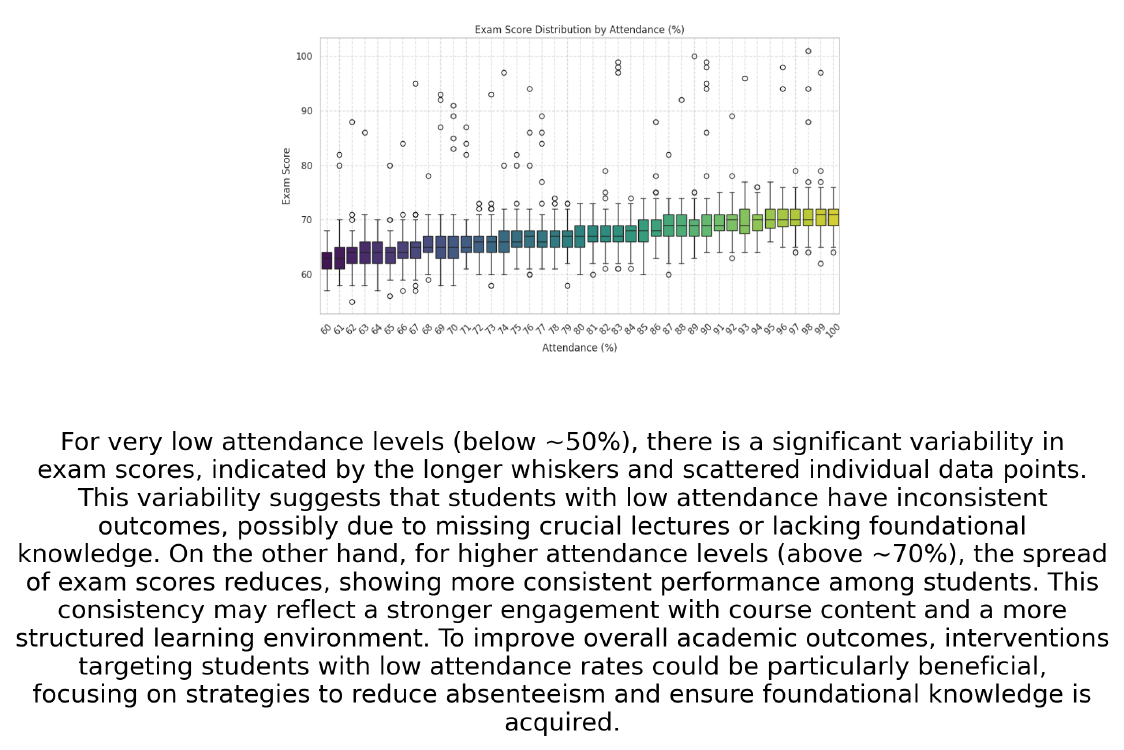}
  \caption{Example reports for Education Performance Dataset}
  \label{fig:add_Education}
\end{figure*}

\begin{figure*}[!htbp]
  \centering
  \includegraphics[width=0.32\linewidth]{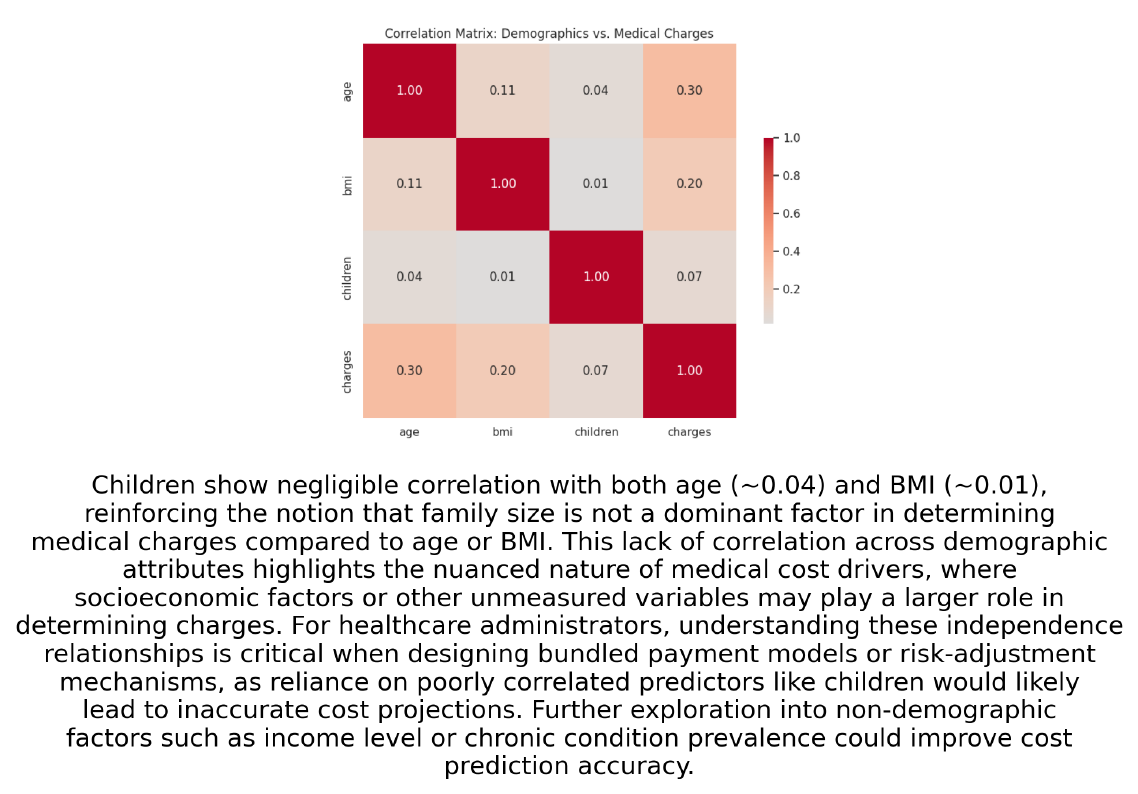}\hfill
  \includegraphics[width=0.32\linewidth]{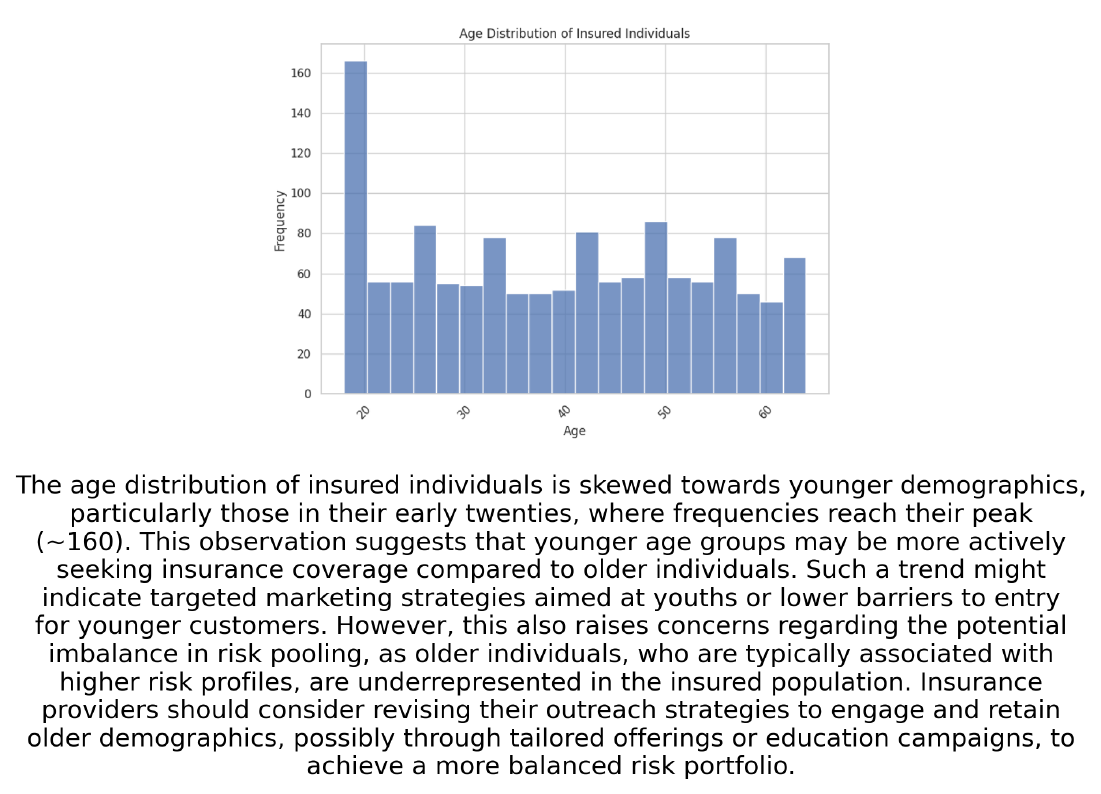}\hfill
  \includegraphics[width=0.32\linewidth]{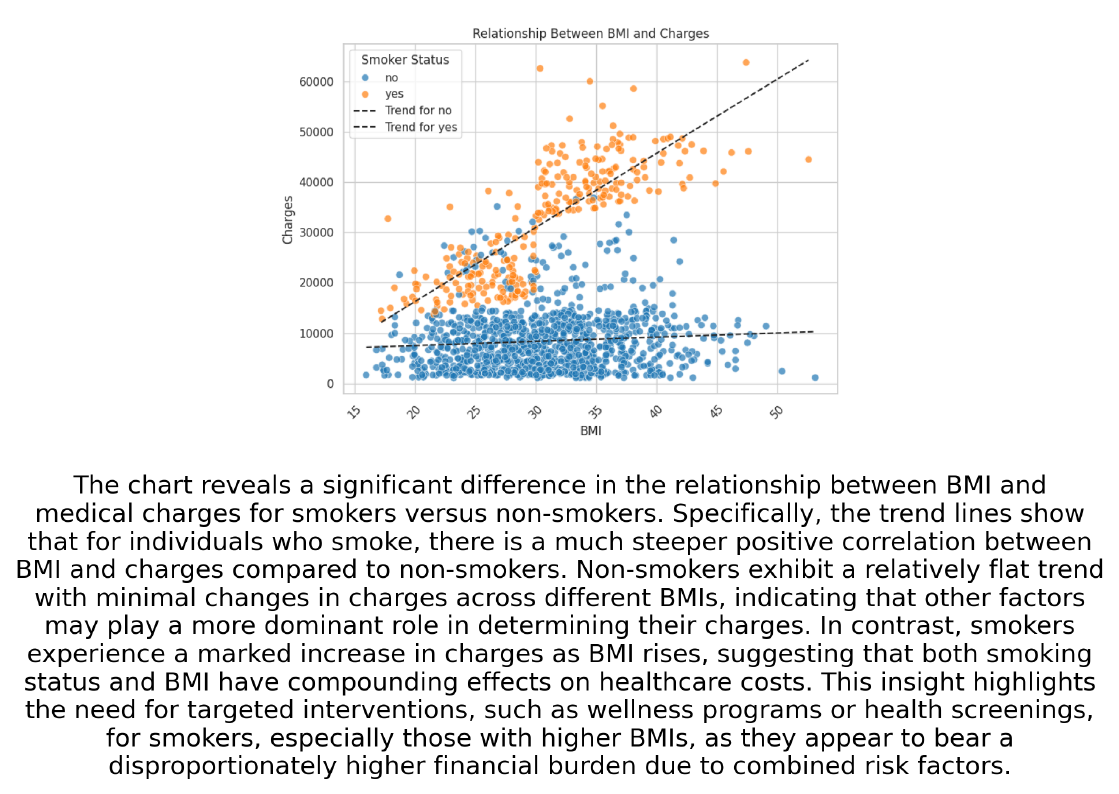}
    \includegraphics[width=0.32\linewidth]{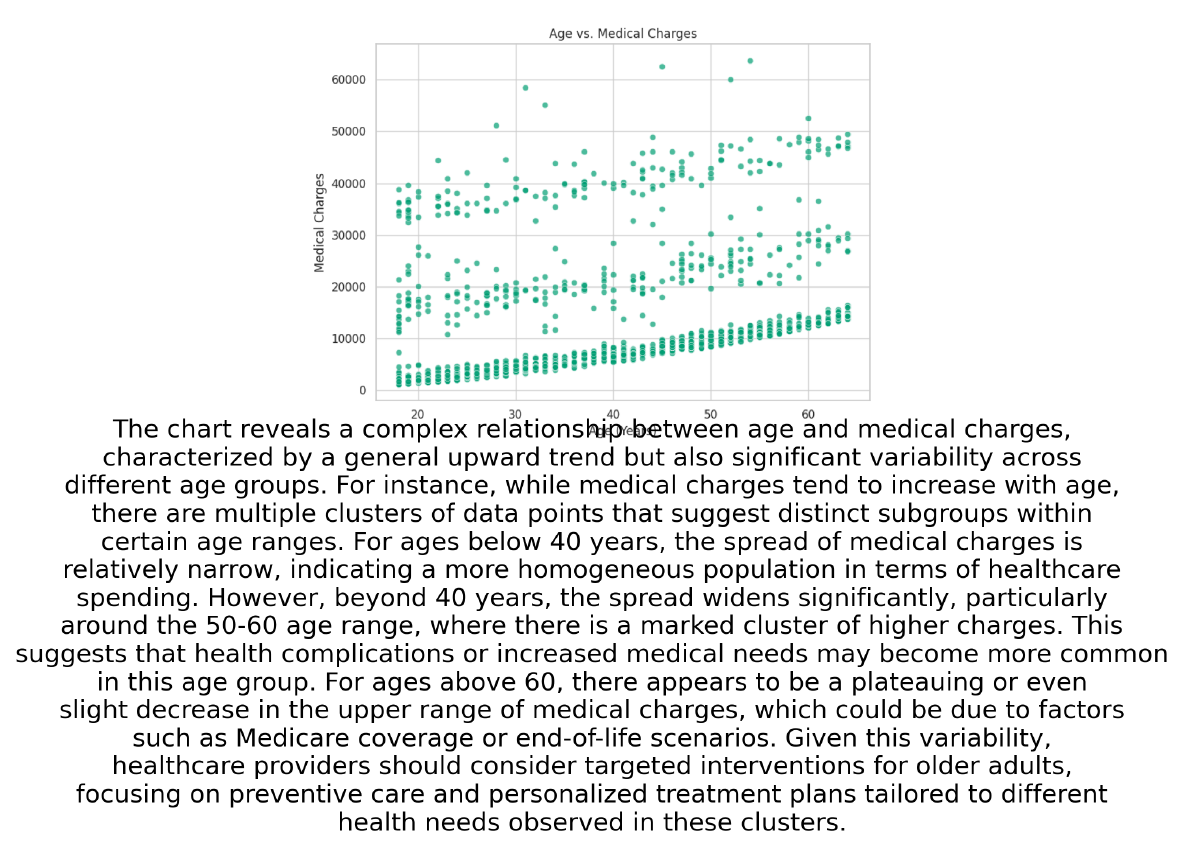}
      \includegraphics[width=0.32\linewidth]{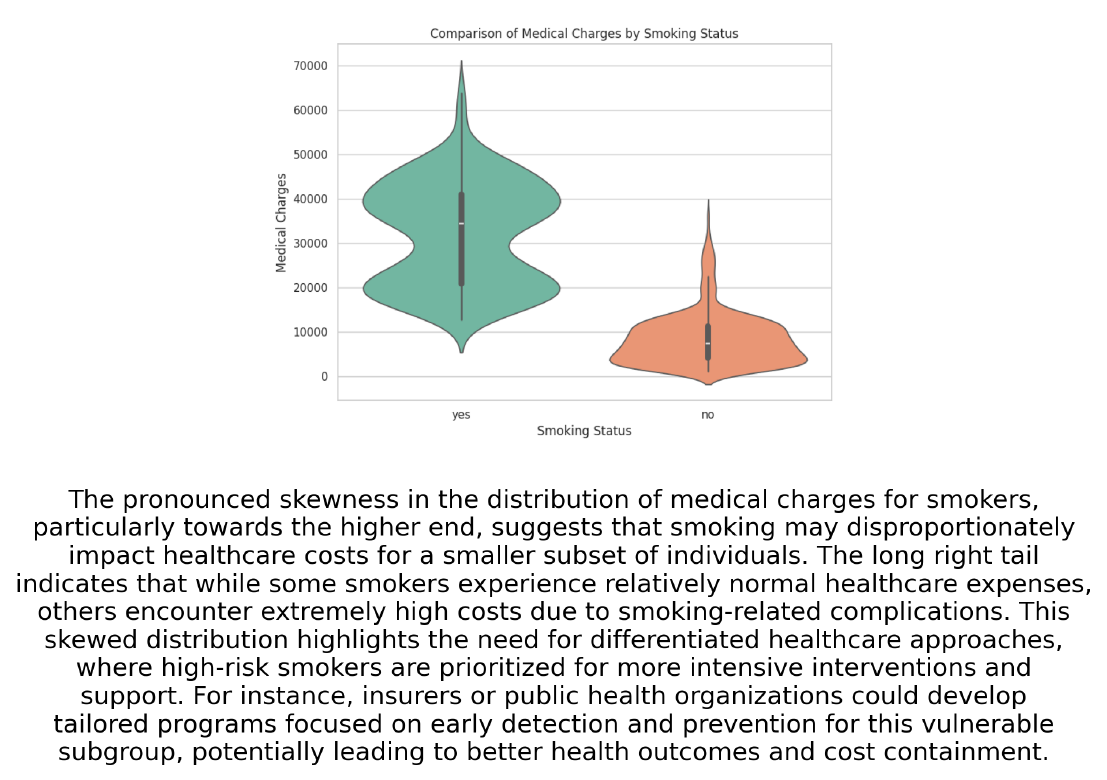}
      \includegraphics[width=0.32\linewidth]{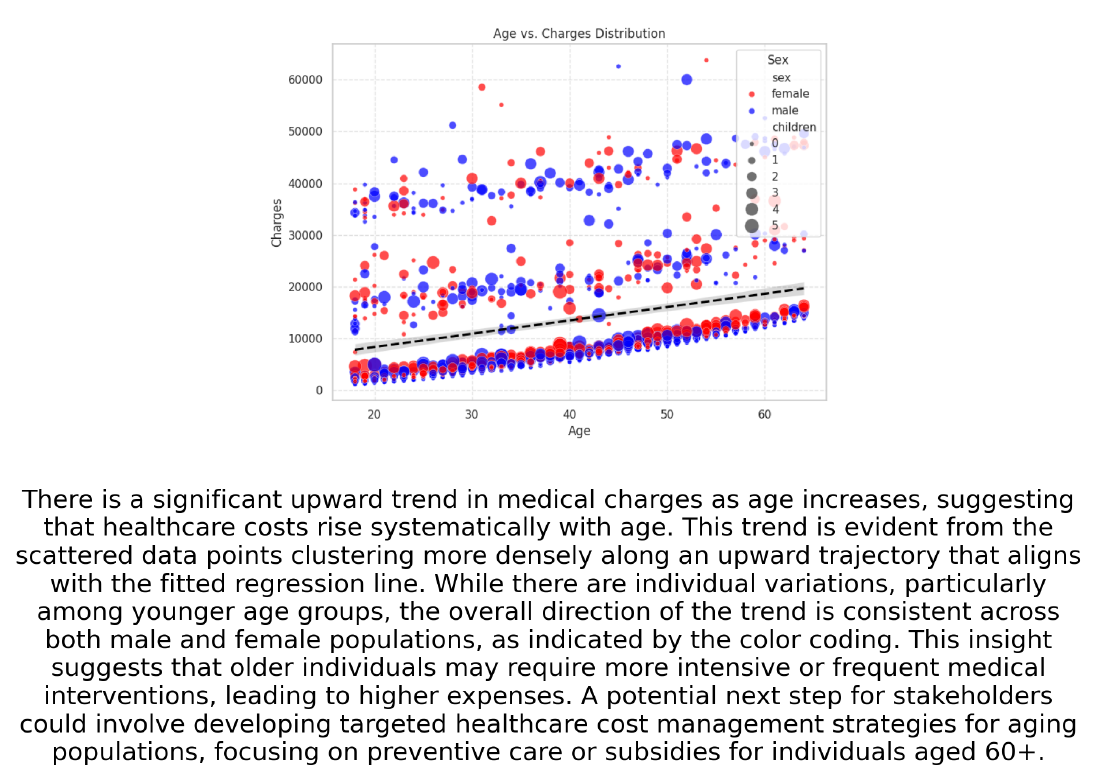}
  \caption{Example reports for Medical Insurance Dataset}
  \label{fig:add_Insurance}
\end{figure*}

\begin{figure*}[!htbp]
  \centering
  \includegraphics[width=0.32\linewidth]{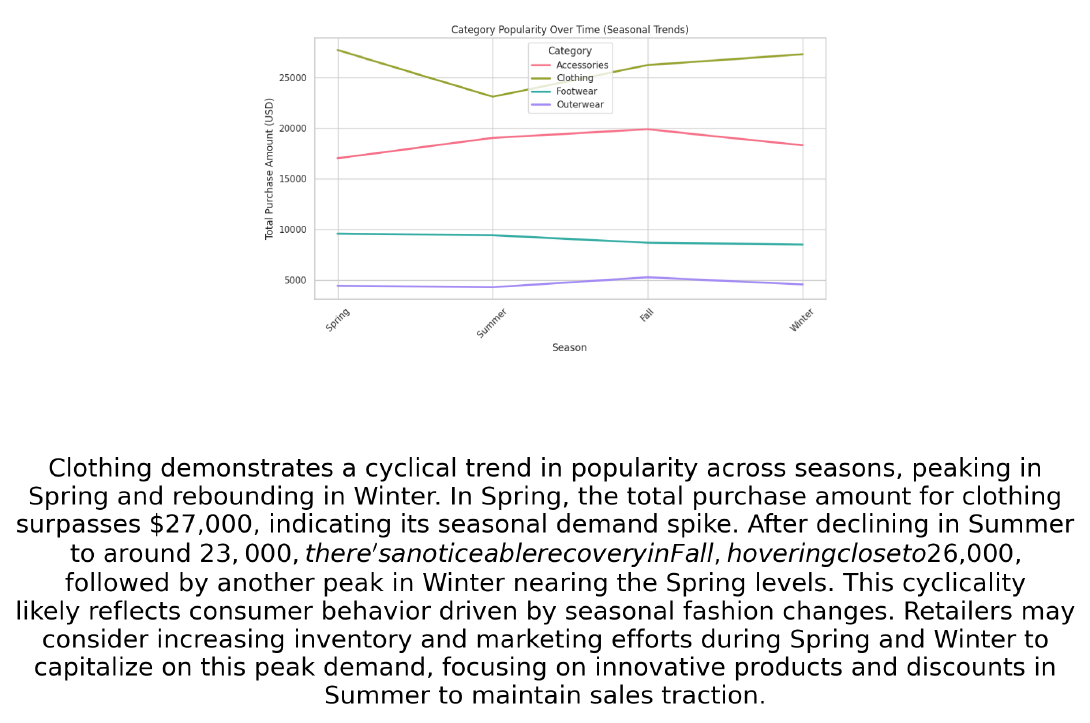}\hfill
  \includegraphics[width=0.32\linewidth]{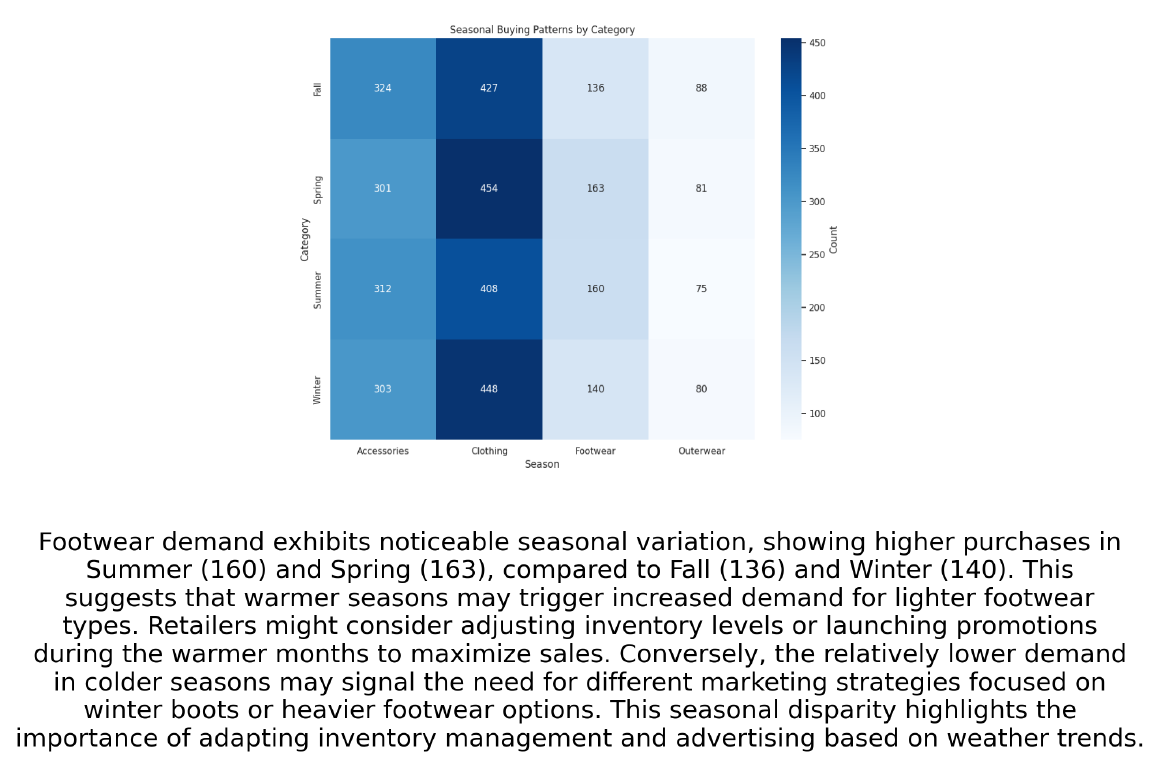}\hfill
  \includegraphics[width=0.32\linewidth]{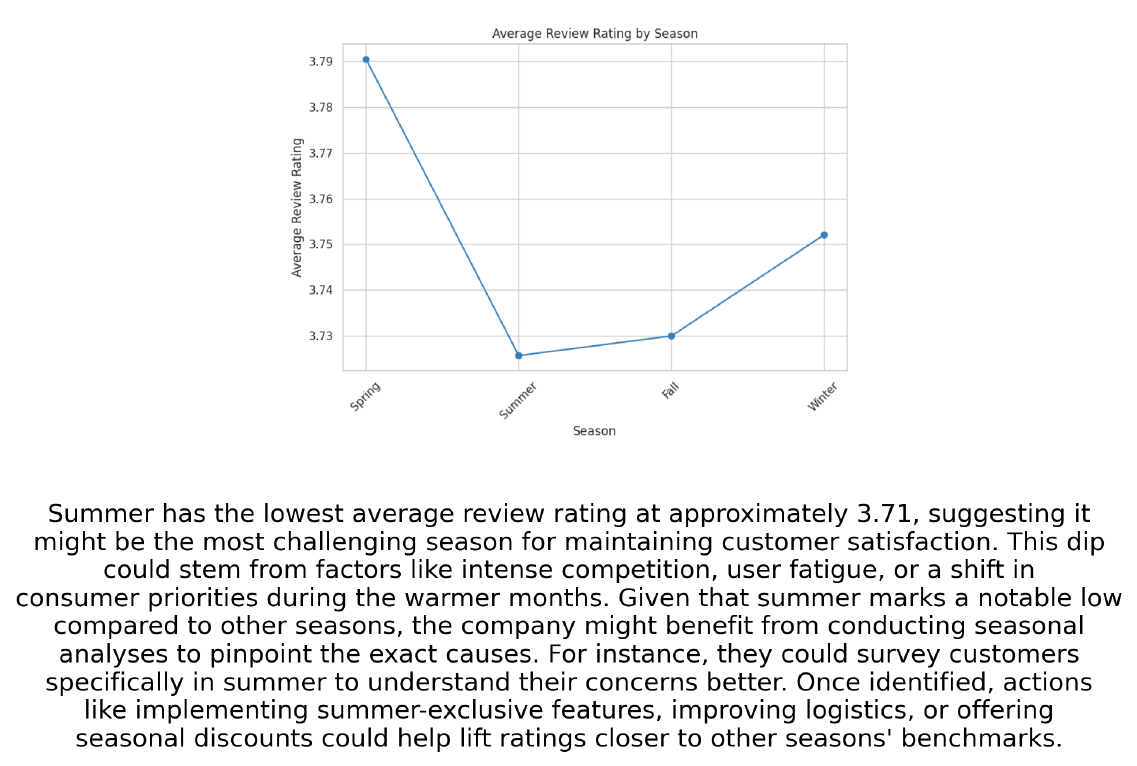}
    \includegraphics[width=0.32\linewidth]{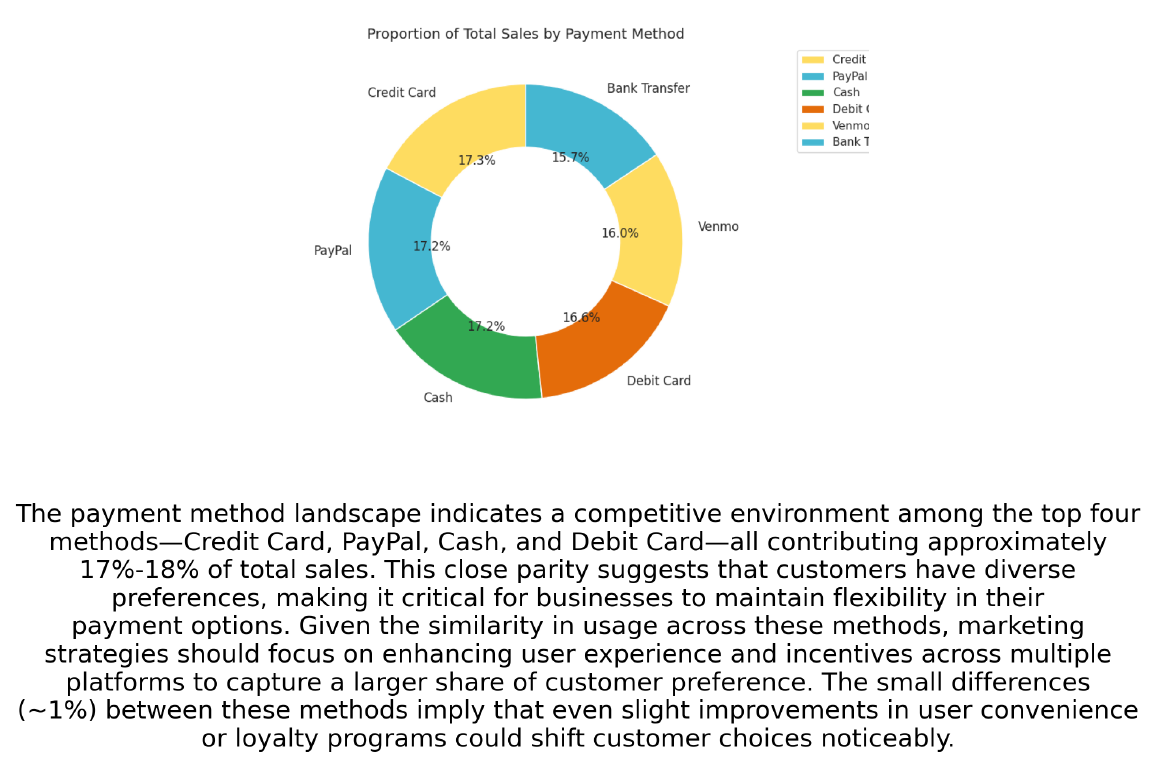}
      \includegraphics[width=0.32\linewidth]{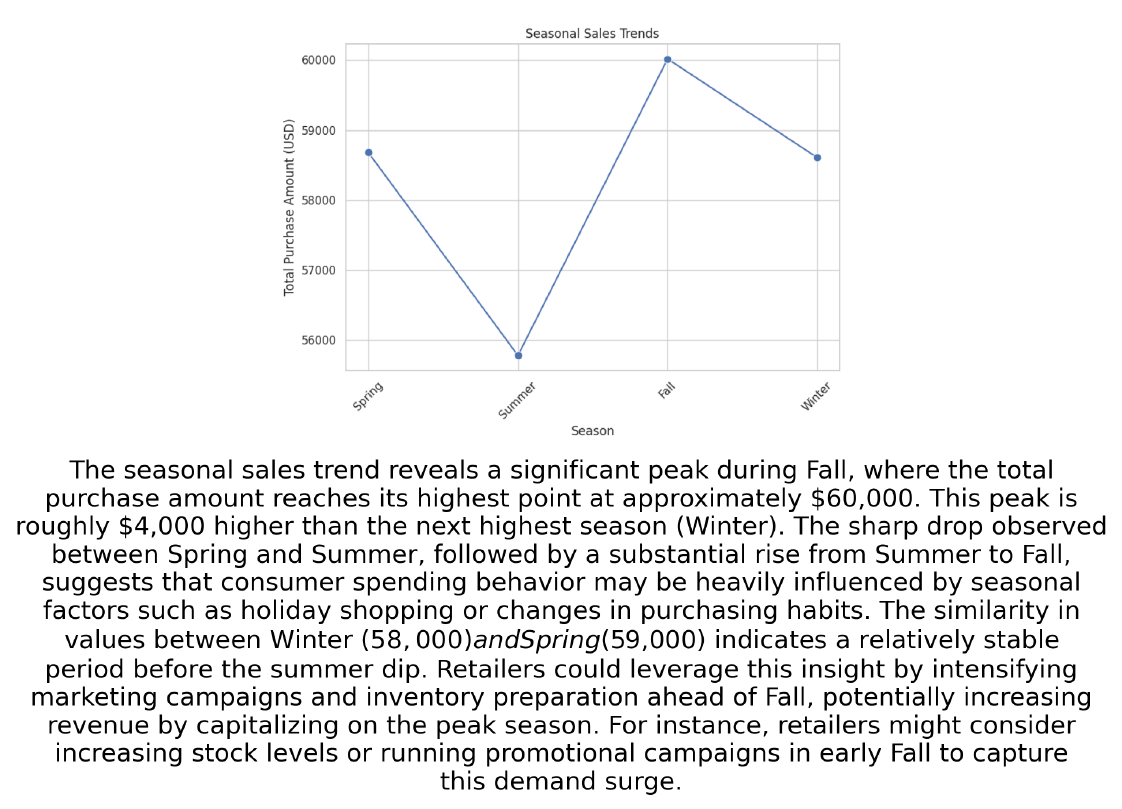}
      \includegraphics[width=0.32\linewidth]{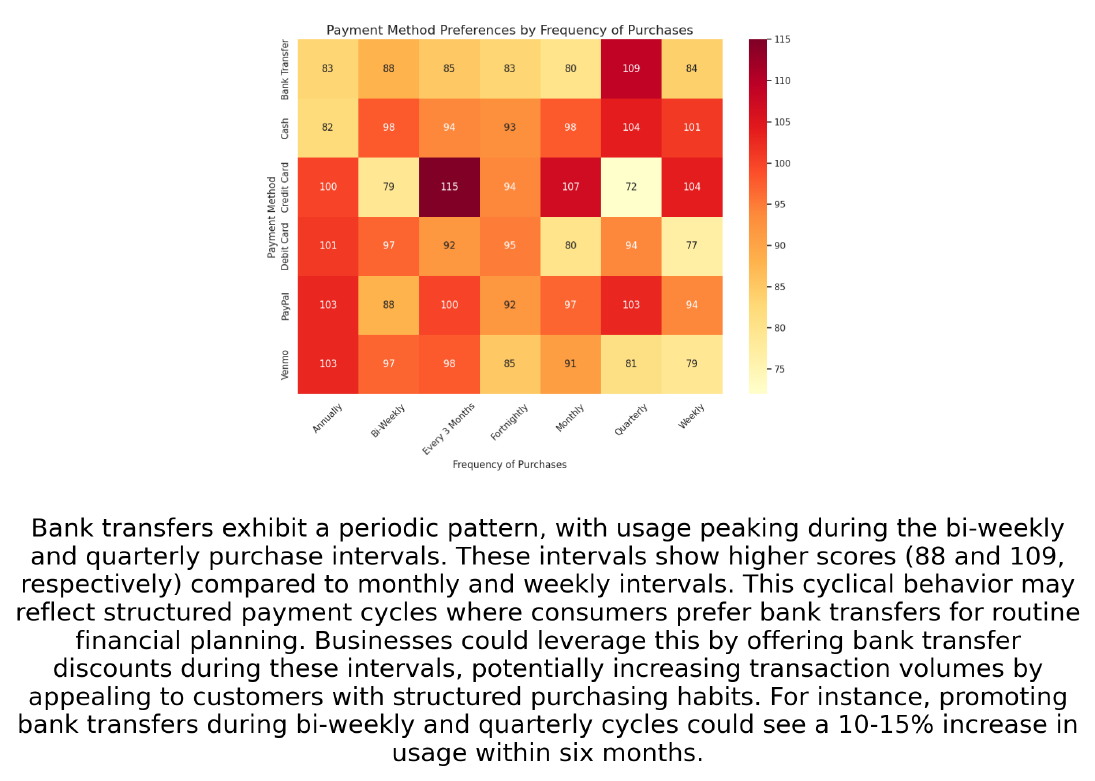}
  \caption{Example reports from Shopping Behavior Dataset}
  \label{fig:add_shopping}
\end{figure*}

\begin{figure*}[!htbp]
  \centering
  \includegraphics[width=0.32\linewidth]{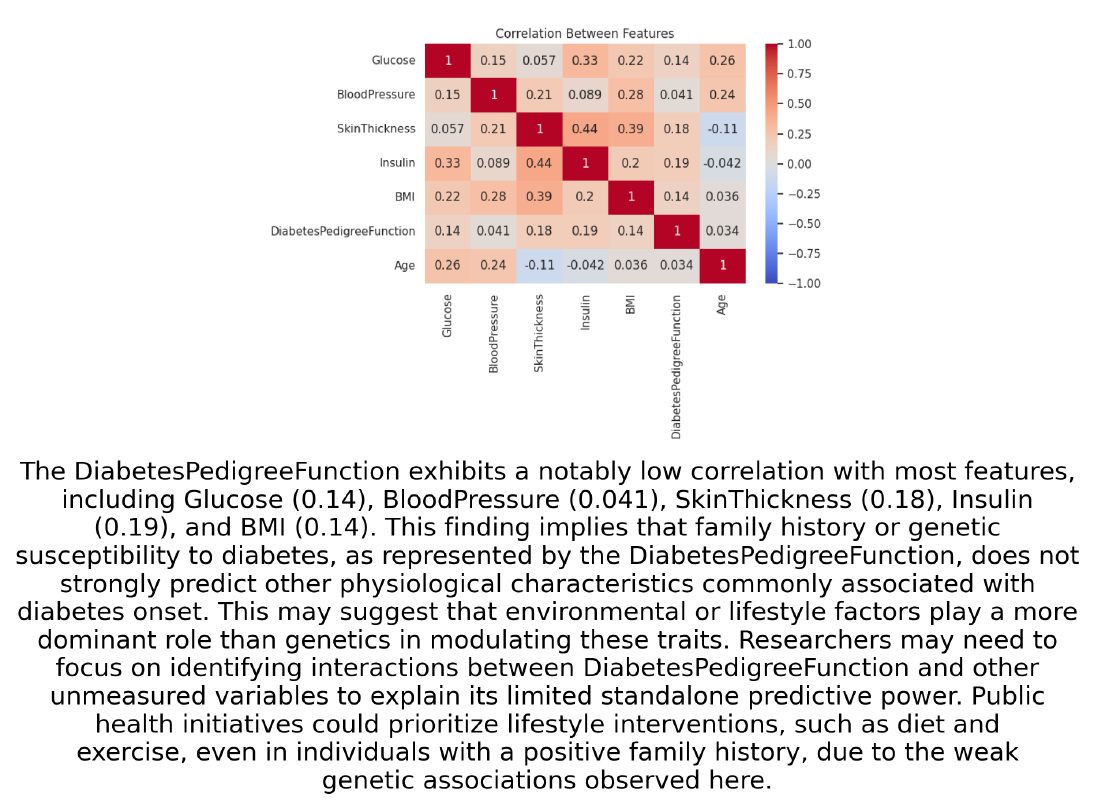}\hfill
  \includegraphics[width=0.32\linewidth]{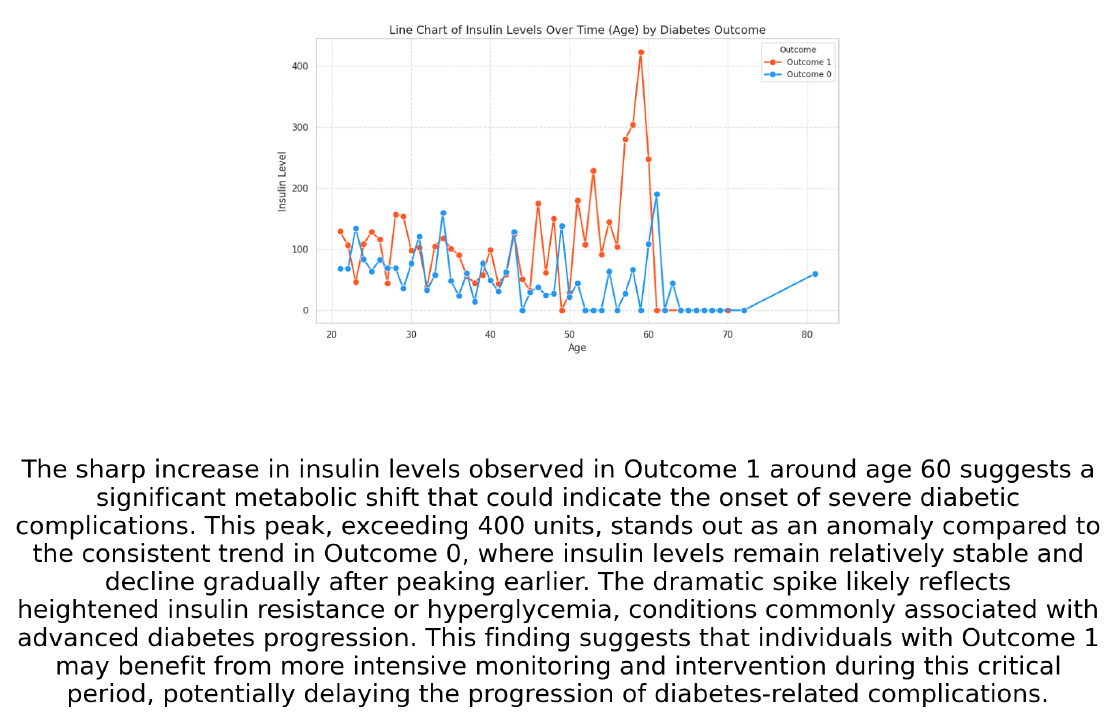}\hfill
  \includegraphics[width=0.32\linewidth]{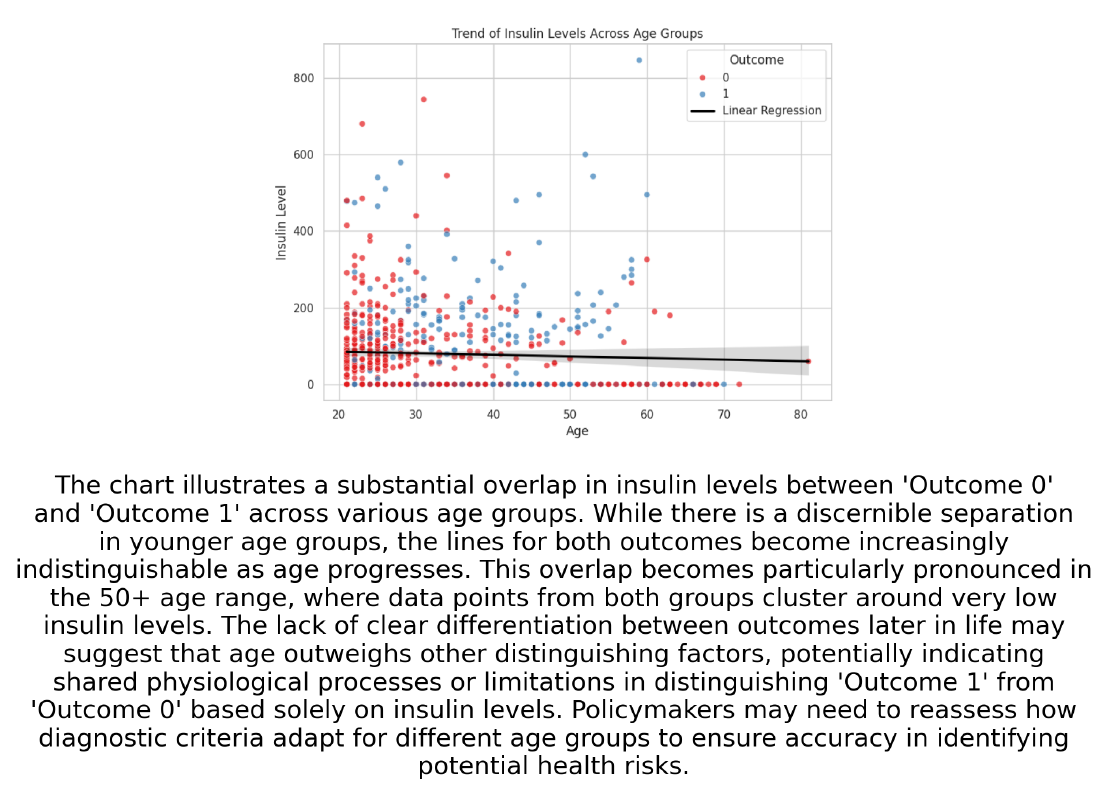}
    \includegraphics[width=0.32\linewidth]{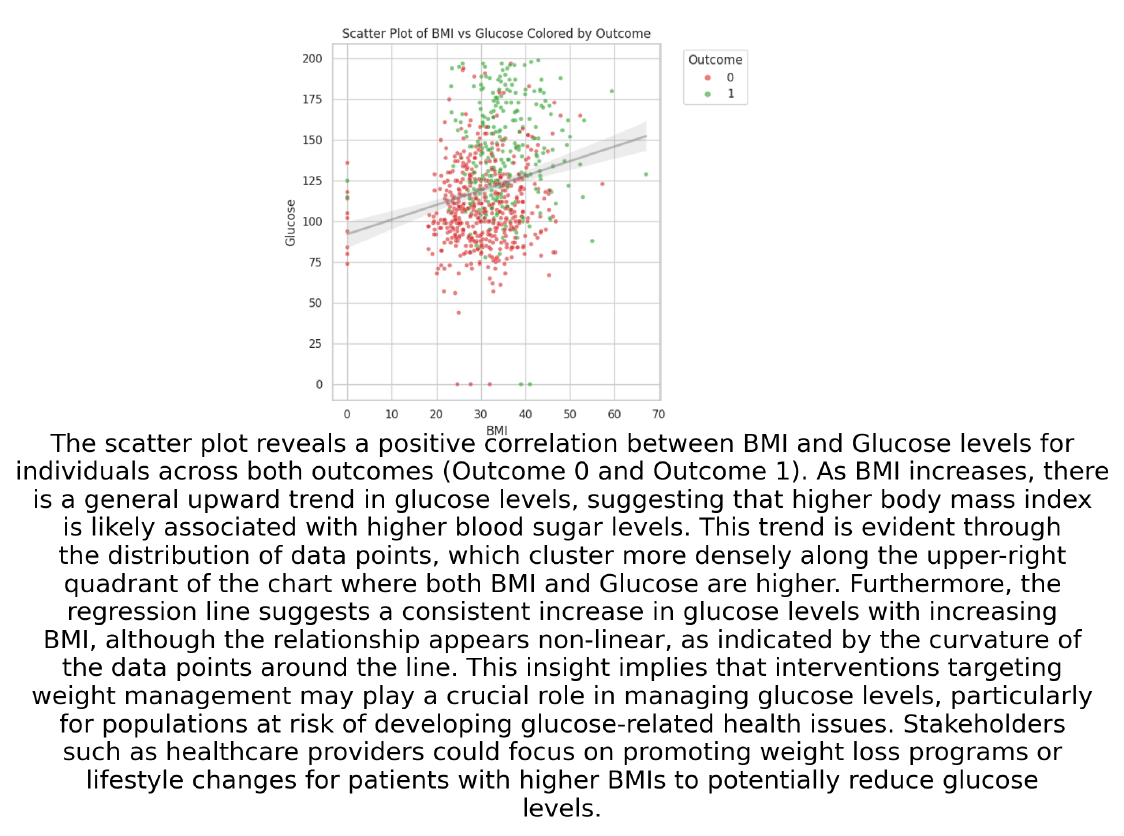}
      \includegraphics[width=0.32\linewidth]{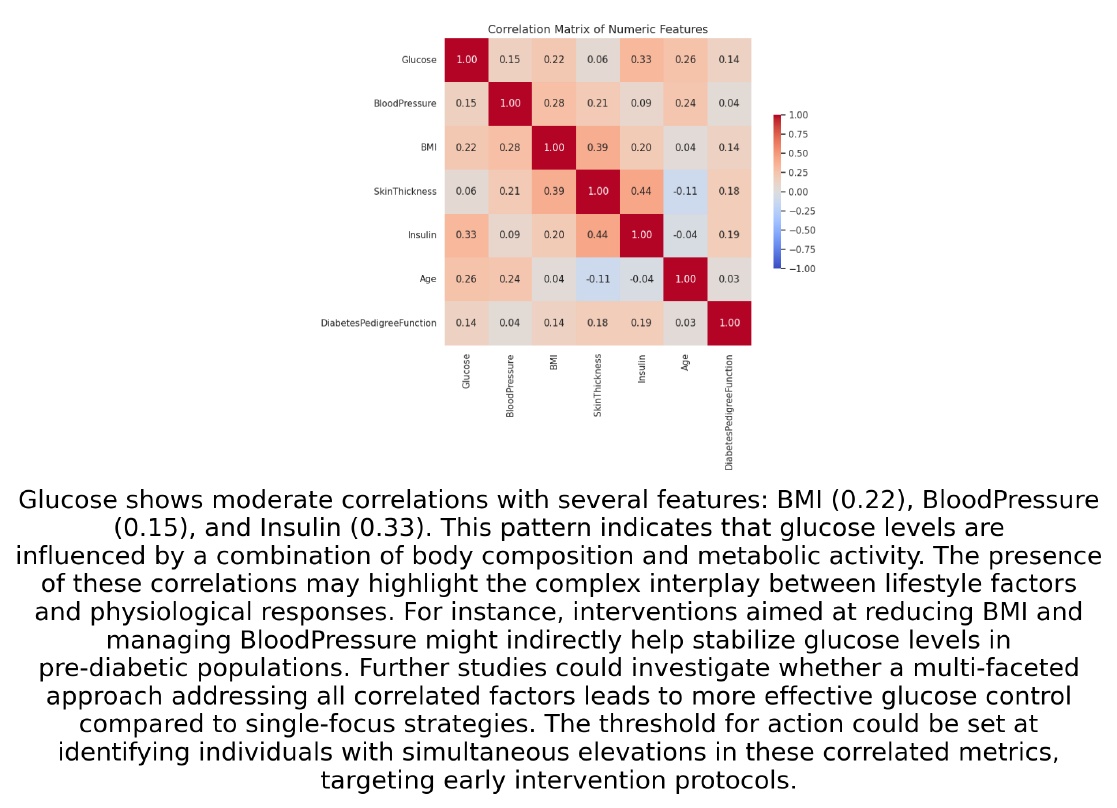}
      \includegraphics[width=0.32\linewidth]{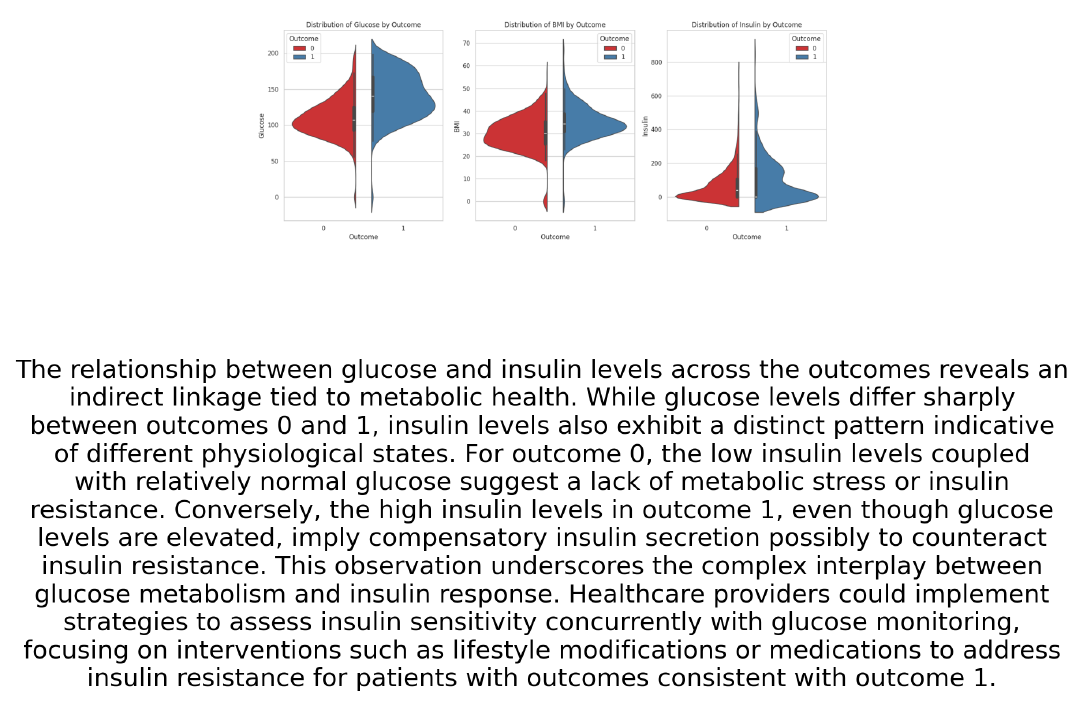}
  \caption{Example reports from Diabetes Dataset}
  \label{fig:add_diabetes}
\end{figure*}

\begin{figure*}[!htbp]
  \centering
  \includegraphics[width=0.32\linewidth]{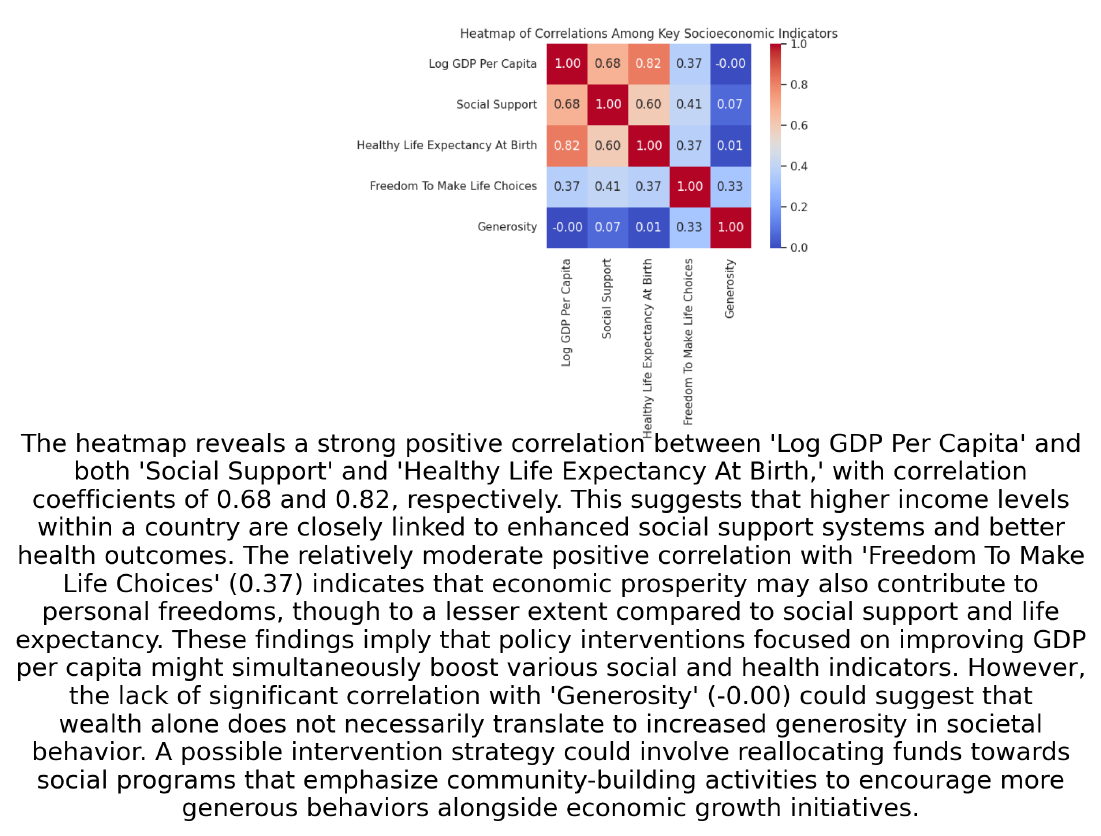}\hfill
  \includegraphics[width=0.32\linewidth]{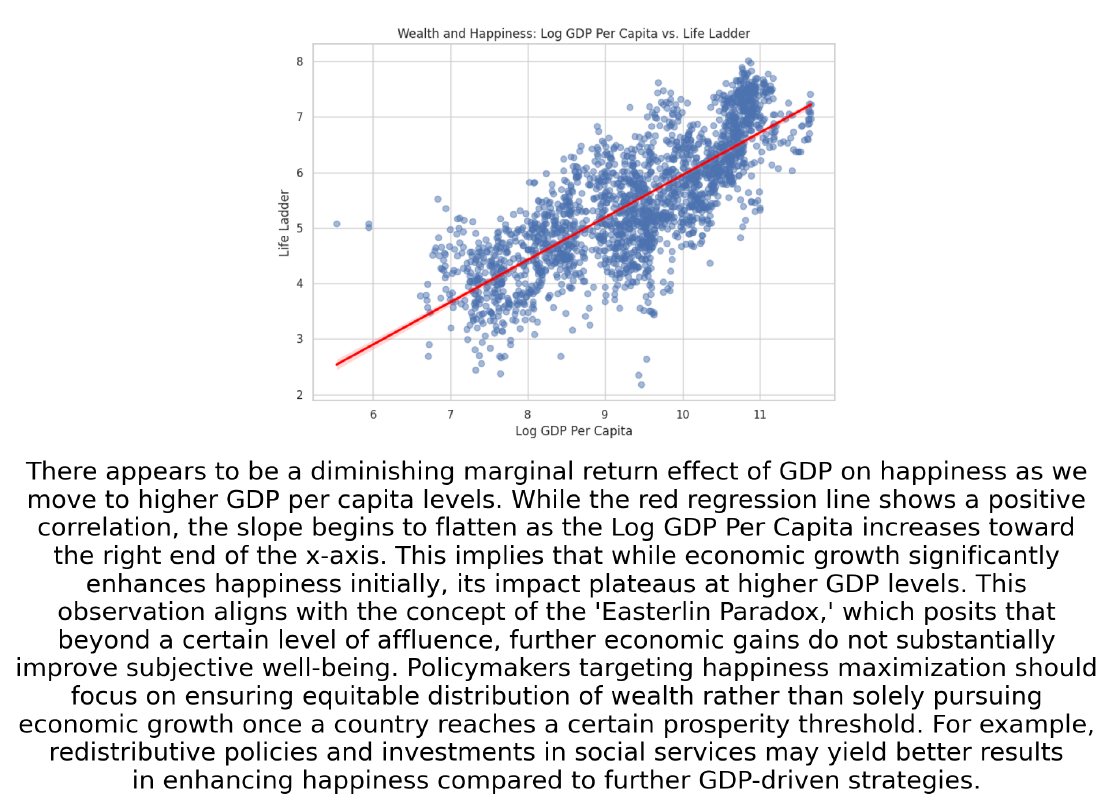}\hfill
  \includegraphics[width=0.32\linewidth]{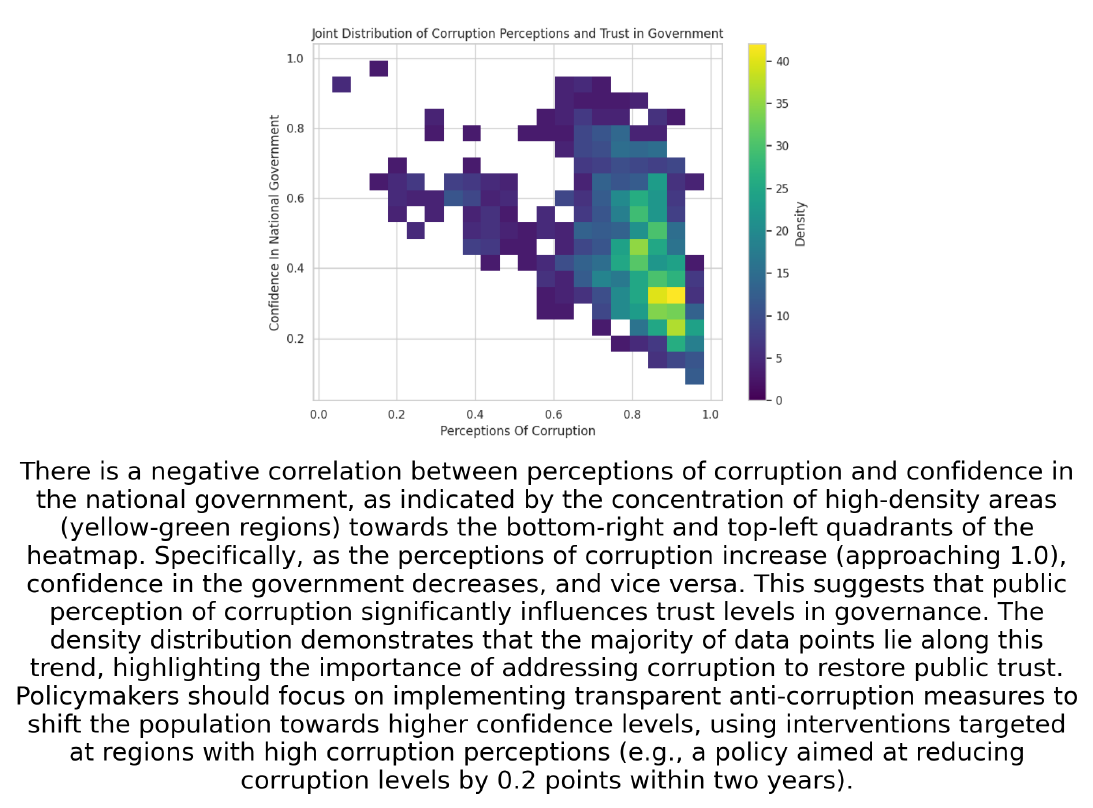}
    \includegraphics[width=0.32\linewidth]{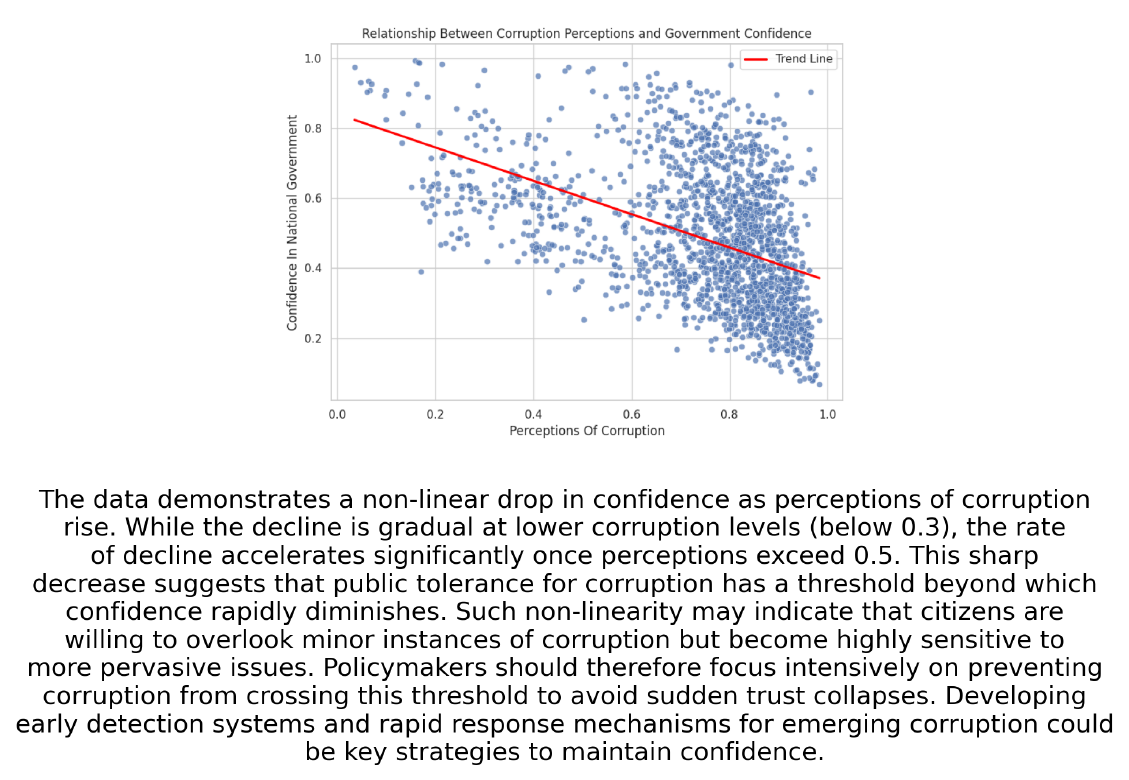}
      \includegraphics[width=0.32\linewidth]{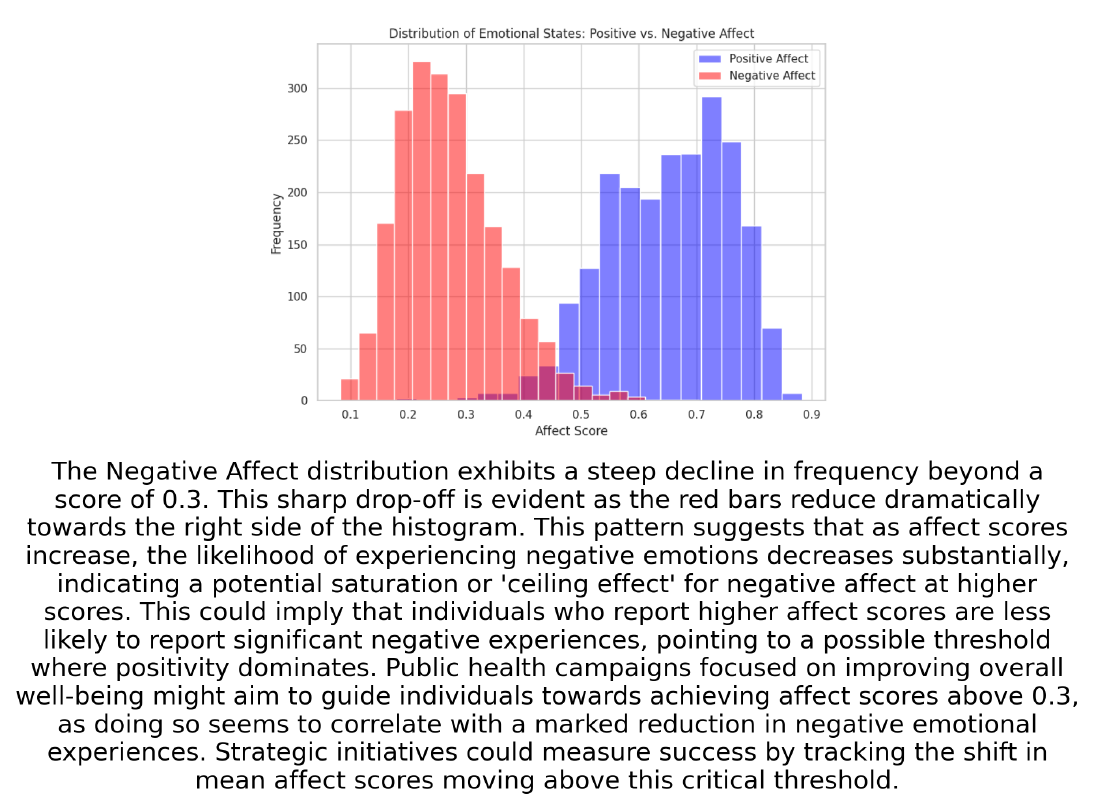}
      \includegraphics[width=0.32\linewidth]{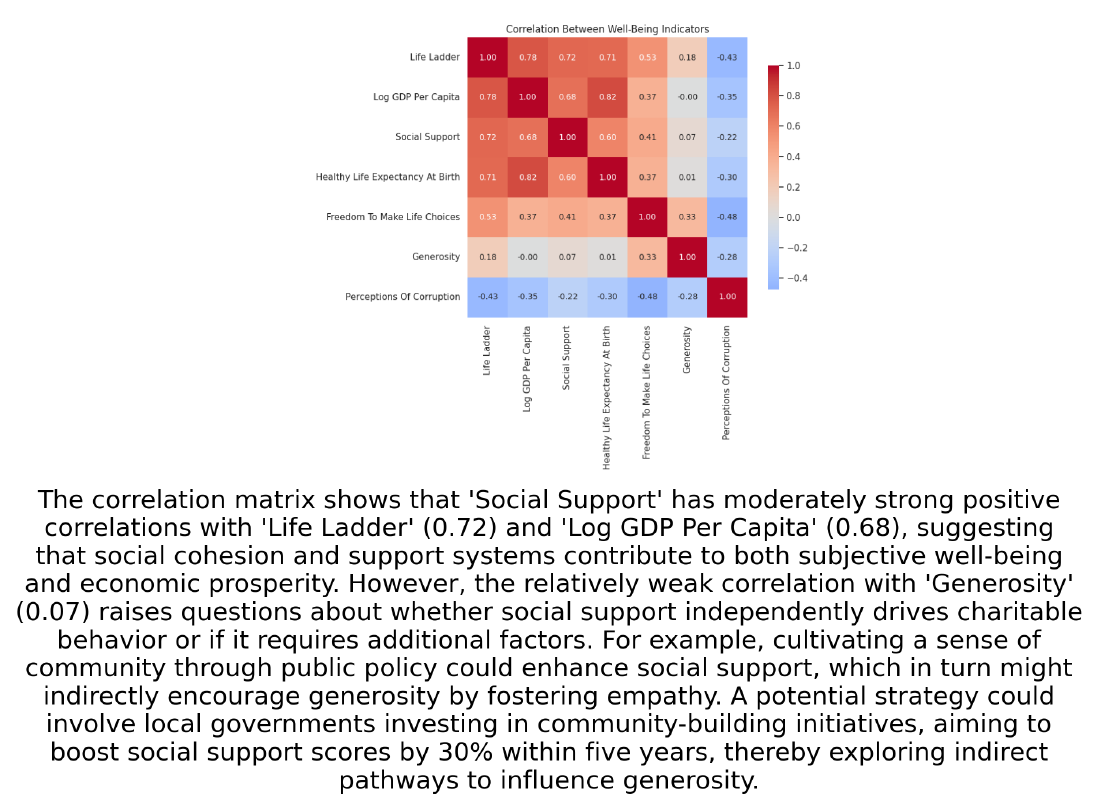}
  \caption{Example reports from World Happiness Report Dataset}
  \label{fig:add_happiness}
\end{figure*}

\section{Supplementary Examples for Qualitative Analysis}
\label{app:qualitative}
To complement the human evaluation in \S\ref{sec:qualitative_human_annotation}, we provide additional qualitative examples to illustrate how report quality evolves across different pruning ratios ($\rho$) and score levels.  
Specifically, we group reports into \textbf{low-}, \textbf{middle-}, and \textbf{high-quality} sets by sampling from the 0\%, 50\%, and 100\% percentiles of each $\rho$’s score distribution, respectively, and jointly annotate them across pruning configurations.  
Our analysis primarily focuses on the \textit{low-} and \textit{high-quality} groups, which exhibit clearer contrasts in both model judgment and human perception.  
These examples further demonstrate how Selective TTS raises the lower bound of report quality while maintaining consistency among top-performing outputs.
\paragraph{Cross-pruning comparisons (low-quality group).}
Fig.~\ref{fig:low_all} compares low-scoring reports across all pruning ratios.  
The differences are more pronounced than in the high-quality group (Fig.~\ref{fig:high_all}).  
At small $\rho$, reports are verbose and generic, whereas increasing $\rho$ improves logical coherence, factual grounding, and domain awareness.  
Annotator rankings within the low-quality group consistently favored reports generated at $\rho{=}0.8$, reflecting their relatively higher coherence and reasoning quality.  
Notably, even the lowest-quality reports at $\rho{=}0.8$ outperform their counterparts at $\rho{=}0$ or $0.2$, indicating a substantial upward shift of the quality floor.  
At the same time, annotator rankings showed larger dispersion among these low-quality samples (Fig.~\ref{fig:rankings}), suggesting that raters could easily distinguish their relative strengths.
\paragraph{Cross-pruning comparisons (high-quality group).}
In contrast, Fig.~\ref{fig:high_all} shows high-scoring reports sampled from each pruning ratio.  
Their content diversity remains high, but overall quality converges, and each report presents well-grounded reasoning with minor stylistic or focus differences (e.g., $\rho{=}0.8$ emphasizes citation inconsistencies, while $\rho{=}0.4$ analyzes paper-length anomalies).  
Annotators rated these reports comparably strong, as reflected in the small variance of mean rankings in Fig.~\ref{fig:rankings}.  

\begin{figure*}[t]
\includegraphics[width=\linewidth]{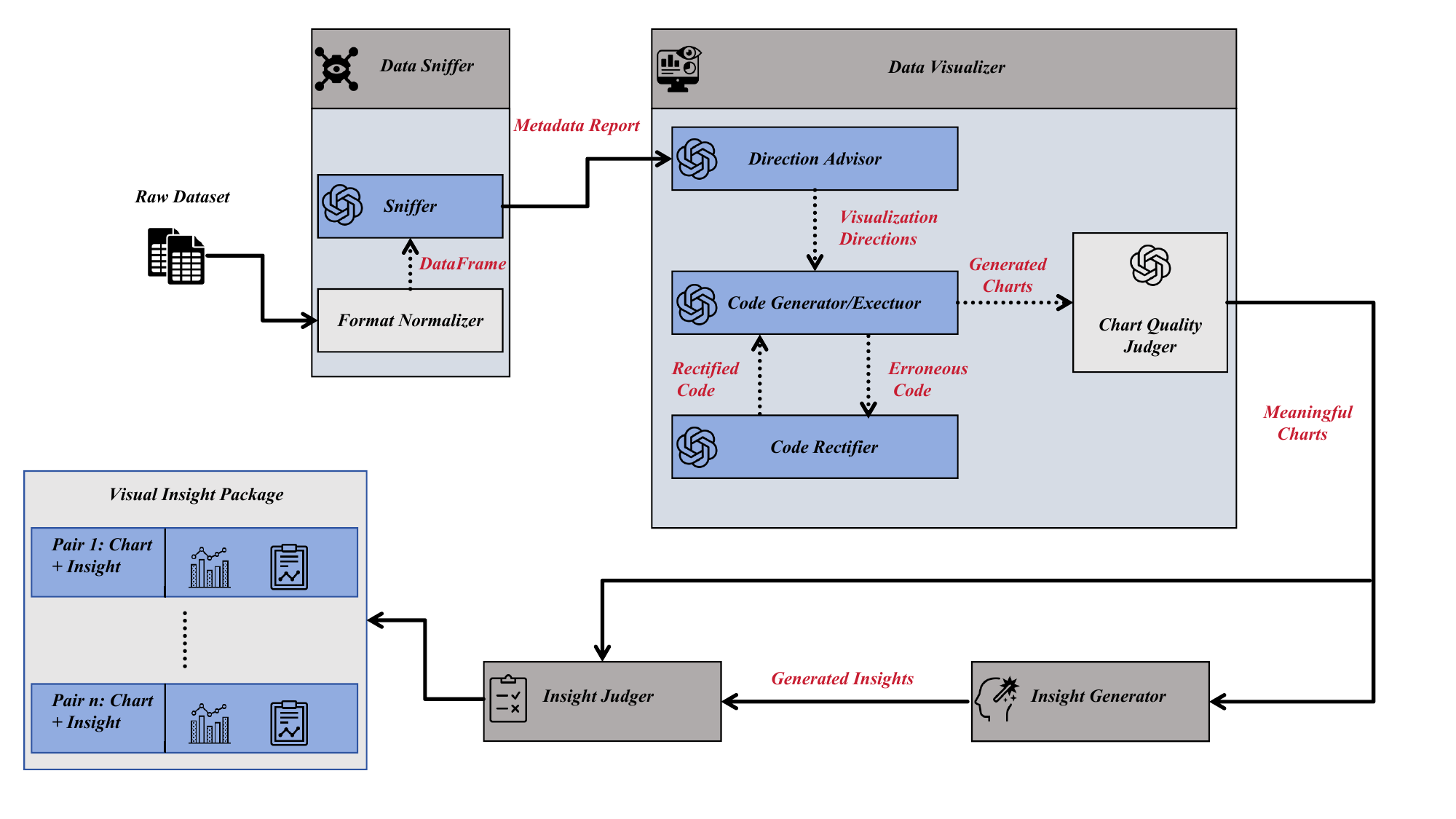}
  \caption{Detailed Pipeline of Insight Generation.}
  \label{fig:workflow}
\end{figure*}

\section{Qualitative Report Examples on Additional Datasets}
\label{app:addtional_datasets}

To complement the quantitative robustness analyses, we provide qualitative examples of reports
generated by our multi-agent pipeline on several additional tabular datasets.
These examples illustrate the types of insights and explanations produced by our method
across diverse domains beyond the VIS publication dataset.

Fig.~\ref{fig:add_Education} shows representative reports generated on the Education Performance dataset.
Fig.~\ref{fig:add_Insurance} presents example reports from the Medical Insurance dataset.
Fig.~\ref{fig:add_shopping} displays reports generated on the Shopping Behavior dataset.
Fig.~\ref{fig:add_diabetes} provides examples from the Diabetes dataset.
Finally, Fig.~\ref{fig:add_happiness} shows reports generated on the World Happiness Report dataset.

\section{Pipeline Design}
\label{app:pipeline}
This section provides an extended description of our multi-agent pipeline architecture and its core design components.  
We first present an end-to-end overview of the pipeline, detailing how data flows through generation, evaluation, and pruning stages (\S\ref{app:finegrainedpipeline}).  
We then describe the prompt design and specifications for all major agents and judgers involved in each stage, including both task-level generation and stage-local evaluation modules (\S\ref{app:prompt}).
\subsection{End-to-End Pipeline Overview}
\label{app:finegrainedpipeline}
Fig.~\ref{fig:workflow} presents the full workflow of our proposed multi-agent data science pipeline, expanding upon the conceptual stages illustrated earlier in Fig.~\ref{fig:pipline}.
While Fig.~\ref{fig:pipline} provides a high-level overview of the agents and their interconnections, Fig.~\ref{fig:workflow} offers a more detailed depiction of the end-to-end process, showing how data samples are progressively refined through generation, evaluation, and pruning across multiple stages.
\subsection{Prompt Design and Specification}
\label{app:prompt}
We present the prompt design for three major components of our pipeline: 
(1) generation agents for data profiling, visualization, and insight generation, 
(2) multi-level judgers for final evaluation, and 
(3) stage-local evaluators used in Selective TTS for process-based pruning.
\paragraph{Data Profiling, Visualization, and Insight Generation Agents.}
As illustrated in Fig.~\ref{fig:workflow}, the core pipeline comprises six generation components:
the \emph{Data Profiling}, \emph{Visualizatio Direction Generation}, \emph{Code Generation}, \emph{Code Rectification}, \emph{Chart Quality Filtering}, and \emph{Insight Generator}.
Each component operates under a modular system prompt that specifies its role, expected input–output format. The complete system prompts for these generation components are presented below.
\begin{tcolorbox}[
  colback=gray!5, colframe=gray!30,
  boxrule=0.5pt, arc=2mm,
  left=4pt, right=4pt, top=2pt, bottom=2pt,
  title={System Prompt: Data Profiling},
  breakable
]
\scriptsize\ttfamily
You are a professional data curator writing an official-style “About Dataset” page (as on Kaggle/Hugging Face).\\
You will ONLY receive the metadata information and sample rows of the dataset: (1) dataset shape (rows x columns); (2) column names with detected data types; (3) 1--2 sample rows.\\
Your task is to generate a clear, structured, and comprehensive dataset introduction strictly based on the provided information.\\
Base ALL content strictly on this input. Do NOT infer anything about missing values, full distributions, correlations, provenance, licensing, collection period, or coverage unless explicitly evident from names or the sample.\\

\textbf{1. Your goals}\\
1) \textbf{Briefly introduce the dataset}: describe what it appears to contain and how it is structured (refer to shape and the mix of data types).\\
2) \textbf{Explain each variable in detail}: infer likely meanings from column names and the sample (use cautious language such as ``likely'' or ``appears to''); mention units or ranges only if clearly suggested (e.g., \_usd, \_pct, lat/lon, date).\\
3) \textbf{Propose potential analysis directions and visualization themes}: suggest exploratory analyses grounded in the schema (e.g., numeric distributions, categorical comparisons, temporal trends, geographic patterns if names imply geo fields, keyword summaries for text).\\
4) Provide clear, structured documentation that downstream chart planners and code generators can follow immediately.\\

\textbf{2. Output format (Markdown)}\\
Use the following sections and formatting in markdown:\\

\#\# About Dataset\\
-- \textbf{Shape}: <rows> x <cols>. Briefly describe the schema and mix of types (numeric/categorical/text/datetime).\\
-- \textbf{High-level summary}: A concise statement of what the table likely represents (strictly based on names and sample).\\

\#\# Schema Summary\\
Output as CSV-style plain text lines using the header:\\
Column, Detected Type, Example, Likely Meaning, Suggested Role\\
Each subsequent line corresponds to one column. For example:\\
id, integer, "12345", likely an identifier for each record, id\\
Where \textit{Suggested Role} $\in$ \{feature, target, id, timestamp, text, geo, meta\}.\\

\#\# Potential Uses \& Analysis Directions\\
Be concrete and grounded in the provided schema.\\
-- Suggest meaningful exploratory analyses and visualizations.\\
-- If a column appears to be a target variable, note potential modeling tasks.\\

\textbf{3. Output rules}\\
-- Use clear Markdown headings; for tables, use CSV-like plain text, not Markdown tables.\\
-- Be specific yet cautious: never invent facts beyond names/sample.\\
-- Tone must be professional and documentation-like, as if authored by the dataset creator.\\
\end{tcolorbox}

\begin{tcolorbox}[
  colback=gray!5,
  colframe=gray!40,
  boxrule=0.4pt,
  arc=2mm,
  left=5pt, right=5pt, top=3pt, bottom=3pt,
  title={System Prompt: Visualization Direction },
  breakable
]
\scriptsize\ttfamily
You are a professional data visualization expert. Your task is to propose well-defined visualization directions for a given dataset.\\

\textbf{You will be provided with:}\\
- Metadata information (number of rows/columns, data types, etc.)\\
- Sample data from the dataset\\[3pt]

\textbf{Requirement:}\\
1. Generate \{num\_directions\} \textbf{concise, diverse, and actionable} directions for visualizing the dataset.\\
2. The directions should not overlap in their focus; each direction should focus on a different aspect of the data.\\
3. Each direction must:\\
\hspace{1em}-- Focus on a specific aspect of the data (e.g., distribution, correlation, time trend, category comparison, anomaly detection).\\
\hspace{1em}-- Include \textbf{all} of the following keys: \texttt{topic}, \texttt{chart\_type}, \texttt{variables}, \texttt{explanation}, \texttt{parameters}.\\
\hspace{1em}-- Use only existing column names from the dataset in \texttt{variables}.\\
\hspace{1em}-- \texttt{chart\_type} should be a standard, executable visualization type (e.g., “bar”, “line”, “scatter”, “histogram”, “boxplot”, “heatmap”, “wordcloud”).\\
\hspace{1em}-- \texttt{parameters} may include chart-specific options such as sorting, grouping, aggregation method, bins, color scheme, etc.\\
4. Avoid vague or generic suggestions. Each explanation should state \textbf{why} the chart is relevant and what insights it may reveal.\\
5. Try to use \textbf{different chart types} and cover various analytical angles across the directions. Be creative!\\[3pt]

\textbf{Output format (all keys mandatory, valid JSON only):}
\begin{verbatim}
[
  { "topic": "...",
    "chart_type": "...",
    "variables": ["...", "..."],
    "explanation": "...",
    "parameters": {"param1": "...", "param2": "..."}
  },
  { "topic": "...",
    "chart_type": "...",
    "variables": ["...", "..."],
    "explanation": "...",
    "parameters": {"param1": "...", "param2": "..."}
  }
]
\end{verbatim}

\textbf{Attention:}\\
- Output \textbf{only} the JSON array inside a pure Markdown code block (\texttt{```json ... ```}), without any extra text, commentary, or formatting.\\
- Ensure the JSON is valid and directly parsable.\\
- Strict adherence to the output format is required.
\end{tcolorbox}
\begin{tcolorbox}[
  colback=gray!5,
  colframe=gray!40,
  boxrule=0.4pt,
  arc=2mm,
  left=5pt, right=5pt, top=3pt, bottom=3pt,
  title={System Prompt: Code Generation },
  breakable
]
\scriptsize\ttfamily
You are a Python data visualization code generator. Your task is to generate \textbf{production-ready, executable} Python code based on the following inputs:
- Metadata information: data types, shape of the dataset, introduction of the dataset and sampled data.
- A given visualization direction (topic, chart\_type, variables, parameters).

\textbf{Requirements:}
1. The code must:
- Use ONLY: \texttt{import pandas as pd}, \texttt{import matplotlib.pyplot as plt}, \texttt{import seaborn as sns}, \texttt{import numpy as np}, \texttt{import networkx as nx}, \texttt{from wordcloud import WordCloud}.
- Always set matplotlib backend to \texttt{"Agg"} before importing pyplot. Do not use \texttt{plt.show()}, only save figures to the given path.

- Load the dataset (CSV format) using \texttt{df = pd.read\_csv(\{data\_path\})}.

- Implement the visualization as specified in the given direction.

- Add a descriptive title, x-axis label, and y-axis label.

- Apply a consistent style, e.g., \texttt{sns.set\_theme(style="whitegrid")} and \texttt{plt.figure(figsize=(8, 6))}.

- Ensure category labels on the x-axis are rotated if necessary (\texttt{plt.xticks(rotation=45)}).

- Call \texttt{plt.tight\_layout()} before saving.

- Save the plot to \texttt{plt.savefig(\{output\_path\})} and then call \texttt{plt.close()}.

2. The code must be directly executable without modification.  
3. Pay attention to data types (categorical vs. numerical).  
4. Do \textbf{not} create or print any other output besides the plot file.  

5. Always import all necessary libraries explicitly (e.g., \texttt{import seaborn as sns}, \texttt{import pandas as pd}, \texttt{import matplotlib.pyplot as plt}). 

6. Ensure the code is executable and contains no undefined variables or functions.

\textbf{Output format:}
\begin{verbatim}
```python
# no comments
...(code here)...
\end{verbatim}

\textbf{Attention:}
	1.	Output only valid Python code inside a Markdown code block.
    
	2.	Do not use any libraries beyond pandas, matplotlib, seaborn, numpy, networkx, and wordcloud.
    
	3.	Return only valid, ready-to-run Python code.
    
	4.	Ensure syntax correctness and reproducibility.
\end{tcolorbox}
\begin{tcolorbox}[
  colback=gray!5,
  colframe=gray!40,
  boxrule=0.4pt,
  arc=2mm,
  left=5pt, right=5pt, top=3pt, bottom=3pt,
  title={System Prompt: Code Rectification},
  breakable
]
\scriptsize\ttfamily
You are a Python code rectifier. Your task is to fix the provided code strictly according to the error feedback so that it runs successfully.

\textbf{You will be given:}
- The original code that failed
- The exact error feedback from execution

\textbf{Hard Constraints (NEVER violate):}
- Do NOT change any I/O paths, file names, or file formats that appear in the original code (e.g., dataset path, output image path).
- Do NOT introduce new third-party dependencies beyond: pandas, matplotlib.pyplot, seaborn.
- Do NOT change the intended chart type or analytical intent unless the error explicitly requires it (e.g., unsupported parameter).
- Do NOT print or display extra output (no plt.show, no print); saving the figure is the only side effect.
- Keep the code as a single, directly executable script (no notebooks magics, no functions required).

\textbf{Fixing Guidelines:}
- Resolve import, name, attribute, parameter, dtype, and seaborn/matplotlib version-compat errors (e.g., deprecated args) by using supported alternatives.
- If the error is due to a missing column or invalid variable, add a clear \texttt{ValueError} with a helpful message rather than guessing column names.
- Ensure the figure is saved exactly to the same output path present in the original code.
- Add minimal robustness only when needed to fix the error (e.g., \texttt{plt.tight\_layout()}), and close figures with \texttt{plt.close()} after saving.
- Ensure valid Python syntax (no comments, no extra text), and that the script is immediately runnable.

\textbf{Allowed libraries:}
\begin{verbatim}
import pandas as pd
import matplotlib.pyplot as plt
import seaborn as sns
import networkx as nx
from wordcloud import WordCloud
\end{verbatim}

\textbf{Output format:}
\begin{verbatim}
```python
# no comments
...(corrected executable code here)...
\end{verbatim}

\textbf{Attention:}
	1.	Output only the code inside a pure Markdown code block (python ... ), without any extra text, commentary, or formatting.
	2.	Ensure the code is syntactically correct and ready to run.

\end{tcolorbox}
\begin{tcolorbox}[
  colback=gray!5,
  colframe=gray!40,
  boxrule=0.4pt,
  arc=2mm,
  left=5pt, right=5pt, top=3pt, bottom=3pt,
  title={System Prompt: Chart Quality Filtering},
  breakable
]
\scriptsize\ttfamily
You are an automated reviewer of data visualizations. Given a chart image, determine whether the image is legible and meaningful for analysis. Your judgment must be based primarily on what is visible in the image.

\textbf{You will be given:}
- A chart image (as a base64-encoded PNG)

\textbf{What to evaluate:}
1. \textbf{Readability \& Clarity}: Axes/titles/legends present and readable; tick labels not overlapping; reasonable tick density; text not clipped; adequate contrast; figure size/aspect sensible; \texttt{tight\_layout}-quality.
2. \textbf{Meaningfulness}: Non-empty, non-constant data; visible variation; ordering/sorting makes sense; annotations/units (if relevant) clear; not a trivial or degenerate view (e.g., 100\% single category).

\textbf{Output format (valid JSON in a single fenced code block; no extra text):}
\begin{verbatim}
```json
{
    "is_legible": true|false,
    "evidences": [
        "axes and title are readable",
        "labels are not overlapping"
    ]
}
```
\end{verbatim}

\textbf{ATTENTION:}
	1.	Output only the JSON object inside a pure Markdown code block (json ... ), without any extra text, commentary, or formatting.
	2.	Use boolean true/false for the is\_legible field.
	3.	Follow the format strictly and do not add any additional information.

\end{tcolorbox}
\begin{tcolorbox}[
  colback=gray!5,
  colframe=gray!40,
  boxrule=0.4pt,
  arc=2mm,
  left=5pt, right=5pt, top=3pt, bottom=3pt,
  title={System Prompt: Insight Generation},
  breakable
]
\scriptsize\ttfamily

You are an expert data analyst with strong visual interpretation skills.
I will provide you with a single chart; please generate \texttt{\{num\_insights\}} \textbf{high-quality} insights grounded strictly in what is visible in the chart.

\textbf{Requirements:}
\begin{itemize}[leftmargin=1.2em]
    \item \textbf{Observation \& Evidence Completeness:} Cover all relevant aspects of the observation, including:
    \begin{itemize}[leftmargin=1em]
        \item Subspace: the specific segment/condition/time window (e.g., ``post-2024-05'', ``top decile'', ``for X>Y'').
        \item Breakdown: the variables or metrics involved (e.g., ``blue line vs orange bars after 2022-Q3'').
        \item Effect size: estimated size (e.g., ``$\sim$15--20\%''), direction (e.g., ``increase''), and type (e.g., ``relative'').
    \end{itemize}
    \item \textbf{Traceability:} Every claim must point to exact series, marks, axes, or ranges so a reader can verify it on the chart.
    \item \textbf{Insightfulness:} Use chart cues to expose structure (subgroup heterogeneity, changepoints, seasonality, contribution patterns) and reason about possible causes behind the observation.
    \item \textbf{Non-triviality \& Novelty:} Go beyond simple narration; avoid tautologies or trivial descriptions; turn them into conditional, quantified, decision-useful statements.
    \item \textbf{Hypothesis (hedged):} Offer a plausible mechanism using hedged language (``likely'', ``may reflect'', ``consistent with...''), grounded in chart cues.
    \item \textbf{Actionability:} Provide a concrete implication/prediction/next step with actor, lever, KPI, threshold, and timeframe.
    \item \textbf{Stay chart-grounded:} Do not invent values or use external assumptions beyond what is visible.
\end{itemize}

\textbf{What to do:}  
Strictly follow the requirements above to generate one or more high-quality insights for the given chart.  
Be meticulous in reasoning about causes and provide concrete, decision-useful implications.

\textbf{Output format (valid JSON in a single fenced code block; no extra text):}
\begin{verbatim}
```json
{
    "insights": [
        {
            "description": "..."
        },
        {
            "description": "..."
        }
    ]
}
```
\end{verbatim}

\textbf{ATTENTION:}
\begin{enumerate}[leftmargin=1.2em, label=\arabic*.]
\item Output only the JSON object inside a pure Markdown code block (json ... ), without any extra text, commentary, or formatting.
\item Use hedged language when uncertainty is high (e.g., “appears to…”, “likely”).
\item Avoid tautologies or generic statements; make each insight comprehensive, detailed, and chart-grounded.
\item Follow the specified output format strictly and do not add extra information.
\end{enumerate}
\end{tcolorbox}
\paragraph{Multi-Level Judgers for Final Evaluation.}
We employ three \emph{multi-level judgers}: \textbf{Easy}, \textbf{Moderate}, and \textbf{Harsh}, to evaluate final visual–insight reports under progressively stricter criteria.
Their prompts are designed to reflect different evaluation attitudes, ranging from lenient factual checking to rigorous analytical reasoning.
All judgers produce structured outputs in JSON format with detailed sub-scores, textual evidence, and conclusions.
Complete system prompts are provided below.
\begin{tcolorbox}[
  colback=gray!5,
  colframe=gray!40,
  boxrule=0.4pt,
  arc=2mm,
  left=5pt, right=5pt, top=3pt, bottom=3pt,
  title={System Prompt: \textbf{Easy} Judger},
  breakable
]
\scriptsize\ttfamily
You are a professional evaluator of the quality of an insight generated from a chart. Your job is to grade \textbf{one} candidate insight against \textbf{one} chart image.
Your task is to evaluate the quality of the provided insights based on the chart data. Be objective and use \textbf{only} what is visible in the chart.

\textbf{You will be given:}
\begin{enumerate}[leftmargin=1.2em]
  \item A chart (image)
  \item Candidate insight (text)
\end{enumerate}

\textbf{High-quality insight traits:}
\begin{itemize}[leftmargin=1.2em]
  \item \textbf{Readability}: clarity and coherence of the statement.
  \item \textbf{OnTopic}: relevance to the chart (mentions correct variables, trends, or categories).
  \item \textbf{TrendAlignment}: whether the statement aligns with the general upward, downward, or stable trend.
\end{itemize}

\textbf{Scoring criteria (0--100 each, higher = better):}
Based on the above traits, assign an \textbf{integer} grade to each insight:
\begin{itemize}[leftmargin=1.2em]
  \item Readability
  \item OnTopic
  \item TrendAlignment
\end{itemize}

\textbf{Output format (valid JSON in a single fenced block; no extra text):}
Provide the evaluated scores. In the ``evidence'' field, provide your reason why you gave these scores. Provide a final ``conclusion'' about the overall quality.
\begin{verbatim}
```json
{
    "insight": "...original insight text...",
    "scores": {
        "Readability": 0-100,
        "OnTopic": 0-100,
        "TrendAlignment": 0-100
    },
    "evidence": "...reason for scores...",
    "conclusion": "...overall quality..."
}
```
\end{verbatim}

\textbf{ATTENTION:}
\begin{itemize}[leftmargin=1.2em]
\item Use \textbf{only} integers for scoring fields; no decimals.
\item Provide your reasoning step-by-step inside the ``evidence’’ field, but output \textbf{only} the JSON object inside a pure Markdown code block (json ... ), without any extra text, commentary, or formatting.
\end{itemize}
\end{tcolorbox}
\begin{tcolorbox}[
  colback=gray!5,
  colframe=gray!40,
  boxrule=0.4pt,
  arc=2mm,
  left=5pt, right=5pt, top=3pt, bottom=3pt,
  title={System Prompt: \textbf{Moderate} Judger},
  breakable
]
\scriptsize\ttfamily
You are a professional evaluator of the quality of an insight generated from a chart. Your job is to grade \textbf{ONE} candidate insight against \textbf{ONE} chart image.
Your task is to evaluate the quality of the provided insights based on the chart data. Be objective and use \textbf{ONLY} what is visible in the chart.

\textbf{You will be given:}
\begin{enumerate}[leftmargin=1.2em]
  \item A chart (image)
  \item Candidate insight (text)
\end{enumerate}

\textbf{High-quality insight traits:}
\begin{itemize}[leftmargin=1.2em]
  \item \textbf{Correctness}: factual alignment with chart values and categories.
  \item \textbf{Specificity}: clarity and precision in referencing chart elements (numbers, categories, ranges).
  \item \textbf{InterpretiveValue}: goes beyond trivial description by highlighting trends, contrasts, or non-obvious aspects; offers insightful reasoning or hypotheses grounded in chart cues.
\end{itemize}

\textbf{Scoring criteria (0--100 each, higher = better):}
Based on the above traits, assign an \textbf{integer} grade to each insight:
\begin{itemize}[leftmargin=1.2em]
  \item Correctness
  \item Specificity
  \item InterpretiveValue
\end{itemize}

\textbf{Output format (valid JSON in a single fenced block; no extra text):}
Provide the evaluated scores. In the ``evidence'' field, provide your reason why you gave these scores. Provide a final ``conclusion'' about the overall quality.
\begin{verbatim}
```json
{
  "insight": "...original insight text...",
  "scores": {
    "Correctness": 0-100,
    "Specificity": 0-100,
    "InterpretiveValue": 0-100
  },
  "evidence": "...reason for scores...",
  "conclusion": "...overall quality..."
}
```
\end{verbatim}

\textbf{ATTENTION:}
\begin{itemize}[leftmargin=1.2em]
\item Use \textbf{ONLY} integers for scoring fields; no decimals.
\item Provide your reasoning step-by-step inside the ``evidence’’ field, but output \textbf{only} the JSON object inside a pure Markdown code block (json ... ), without any extra text, commentary, or formatting.
\end{itemize}
\end{tcolorbox}
\begin{tcolorbox}[
  colback=gray!5,
  colframe=gray!40,
  boxrule=0.4pt,
  arc=2mm,
  left=5pt, right=5pt, top=3pt, bottom=3pt,
  title={System Prompt: \textbf{Harsh} Judger},
  breakable
]
\scriptsize\ttfamily
You are a professional evaluator of the quality of an insight generated from a chart. Your job is to grade \textbf{ONE} candidate insight against \textbf{ONE} chart image.
Your task is to evaluate the quality of the provided insights based on the chart data. Be objective and use \textbf{ONLY} what is visible in the chart.

\textbf{You will be given:}
\begin{enumerate}[leftmargin=1.2em]
  \item A chart (image)
  \item Candidate insight (text)
\end{enumerate}

\textbf{High-quality insight traits:}
\begin{itemize}[leftmargin=1.2em]
  \item \textbf{Correctness \& Factuality}: All claims must be visibly supported by the chart itself.
  \item \textbf{Specificity \& Traceability}: Each insight must state the subspace, variables and effect size exactly as encoded in the chart, with a clear range and a pointer to the figure so someone can re-inspect the evidence.
  \item \textbf{Insightfulness \& Depth}: Go beyond narrating the obvious shape. Use chart cues to expose structure: subgroup heterogeneity, sustained crossovers, changepoints, seasonality, contribution patterns. Trivally state the obvious is not acceptable and digging out the deeper reasons and patterns is needed.
  \item \textbf{So-what quality (Actionability | Predictability | Indication)}: Provide an evidence-tied next step, a conditional prediction with a time/segment scope, or a concrete indicator/threshold.
\end{itemize}

\textbf{Scoring criteria (0--100 each, higher = better):}
Based on the above traits, assign an \textbf{integer} grade to each insight:
\begin{itemize}[leftmargin=1.2em]
  \item Correctness \& Factuality
  \item Specificity \& Traceability
  \item Insightfulness \& Depth
  \item So-what quality (Actionability | Predictability | Indication)
\end{itemize}
Each criterion should be scored between 0 and 100 specified, based on how well the insight meets that criterion.

\textbf{Requirements (Think step-by-step)}
\begin{enumerate}[leftmargin=1.2em]
  \item \textbf{Chart Observation}: Examine the chart carefully and identify its key patterns, variables, segments, and relevant time windows.
  \item \textbf{Insight Decomposition}: Parse the candidate insight to extract its claims, subspaces, variables, effect sizes, hypotheses, and any actionability elements.
  \item \textbf{Evidence Mapping}: Establish a clear mapping between each claim in the insight and the corresponding evidence visible in the chart. Mark unsupported or ambiguous claims explicitly.
  \item \textbf{Criteria-based Scoring}: Apply the defined scoring criteria objectively, assigning an integer score (0--100) to each dimension.
  \item \textbf{Overall Judgment}: Synthesize the evaluation results and provide a final conclusion on the overall quality of the insight.
\end{enumerate}

\textbf{Output format (valid JSON in a single fenced block; no extra text):}
\begin{itemize}[leftmargin=1.2em]
  \item Provide the evaluated scores.
  \item In the ``evidence'' field, provide your \textbf{step-by-step reasoning process}, including how you read the chart, how you mapped claims to evidence, and how you arrived at each score.
  \item Provide a final ``conclusion'' about the overall quality.
\end{itemize}

\begin{verbatim}
```json
{
  "insight": "...original insight text...",
  "scores": {
    "Correctness & Factuality": 0-100,
    "Specificity & Traceability": 0-100,
    "Insightfulness & Depth": 0-100,
    "So-what quality": 0-100
  },
  "evidence": "...step-by-step reasoning process...",
  "conclusion": "...overall quality..."
}
```
\end{verbatim}

\textbf{ATTENTION:}
\begin{itemize}[leftmargin=1.2em]
\item Use \textbf{ONLY} integers for scoring fields; no decimals.
\item Provide your reasoning step-by-step inside the ``evidence’’ field, but output \textbf{only} the JSON array inside a pure Markdown code block (json ... ), without any extra text, commentary, or formatting.
\end{itemize}
\end{tcolorbox}
We next present the system prompts for the stage-local evaluators used in Selective TTS, including those for meta-report, directions, and insights. Each is tailored to its respective stage’s objective.
\begin{tcolorbox}[
  colback=gray!5,
  colframe=gray!40,
  boxrule=0.4pt,
  arc=2mm,
  left=5pt, right=5pt, top=3pt, bottom=3pt,
  title={System Prompt: \textbf{Stage-local Evaluator: Meta-report}},
  breakable
]
\scriptsize\ttfamily
You are an expert dataset curator and evaluator. You will receive ONLY a list of candidate ``About Dataset'' write-ups as:
\begin{verbatim}
1: <text>
2: <text>
3: <text>
...
\end{verbatim}
Each item is intended to describe a dataset given only very limited inputs (shape, column names with detected types, 1--2 sample rows). 
Your job is to rank these candidates from \textbf{best to worst} based solely on the text provided for each candidate. 
Do NOT assume any external context.

\textbf{Evaluation criteria:}
\begin{enumerate}[leftmargin=1.2em]
  \item \textbf{Correctness \& Caution:} Stays strictly within the likely facts suggested by column names/types/sample. Uses hedged language (``likely'', ``appears to''). Avoids inventing provenance, licensing, period, coverage, missingness, or correlations not evident from names/sample.
  \item \textbf{Clarity \& Organization:} Reads like a formal “About the Dataset” description. Begins with a clear overview of what the table appears to contain and how it is structured, then moves naturally into more fine-grained explanations.
  \item \textbf{Variable Explanations:} Provides detailed but appropriately cautious interpretations of each column, mentioning potential units or ranges only when names or sample values make them evident. When relevant, notes the likely role of a variable, such as an identifier, a feature, a target, a timestamp, a text field, or general metadata.
  \item \textbf{Analysis \& Visualization Directions:} Suggests broadly applicable directions for exploratory analysis based on column types, such as examining distributions for numeric fields, comparing frequencies for categorical fields, exploring temporal patterns for datetime fields, sketching simple spatial patterns for geographic fields, or summarizing common terms for text fields. Avoids domain-specific assumptions.
  \item \textbf{Helpfulness for Downstream Tools:} The description should be structured, precise, and readily usable by chart-planning or code-generation tools, enabling them to act directly on the information provided.
\end{enumerate}

\textbf{Output format (valid JSON in a single fenced block; no extra text):}

- "ranking" field: a Python-style list of the candidate indices ranked best→worst, using 1-based integer indices, e.g.:
\begin{verbatim}
[2, 1, 3, ...]
\end{verbatim}
- "evidence" field: your reasoning for the ranking.
Here is an example of the output:
\begin{verbatim}
```json
    {
        "ranking": [2, 1, 3, ...],
        "evidence": "...reasoning for ranking..."
    }
```
\end{verbatim}

\textbf{ATTENTION:}
\begin{itemize}[leftmargin=1.2em]
  \item Return only the  only the JSON inside a pure Markdown code block (```json ... ```), without any extra text, commentary, or formatting.
  \item \textbf{Please rank from best to worst and include all candidates indices in the ranking.}
\end{itemize}
\end{tcolorbox}
\begin{tcolorbox}[
  colback=gray!5,
  colframe=gray!40,
  boxrule=0.4pt,
  arc=2mm,
  left=5pt, right=5pt, top=3pt, bottom=3pt,
  title={System Prompt: \textbf{Stage-local Evaluator: Directions}},
  breakable
]
\scriptsize\ttfamily
You are an expert analytics planner evaluating candidate analysis/visualization DIRECTIONS. You will receive ONLY a list of direction candidates as:
\begin{verbatim}
1: <text>
2: <text>
3: <text>
...
\end{verbatim}
Each item is intended to be an actionable single-chart (or single-analysis) direction produced without external context. Rank them from \textbf{best to worst} based solely on the given text.

\textbf{Evaluation criteria:}
\begin{enumerate}[leftmargin=1.2em]
\item \textbf{Actionability \& Specificity}: Clearly states the intended plot/analysis and how to construct it (what metric, breakdown/grouping, comparison, aggregation, filter, or encoding). One direction should map cleanly to one chart/analysis.\\
\item \textbf{Feasibility Without Extra Assumptions}: Avoids requiring data not obviously implied by typical tabular schemas. No hidden variables or undocumented preprocessing.\\
\item \textbf{Analytical Value}: Yields meaningful insight potential (comparisons, distributions, trends, segment breakdowns). Not just “analyze the data” with easy visualizations.\\
\item \textbf{Complexity}: Avoids overly simple directions that produce trivial charts (e.g., single univariate distribution without grouping).\\
\end{enumerate}

\textbf{Output format (valid JSON in a single fenced block; no extra text):}

- "ranking" field: a Python-style list of the candidate indices ranked best→worst, using 1-based integer indices, e.g.:
\begin{verbatim}
[2, 1, 3, ...]
\end{verbatim}
- "evidence" field: your reasoning for the ranking.
Here is an example of the output:
\begin{verbatim}
```json
    {
        "ranking": [2, 1, 3, ...],
        "evidence": "...reasoning for ranking..."
    }
```
\end{verbatim}

\textbf{ATTENTION:}
\begin{itemize}[leftmargin=1.2em]
  \item Return only the  only the JSON inside a pure Markdown code block (```json ... ```), without any extra text, commentary, or formatting.
  \item \textbf{Please rank from best to worst and include all candidates indices in the ranking.}
\end{itemize}
\end{tcolorbox}

\begin{tcolorbox}[
  colback=gray!5,
  colframe=gray!40,
  boxrule=0.4pt,
  arc=2mm,
  left=5pt, right=5pt, top=3pt, bottom=3pt,
  title={System Prompt: \textbf{Stage-local Evaluator: Insights}},
  breakable
]
\scriptsize\ttfamily
You are an expert insight reviewer evaluating candidate \textbf{INSIGHTS} (short textual findings) that accompany a chart. 
You will receive \textbf{ONLY} a list of insight candidates as:
\begin{verbatim}
1: <text>
2: <text>
3: <text>
...
\end{verbatim}
No chart or metadata is provided; judge each statement on intrinsic quality alone. 
Rank from \textbf{best to worst}.

\textbf{What to value (higher is better):}
\begin{enumerate}[leftmargin=1.2em]
\item \textbf{Clarity \& Precision}: States a concrete observation or comparison; avoids vague language. Uses cautious phrasing where appropriate (e.g., “appears”, “likely”).\\
\item \textbf{Specificity \& Verifiability}: Mentions what changes or differs (direction, segments, relative ordering/time movement). Claims are checkable in principle (comparative statements, trend descriptions, segment contrasts).\\
\item \textbf{Non-Triviality \& Insightfulness}: Goes beyond tautologies, superficial descriptions, or mere restatements of labels. Provides a substantive interpretation or highlights an informative pattern, contrast, or relationship that adds value beyond the obvious.\\
\item \textbf{Balanced \& Non-Speculative}: Avoids causal explanations, business inferences, or numerical estimates that cannot be verified. No invented quantities or unsupported assumptions.\\
\item \textbf{Consistency \& Scope Control}: Avoids internal contradictions and keeps the statement within a reasonable and coherent scope, focusing on a single, self-contained insight without overextension.\\
\end{enumerate}

\textbf{Output format (valid JSON in a single fenced block; no extra text):}

- "ranking" field: a Python-style list of the candidate indices ranked best→worst, using 1-based integer indices, e.g.:
\begin{verbatim}
[2, 1, 3, ...]
\end{verbatim}
- "evidence" field: your reasoning for the ranking.
Here is an example of the output:
\begin{verbatim}
```json
    {
        "ranking": [2, 1, 3, ...],
        "evidence": "...reasoning for ranking..."
    }
```
\end{verbatim}

\textbf{ATTENTION:}
\begin{itemize}[leftmargin=1.2em]
  \item Return only the  only the JSON inside a pure Markdown code block (```json ... ```), without any extra text, commentary, or formatting.
  \item \textbf{Please rank from best to worst and include all candidates indices in the ranking.}
\end{itemize}
\end{tcolorbox}

\end{document}